\documentclass[twoside,11pt]{article}

\usepackage{jmlr2e}

\jmlrheading{1}{2020}{1-48}{4/00}{10/00}{kreidieh2019a}{Abdul Rahman Kreidieh and Alexandre M. Bayen}

\ShortHeadings{}{}
\firstpageno{1}

\usepackage{microtype}
\usepackage{graphicx}
\usepackage{subcaption, floatrow}
\usepackage{booktabs} 
\usepackage{wrapfig}  

\usepackage[utf8]{inputenc} 
\usepackage[T1]{fontenc}    
\usepackage{hyperref}       
\usepackage{url}            
\usepackage{booktabs}       
\usepackage{amsfonts}       
\usepackage{nicefrac}       
\usepackage{microtype}      
\usepackage{amsmath}
\usepackage{caption}
\usepackage{algorithm}
\usepackage{algpseudocode}
\algnewcommand{\algorithmicforeach}{\textbf{for each}}
\algdef{SE}[FOR]{ForEach}{EndForEach}[1]
  {\algorithmicforeach\ #1\ \algorithmicdo}
  {\algorithmicend\ \algorithmicforeach}
\newcommand{\pushright}[1]{\ifmeasuring@#1\else\omit\hfill$\displaystyle#1$\fi\ignorespaces}
\algrenewcommand\alglinenumber[1]{\tiny #1:}

\usepackage{enumitem}
\usepackage{todonotes}
\usepackage{soul}
\usepackage{amsmath}
\usepackage{listings}  
\usepackage{xspace}

\usepackage{floatrow}
\newfloatcommand{capbtabbox}{table}[][\FBwidth]

\newcommand{\methodName}[0]{CHER\xspace}
\newcommand{\antgather}[0]{Ant Gather\xspace}
\newcommand{\antmaze}[0]{Ant Maze\xspace}
\newcommand{\antfourrooms}[0]{Ant Four Rooms\xspace}

\newcommand{\highwaysingle}[0]{Highway\xspace}
\newcommand{\ringroad}[0]{Ring Road\xspace}

\usepackage{pgf,pgfplots}
\usetikzlibrary{babel} 
\usetikzlibrary{arrows,decorations.markings}
\usepackage{multicol}
\usepgfplotslibrary{fillbetween}
\usetikzlibrary{patterns}

\definecolor{color1}{RGB}{0, 0, 255}
\definecolor{color2}{RGB}{255, 129, 19} 
\definecolor{color3}{RGB}{157, 116, 195} 
\definecolor{color4}{RGB}{52, 164, 52} 
\definecolor{color5}{RGB}{144, 93, 82} 
\definecolor{color6}{RGB}{200, 0, 0} 
\definecolor{limegreen}{RGB}{50, 205, 50} 
\allowdisplaybreaks[1]

\usepackage{tabularx,caption,ragged2e}
\newcolumntype{Y}{>{\RaggedRight\arraybackslash}X}

\begin{document}

\title{Inter-Level Cooperation in\\ Hierarchical Reinforcement Learning}

%
\author{\name Abdul Rahman Kreidieh 
	\email aboudy@berkeley.edu \\[-15pt]
    \AND
    \name Glen Berseth 
	\email gberseth@berkeley.edu \\[-15pt]
    \AND
    \name Brandon Trabucco
	\email btrabucco@berkeley.edu \\[-15pt]
    \AND
    \name Samyak Parajuli
	\email samyak.parajuli@berkeley.edu \\[-15pt]
    \AND
    \name Sergey Levine
	\email svlevine@eecs.berkeley.edu \\[-15pt]
    \AND
    \name Alexandre M. Bayen
	\email bayen@berkeley.edu \\
    \addr University of California, Berkeley
}
\date{\vspace{-5ex}}
\editor{}
\maketitle

\medskip


\begin{abstract}
Hierarchies of temporally decoupled policies present a promising approach for enabling structured exploration in complex long-term planning problems. To fully achieve this approach an end-to-end training paradigm is needed. However, training these multi-level policies has had limited success due to challenges arising from interactions between the goal-assigning and goal-achieving levels within a hierarchy. In this article, we consider the policy optimization process as a multi-agent process. This allows us to draw on connections between communication and cooperation in multi-agent RL, and demonstrate the benefits of increased cooperation between sub-policies on the training performance of the overall policy. We introduce a simple yet effective technique for inducing inter-level cooperation by modifying the objective function and subsequent gradients of higher-level policies. Experimental results on a wide variety of simulated robotics and traffic control tasks demonstrate that inducing cooperation results in stronger performing policies and increased sample efficiency on a set of difficult long time horizon tasks. We also find that goal-conditioned policies trained using our method display better transfer to new tasks, highlighting the benefits of our method in learning task-agnostic lower-level behaviors. Videos and code are available at: \url{https://sites.google.com/berkeley.edu/cooperative-hrl}.
\end{abstract}



\section{Introduction} \label{sec:introduction}

To solve interesting problems in the real world agents must be adept at planning and reasoning over long time horizons. For instance, in robot navigation and interaction tasks, 
agents must learn to compose lengthy sequences of actions to achieve long-term goals. In other environments, such as mixed-autonomy traffic control settings~\citep{wu2017emergent, vinitsky2018benchmarks}, 
exploration is delicate, as individual actions may not influence the flow of traffic until multiple timesteps in the future. RL has had limited success in solving long-horizon planning problems such as these without relying on task-specific reward shaping strategies that limit the performance of resulting policies~\citep{wu2017flow} or additional task decomposition techniques that are not transferable to different tasks~\citep{sutton1999between, kulkarni2016hierarchical, 2017-TOG-deepLoco, florensa2017stochastic}.

\begin{figure*}
\centering
\begin{subfigure}[b]{\textwidth}
\centering
\begin{tikzpicture}
    \newcommand \boxwidth {1.0}
    \newcommand \boxheight {\textwidth}
    \newcommand \boxoffset {0}

    \draw (0, -0.15) node {};

    \filldraw [limegreen] (0.305 * \textwidth,\boxwidth + \boxoffset - 0.5) circle (0.125);
    \draw [anchor=west]
    (0.32 * \textwidth, \boxwidth + \boxoffset - 0.5) node {\scriptsize Apple ($+1$)};

    \filldraw [red] (0.46 * \textwidth,\boxwidth + \boxoffset - 0.5) circle (0.125);
    \draw [anchor=west]
    (0.475 * \textwidth, \boxwidth + \boxoffset - 0.5) node {\scriptsize Bomb ($-1$)};

    \filldraw [blue] (0.615 * \textwidth, \boxwidth + \boxoffset - 0.5) circle (0.125);
    \draw [anchor=west]
    (0.63 * \textwidth, \boxwidth + \boxoffset - 0.5) node {\scriptsize High-level goal};

    \draw [dashed] 
    (0.790 * \textwidth, \boxwidth + \boxoffset - 0.5) -- 
    (0.825 * \textwidth, \boxwidth + \boxoffset - 0.5); 
    \draw [anchor=west]
    (0.83 * \textwidth, \boxwidth + \boxoffset - 0.5) node {\scriptsize Agent trajectory};
\end{tikzpicture}
\end{subfigure} \\ \vspace{-0.4cm}
\begin{tikzpicture}
    \draw[->] (0, 0) -- (0.06*\textwidth, 0);
    \draw[anchor=west] (0.065*\textwidth, 0) node {\scriptsize training iteration};

    \draw[->] (0.52*\textwidth, 0) -- (0.58*\textwidth, 0);
    \draw[anchor=west] (0.585*\textwidth, 0) node {\scriptsize training iteration};
    \draw (\textwidth, 0) {};
\end{tikzpicture}\\[-15pt]
\begin{subfigure}[b]{0.48\textwidth}
    \begin{subfigure}[b]{0.3\textwidth}
        \includegraphics[width=\linewidth]{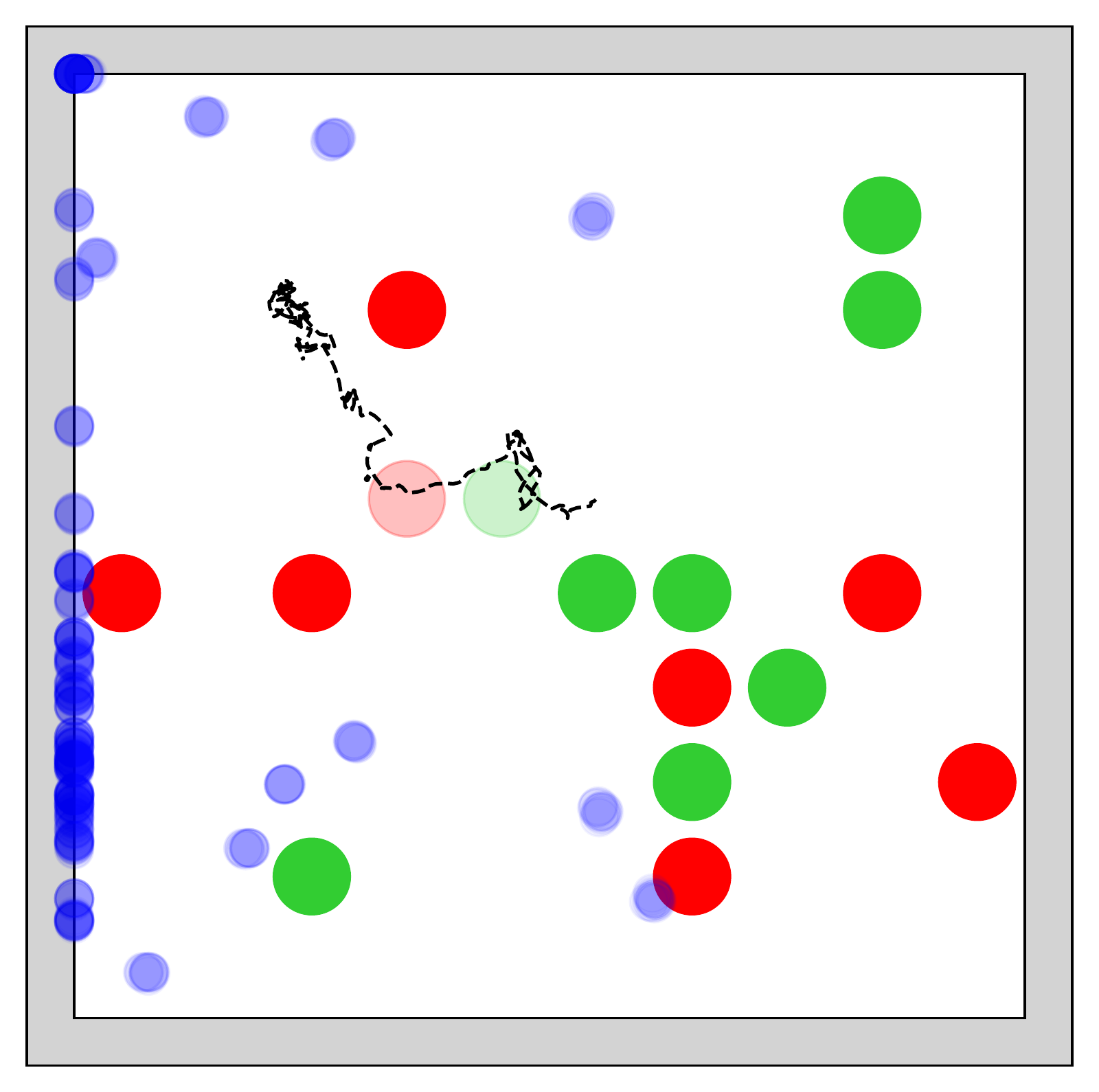}
    \end{subfigure}
    \hfill
    \begin{subfigure}[b]{0.3\textwidth}
        \includegraphics[width=\linewidth]{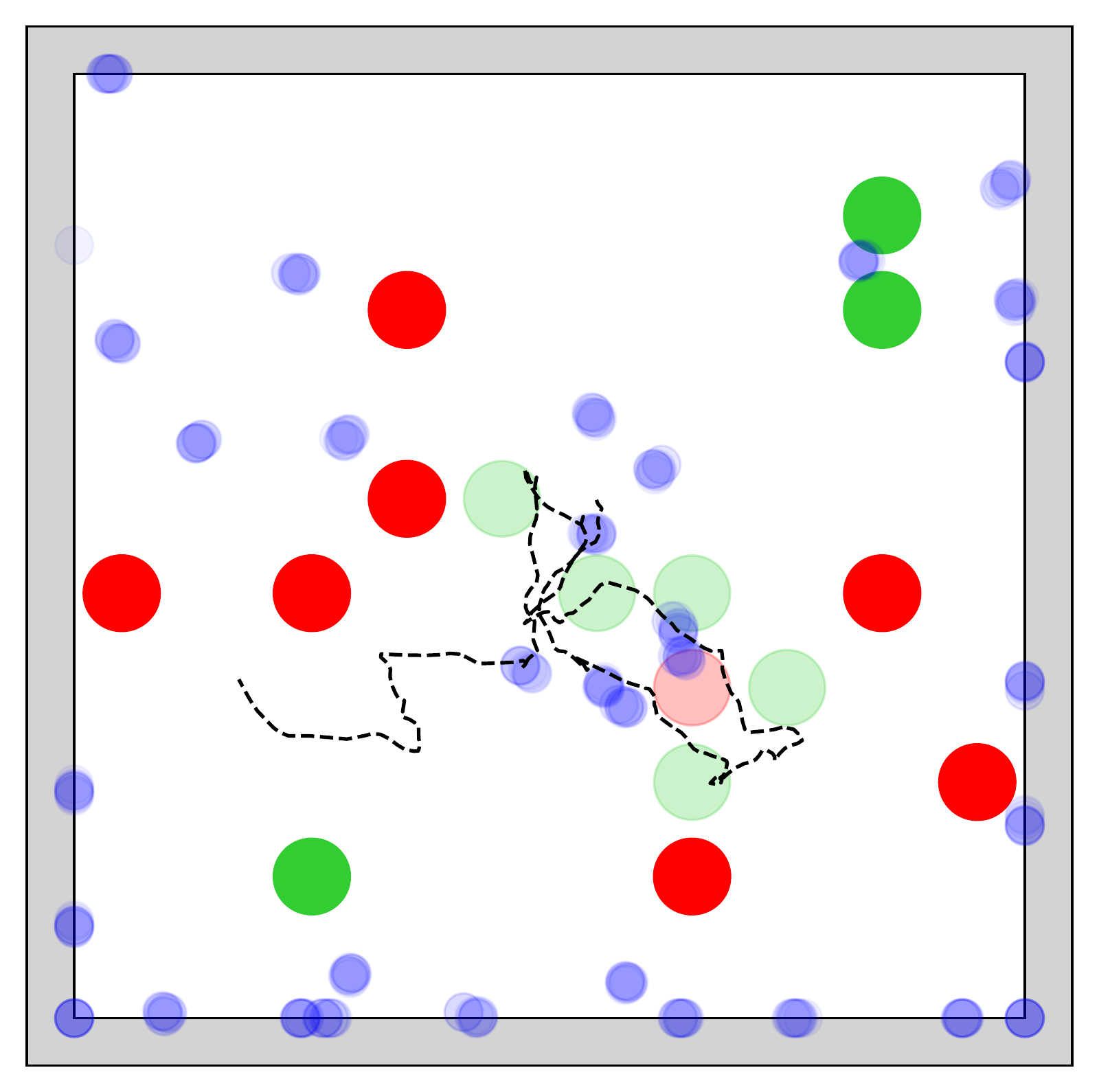}
    \end{subfigure}
    \hfill
    \begin{subfigure}[b]{0.3\textwidth}
        \includegraphics[width=\linewidth]{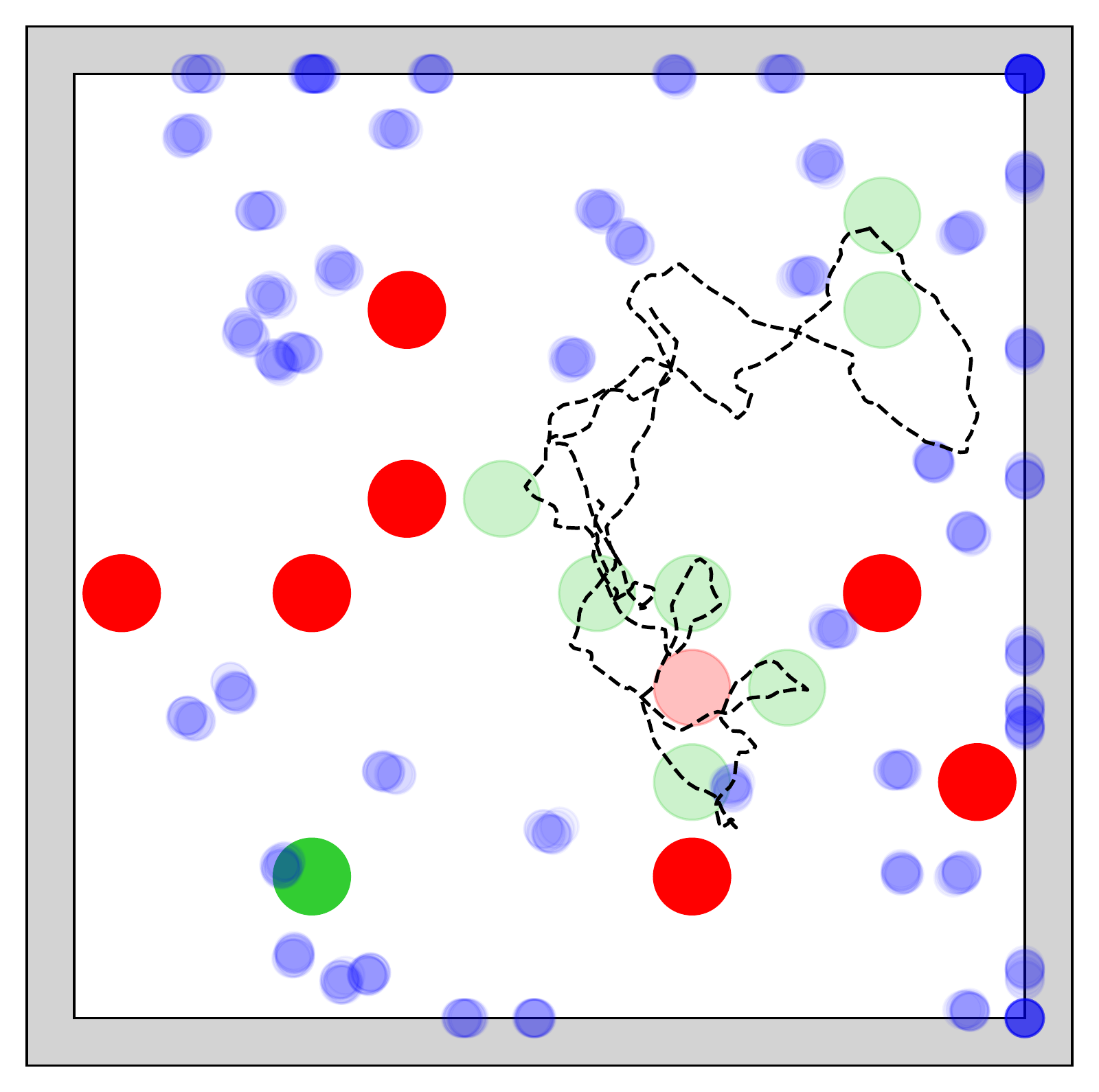}
    \end{subfigure}
    \caption{Standard HRL}
    \label{fig:goaldists-standrd}
\end{subfigure}
\hfill
\begin{tikzpicture}
    \draw [dashed] (0,0) -- (0,3.25);
\end{tikzpicture}
\hfill
\begin{subfigure}[b]{0.48\textwidth}
    \begin{subfigure}[b]{0.3\textwidth}
        \includegraphics[width=\linewidth]{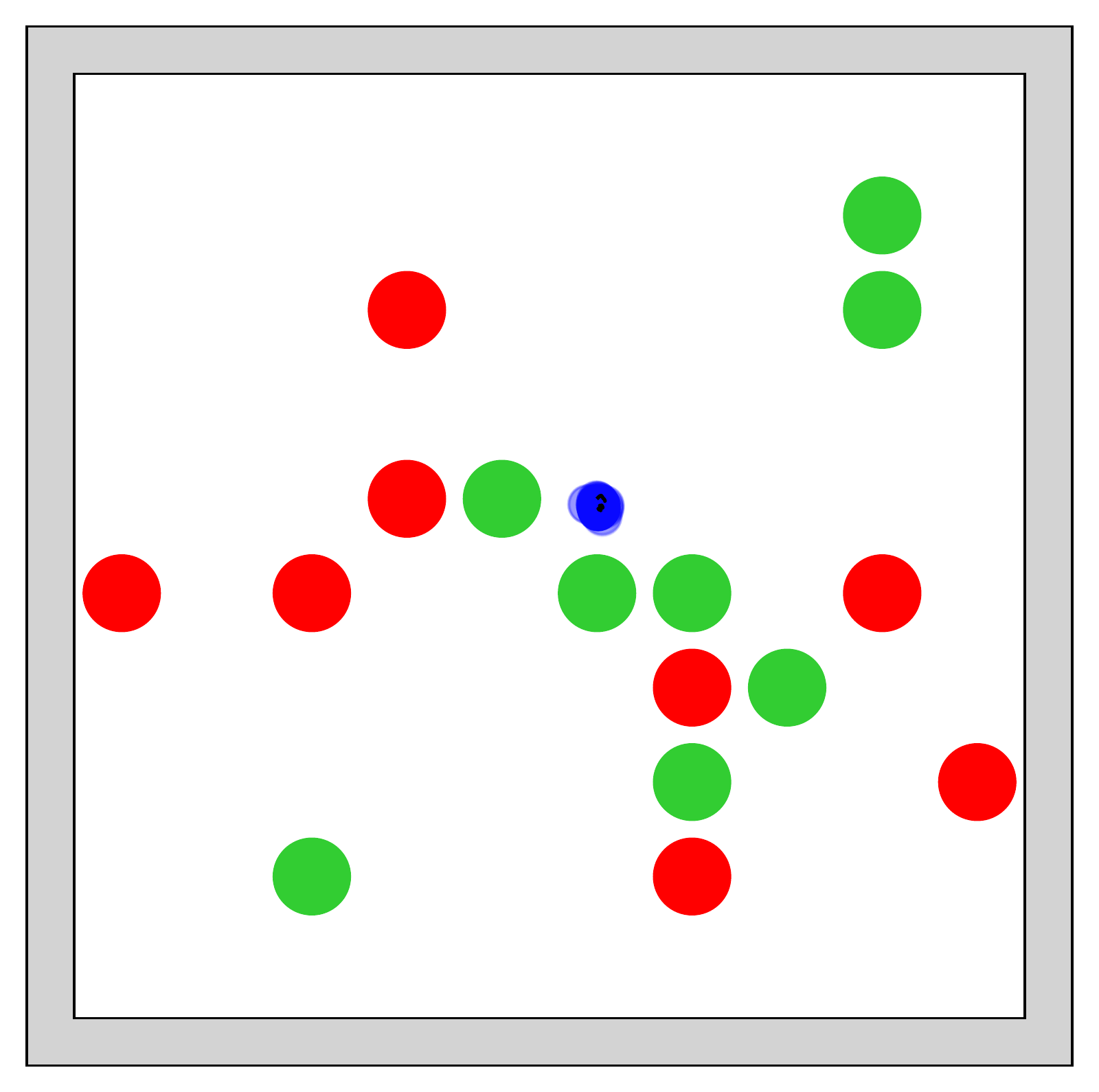}
    \end{subfigure}
    \hfill
    \begin{subfigure}[b]{0.3\textwidth}
        \includegraphics[width=\linewidth]{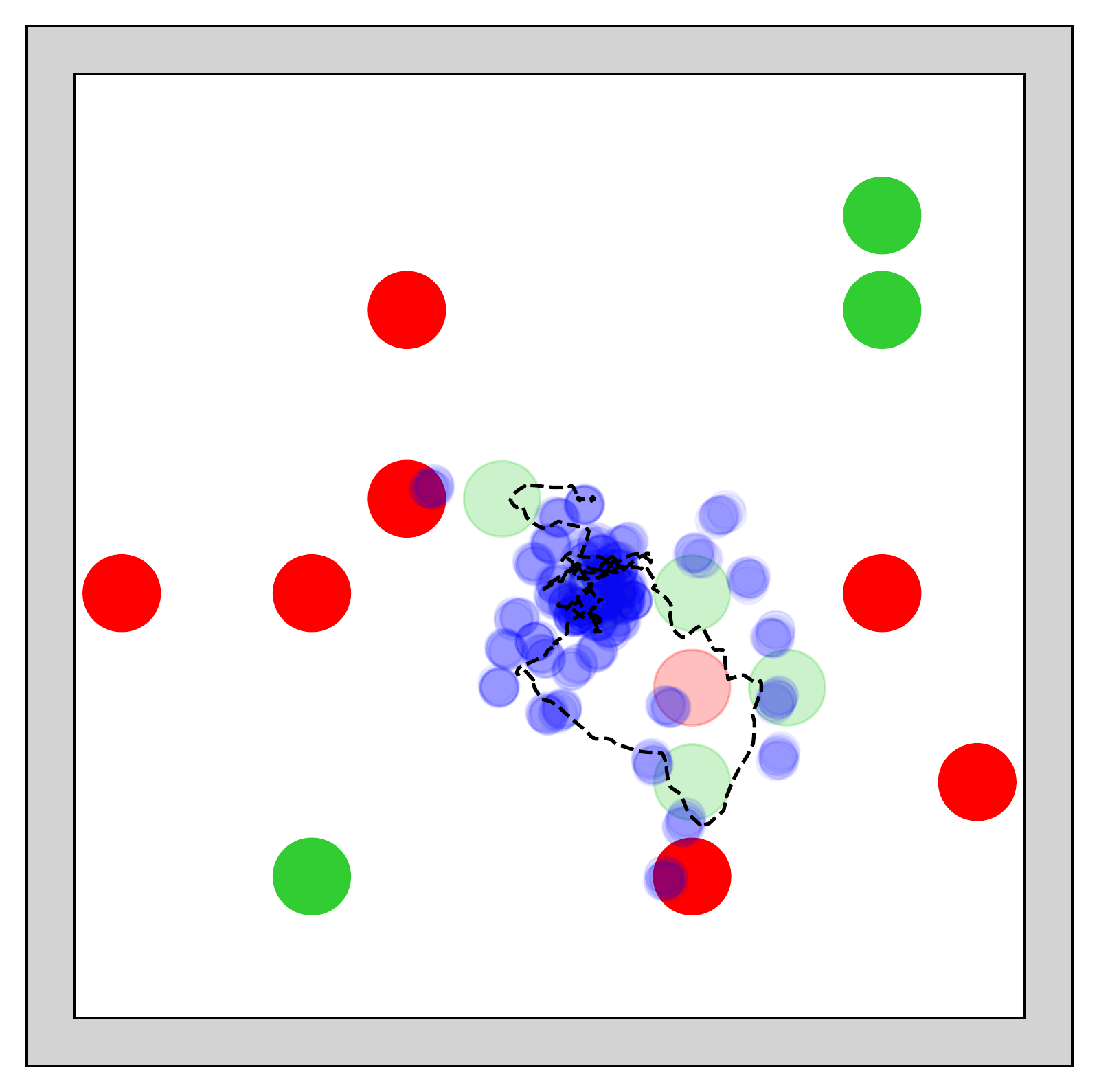}
    \end{subfigure}
    \hfill
    \begin{subfigure}[b]{0.3\textwidth}
        \includegraphics[width=\linewidth]{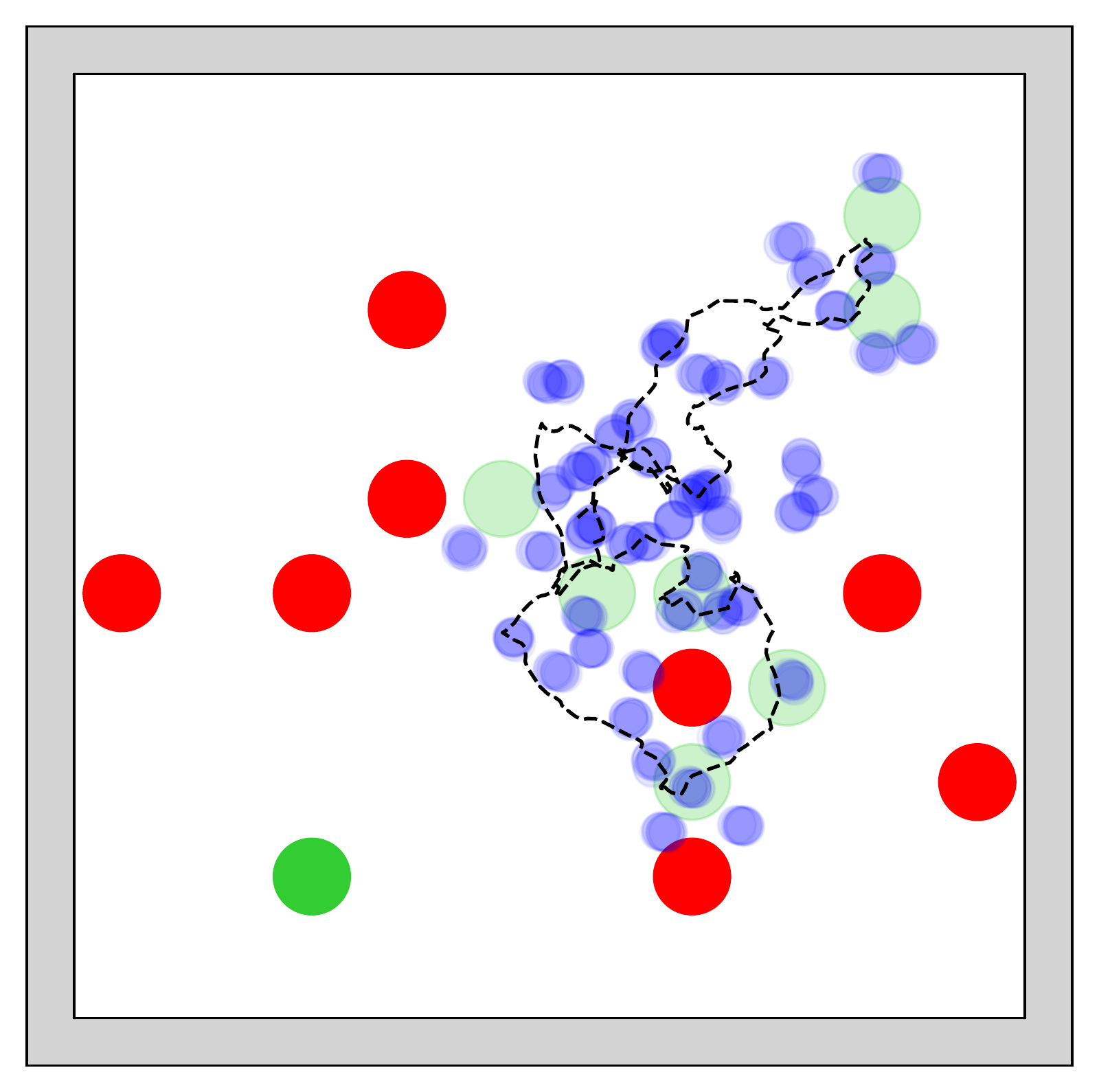}
    \end{subfigure}
    \caption{Cooperative HRL}
    \label{fig:goaldists-cooperative}
\end{subfigure}
\caption{Here the learned goal proposal distributions are shown for normal HRL and our method. Our agents develop more reasonable goal proposals that allow low-level policies to learn goal reaching skills quicker. The additional communication also allows the high level to better understand why proposed goals failed.}
\label{fig:goaldists}
\end{figure*}

Concurrent learning methods in hierarchical RL can improve the quality of learned hierarchical policies by flexibly updating both goal-assignment and goal-reaching policies to be better adapted to a given task~\citep{levy2017hierarchical,nachum2018data,li2019sub}. 
The process of simultaneously learning diverse skills and exploiting these skills to achieve a high-level objective, however, 
is an unstable and non-stationary optimization procedure that can be difficult to solve in practice.
In particular, at the early stages of training lower-level policies are unable to reach most goals assigned to them by a higher-level policy, and instead must learn to do so as training progresses. This inability of the lower-level to be able to reach assigned goals exacerbates the credit assignment problem from the perspective of the higher-level policy. It causes the higher-level policy to be unable to identify whether a specific goal under-performed as a result of the choice of goal or the lower-level policy's inability to achieve it. In practice, this results in the highly varying or random goal assignment strategies in~Figure~\ref{fig:goaldists-standrd} that require a large number of samples~\citep{li2019sub} and some degree of feature engineering~\citep{nachum2018data} to optimize over.

In this work, we show how adding cooperation between internal levels within a hierarchy\footnote{The connection of this problem to cooperation in multiagent RL is discussed in Section~\ref{sec:algorithm}.}, and subsequently introduce mechanisms that promote variable degrees of cooperation in HRL. Our method named \emph{Cooperative HiErarchical RL}, \methodName, improves cooperation by encouraging higher-level policies to specify goals that lower-level policies can succeed at, thereby disambiguating under-performing goals from goals that were unachievable by the lower-level policy. In Figure~\ref{fig:goaldists}(b) we show how \methodName changes the high-level goal distributions to be within the capabilities of the agent. This approach results in more informative communication between the policies. 
The distribution of goals or tasks the high-level will command of the lower level expands over time as the lower-level policy's capabilities increase.


A key finding in this article is that regulating the degree of cooperation between HRL layers can significantly impact the learned behavior by the policy. Too little cooperation may introduce no change to the goal-assignment behaviors and subsequent learning, while excessive cooperation may disincentivize an agent from making forward progress. We accordingly introduce a constrained optimization that serves to regulate the degree of cooperation between layers and ground the notion of cooperation in HRL to quantitative metrics within the lower-level policy. This results in a general and stable method for optimizing hierarchical policies concurrently.

We demonstrate the performance of \methodName on a collection of standard HRL environments and two previously unexplored mixed autonomy traffic control tasks. 
    For the former set of problems, we find that our method can achieve better performance compared to recent sample efficient off-policy and HRL algorithms. For the mixed autonomy traffic tasks, the previous HRL methods struggle while our approach subverts overestimation biases that emerge in the early stages of training, thereby allowing the controlled (autonomous) vehicles to regulate their speeds around the optimal driving of the task as opposed to continuously attempting to drive as fast as possible.
    When transferring lower-level policies between tasks, we find that policies learned via inter-level cooperation perform significantly better in new tasks without the need for additional training. This highlights the benefit of our method in learning generalizable policies.


\section{Background} \label{sec:background}

RL problems are generally studied as a \emph{Markov decision problem} (MDP) \citep{bellman1957markovian}, defined by the tuple: $(\mathcal{S}, \mathcal{A}, \mathcal{P}, r, \rho_0, \gamma, T)$, where $\mathcal{S} \subseteq \mathbb{R}^n$ is an $n$-dimensional state space, $\mathcal{A} \subseteq \mathbb{R}^m$ an $m$-dimensional action space, $\mathcal{P} : \mathcal{S} \times \mathcal{A} \times \mathcal{S} \to \mathbb{R}_+$ a transition probability function, $r : \mathcal{S} \to \mathbb{R}$ a reward function, $\rho_0 : \mathcal{S} \to \mathbb{R}_+$ an initial state distribution, $\gamma \in (0,1]$ a discount factor, and $T$ a time horizon. 

In a MDP, an \textit{agent} is in a state $s_t \in \mathcal{S}$ in the environment and interacts with this environment by performing actions $a_t \in \mathcal{A}$. The agent's actions are defined by a policy $\pi_\theta : \mathcal{S} \times \mathcal{A} \to \mathbb{R}_+$ parametrized by $\theta$. The objective of the agent is to learn an optimal policy: $\theta^* := \text{argmax}_\theta J(\pi_\theta)$, where $J(\pi_\theta) = \mathbb{E}_{p\sim \pi_\theta} \left[ \sum_{i=0}^T \gamma^i r_i\right]$ is the expected discounted return.

\subsection{Hierarchical reinforcement learning} \label{sec:background-hrl}
In HRL, the policy is decomposed into a high-level policy that optimizes the environment task reward, and a low-level policy that is conditioned on latent goals from the high-level and executes actions within the environment. The high-level controller is decoupled from the true MDP by operating at a lower temporal resolution and passing goals~\citep{dayan1993feudal} or options~\citep{sutton1999between} to the lower-level. This can reduce the credit assignment problem from the perspective of the high-level controller, and allows the low-level policy to produce action primitives that support short time horizon tasks as well~\citep{sutton1999between}.

\begin{figure*}
\centering
\begin{tikzpicture}
    \newcommand \boxwidth {0.65}
    \newcommand \boxoffset {0.5}
    \newcommand \envwidth {7.75}
    \newcommand \envheight {0.65}
    \newcommand \layerskip {1.4}

    \fill [orange!30!white] (0,0) rectangle (\envwidth, \envheight);
    \draw [brown] (0,0) -- (\envwidth,0);
    \draw [brown] (0, \envheight) -- (\envwidth, \envheight);
    \draw (\envwidth/2, \envheight/2) node {\small Environment};

    \filldraw [black!40!white, draw=black!90!white] 
    (2*\boxoffset, 3.5*\envheight) rectangle 
    (2*\boxoffset + \boxwidth, 3.5*\envheight + \boxwidth);
    \draw
    (2.0*\boxoffset + 0.5*\boxwidth, 3.5*\envheight + 0.5*\boxwidth) node {\small $\pi_m$};

    \filldraw [black!20!white, draw=black!80!white] 
    (4.5*\boxoffset, 2*\envheight) rectangle 
    (4.5*\boxoffset + \boxwidth, 2*\envheight + \boxwidth);
    \draw
    (4.5*\boxoffset + 0.5*\boxwidth, 2*\envheight + 0.5*\boxwidth) node {\small $\pi_w$};

    \filldraw [black!20!white, draw=black!80!white] 
    (6.0*\boxoffset + 1.0*\boxwidth, 2*\envheight) rectangle 
    (6.0*\boxoffset + 2.0*\boxwidth, 2*\envheight + \boxwidth);
    \draw
    (6.0*\boxoffset + 1.5*\boxwidth, 2*\envheight + 0.5*\boxwidth) node {\small $\pi_w$};

    \filldraw [black!20!white, draw=black!80!white] 
    (9.0*\boxoffset + 2.0*\boxwidth, 2*\envheight) rectangle 
    (9.0*\boxoffset + 3.0*\boxwidth, 2*\envheight + \boxwidth);
    \draw
    (9.0*\boxoffset + 2.5*\boxwidth, 2*\envheight + 0.5*\boxwidth) node {\small $\pi_w$};


    \draw [->] 
    (1.25 * \boxoffset, \envheight) -- (1.25 * \boxoffset, 4 * \envheight) -- (2*\boxoffset, 4 * \envheight);
    \draw [->] 
    (1.25 * \boxoffset, 2.5*\envheight) -- (4.5*\boxoffset, 2.5*\envheight);
    \draw 
    (1.25 * \boxoffset + 0.25, \envheight + 0.2) node {\small $s_t$}; 

    \draw [->] 
    (2*\boxoffset + \boxwidth, 4*\envheight) -- 
    (4.5*\boxoffset + 0.5*\boxwidth, 4*\envheight) -- 
    (4.5*\boxoffset + 0.5*\boxwidth, 3*\envheight);
    \draw [->] 
    (4.5*\boxoffset + 0.5*\boxwidth, 4*\envheight) -- 
    (4.5*\boxoffset + 2.5*\boxwidth, 4*\envheight) -- 
    (4.5*\boxoffset + 2.5*\boxwidth, 3*\envheight);
    \draw 
    (4.5*\boxoffset + 2.5*\boxwidth, 4*\envheight) -- 
    (5.5*\boxoffset + 2.5*\boxwidth, 4*\envheight); 
    \draw 
    (7.5*\boxoffset + 2.0*\boxwidth, 4.0*\envheight) node {$\cdots$}; 
    \draw [->]
    (8.5*\boxoffset + 2.0*\boxwidth, 4*\envheight) -- 
    (9.0*\boxoffset + 2.5*\boxwidth, 4*\envheight) --
    (9.0*\boxoffset + 2.5*\boxwidth, 3*\envheight); 
    \draw 
    (2*\boxoffset + \boxwidth + 0.3, 4 * \envheight + 0.2) node {\small $g_t$};

    \draw [->] 
    (4.5*\boxoffset + \boxwidth, 2.5*\envheight) --
    (5.0*\boxoffset + \boxwidth, 2.5*\envheight) --
    (5.0*\boxoffset + \boxwidth, 1.0*\envheight);
    \draw
    (5.0*\boxoffset + \boxwidth - 0.25, \envheight + 0.2) node {\small $a_t$}; 

    \draw [->] 
    (5.5*\boxoffset + \boxwidth, 1.0*\envheight) --
    (5.5*\boxoffset + \boxwidth, 2.5*\envheight) --
    (6.0*\boxoffset + \boxwidth, 2.5*\envheight);
    \draw
    (5.5*\boxoffset + \boxwidth + 0.4, \envheight + 0.2) node {\small $s_{t+1}$}; 

    \draw [->] 
    (6.0*\boxoffset + 2.0*\boxwidth, 2.5*\envheight) --
    (6.5*\boxoffset + 2.0*\boxwidth, 2.5*\envheight) --
    (6.5*\boxoffset + 2.0*\boxwidth, 1.0*\envheight);
    \draw
    (6.5*\boxoffset + 2.0*\boxwidth + 0.40, \envheight + 0.2) node {\small $a_{t+1}$}; 

    \draw (7.5*\boxoffset + 2.0*\boxwidth, 2.5*\envheight) node {$\cdots$}; 

    \draw [->] 
    (8.5*\boxoffset + 2.0*\boxwidth, 1.0*\envheight) --
    (8.5*\boxoffset + 2.0*\boxwidth, 2.5*\envheight) --
    (9.0*\boxoffset + 2.0*\boxwidth, 2.5*\envheight);
    \draw
    (8.5*\boxoffset + 2.0*\boxwidth + 0.40, \envheight + 0.2) node {\small $s_{t+k}$}; 

    \draw [->] 
    (9.0*\boxoffset + 3.0*\boxwidth, 2.5*\envheight) --
    (9.5*\boxoffset + 3.0*\boxwidth, 2.5*\envheight) --
    (9.5*\boxoffset + 3.0*\boxwidth, 1.0*\envheight);
    \draw
    (9.5*\boxoffset + 3.0*\boxwidth + 0.40, \envheight + 0.2) node {\small $a_{t+k}$};


    \newcommand \lanewidth {0.5}
    \newcommand \lanepos {5.5*\envheight}

    \filldraw [black!20!white] 
    (0, \lanepos) rectangle (\textwidth, \lanepos + \lanewidth);
    \draw (0, \lanepos) -- (\textwidth, \lanepos);
    \draw (0, \lanepos + \lanewidth) -- (\textwidth, \lanepos + \lanewidth);


    \node[inner sep=0pt] at 
    (1.5, \lanepos + 0.5 * \lanewidth)
    {\includegraphics[origin=c,height=0.5cm]{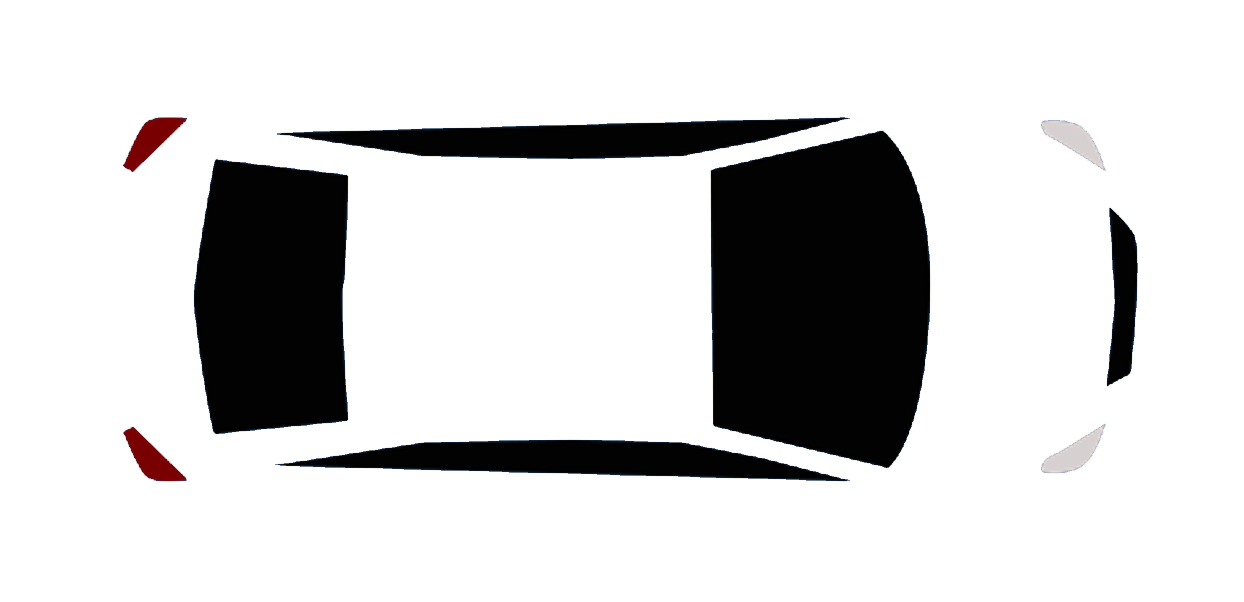}};

    \node[inner sep=0pt] at 
    (4.0, \lanepos + 0.5 * \lanewidth)
    {\includegraphics[origin=c,height=0.5cm]{figures/car.png}};

    \node[inner sep=0pt] at 
    (6.0, \lanepos + 0.5 * \lanewidth)
    {\includegraphics[origin=c,height=0.5cm]{figures/car.png}};

    \node[inner sep=0pt] at 
    (8.0, \lanepos + 0.5 * \lanewidth)
    {\includegraphics[origin=c,height=0.5cm]{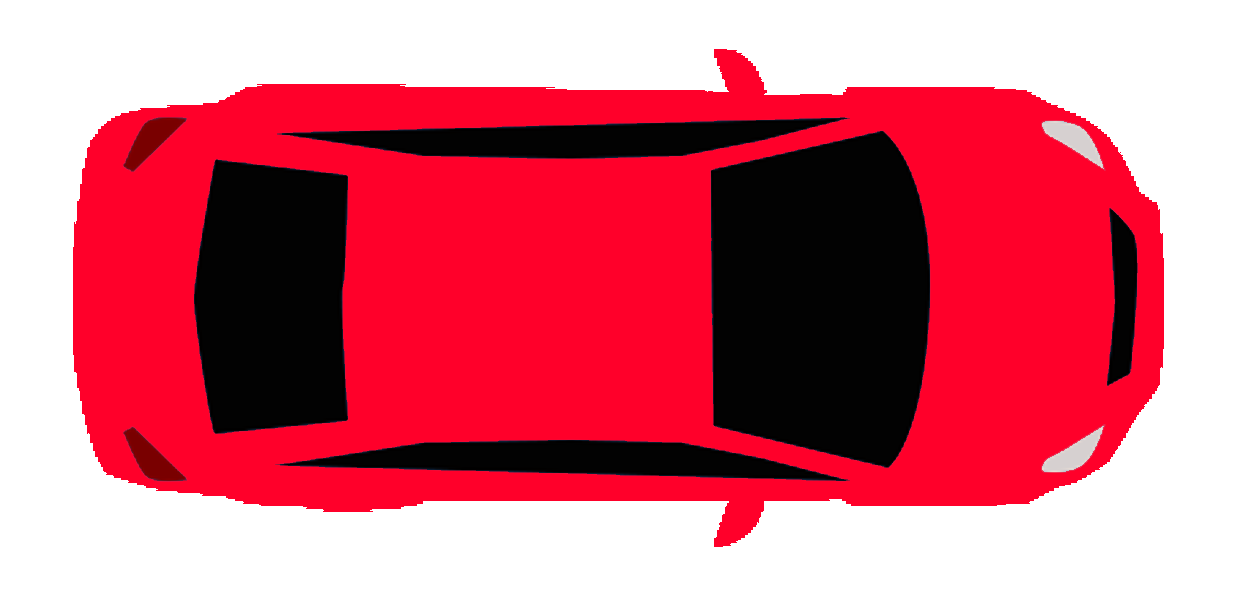}};

    \node[inner sep=0pt] at 
    (10.5, \lanepos + 0.5 * \lanewidth)
    {\includegraphics[origin=c,height=0.5cm]{figures/car.png}};

    \node[inner sep=0pt] at 
    (14.5, \lanepos + 0.5 * \lanewidth)
    {\includegraphics[origin=c,height=0.5cm]{figures/car.png}};
    
    
    \newcommand \speedometerCenterX {9.5}
    \newcommand \speedometerCenterY {1.25}
    \newcommand \speedometerRadius  {1.25}

    \newcommand \cosOne        {1}                    
    \newcommand \sinOne        {0}                    
    \newcommand \cosTwo        {0.9876883405951378}   
    \newcommand \sinTwo        {0.1564344650402308}   
    \newcommand \cosThree      {0.9510565162951535}   
    \newcommand \sinThree      {0.3090169943749474}   
    \newcommand \cosFour       {0.8910065241883679}   
    \newcommand \sinFour       {0.4539904997395467}   
    \newcommand \cosFive       {0.8090169943749475}   
    \newcommand \sinFive       {0.5877852522924731}   
    \newcommand \cosSix        {0.7071067811865476}   
    \newcommand \sinSix        {0.7071067811865476}   
    \newcommand \cosSeven      {0.5877852522924731}   
    \newcommand \sinSeven      {0.8090169943749475}   
    \newcommand \cosEight      {0.4539904997395467}   
    \newcommand \sinEight      {0.8910065241883679}   
    \newcommand \cosNine       {0.3090169943749474}   
    \newcommand \sinNine       {0.9510565162951535}   
    \newcommand \cosTen        {0.1564344650402308}   
    \newcommand \sinTen        {0.9876883405951378}   
    \newcommand \cosEleven     {0}                    
    \newcommand \sinEleven     {1}                    
    \newcommand \cosTwelve     {-0.1564344650402308}  
    \newcommand \sinTwelve     {0.9876883405951378}   
    \newcommand \cosThirteen   {-0.3090169943749474}  
    \newcommand \sinThirteen   {0.9510565162951535}   
    \newcommand \cosFourteen   {-0.4539904997395467}  
    \newcommand \sinFourteen   {0.8910065241883679}   
    \newcommand \cosFifteen    {-0.5877852522924731}  
    \newcommand \sinFifteen    {0.8090169943749475}   
    \newcommand \cosSixteen    {-0.7071067811865476}  
    \newcommand \sinSixteen    {0.7071067811865476}   
    \newcommand \cosSeventeen  {-0.8090169943749475}  
    \newcommand \sinSeventeen  {0.5877852522924731}   
    \newcommand \cosEighteen   {-0.8910065241883679}  
    \newcommand \sinEighteen   {0.4539904997395467}   
    \newcommand \cosNineteen   {-0.9876883405951378}  
    \newcommand \sinNineteen   {0.1564344650402308}   
    \newcommand \cosTwenty     {-0.9510565162951535}  
    \newcommand \sinTwenty     {0.3090169943749474}   
    \newcommand \cosTwentyOne  {-1}                   
    \newcommand \sinTwentyOne  {0}                    

    \newcommand \cosSpeed      {0.20791169081775945}
    \newcommand \sinSpeed      {0.9781476007338056}
    \newcommand \cosGoal       {0.927183854566787}
    \newcommand \sinGoal       {0.374606593415912}

    \draw 
    (8.00, \lanepos - 0.4 * \lanewidth) --
    (8.75, \lanepos - 2.25 * \lanewidth);

    \draw [line width=0.35mm]
    (\speedometerCenterX + 0.75 * \speedometerRadius * \cosOne,
     \speedometerCenterY + 0.75 * \speedometerRadius * \sinOne) --
    (\speedometerCenterX + 1.00 * \speedometerRadius * \cosOne,
     \speedometerCenterY + 1.00 * \speedometerRadius * \sinOne);
    \draw [line width=0.25mm]
    (\speedometerCenterX + 0.90 * \speedometerRadius * \cosTwo,
     \speedometerCenterY + 0.90 * \speedometerRadius * \sinTwo) --
    (\speedometerCenterX + 1.00 * \speedometerRadius * \cosTwo,
     \speedometerCenterY + 1.00 * \speedometerRadius * \sinTwo);
    \draw [line width=0.25mm]
    (\speedometerCenterX + 0.90 * \speedometerRadius * \cosThree,
     \speedometerCenterY + 0.90 * \speedometerRadius * \sinThree) --
    (\speedometerCenterX + 1.00 * \speedometerRadius * \cosThree,
     \speedometerCenterY + 1.00 * \speedometerRadius * \sinThree);
    \draw [line width=0.25mm]
    (\speedometerCenterX + 0.90 * \speedometerRadius * \cosFour,
     \speedometerCenterY + 0.90 * \speedometerRadius * \sinFour) --
    (\speedometerCenterX + 1.00 * \speedometerRadius * \cosFour,
     \speedometerCenterY + 1.00 * \speedometerRadius * \sinFour);
    \draw [line width=0.25mm]
    (\speedometerCenterX + 0.90 * \speedometerRadius * \cosFive,
     \speedometerCenterY + 0.90 * \speedometerRadius * \sinFive) --
    (\speedometerCenterX + 1.00 * \speedometerRadius * \cosFive,
     \speedometerCenterY + 1.00 * \speedometerRadius * \sinFive);
    \draw [line width=0.35mm]
    (\speedometerCenterX + 0.75 * \speedometerRadius * \cosSix,
     \speedometerCenterY + 0.75 * \speedometerRadius * \sinSix) --
    (\speedometerCenterX + 1.00 * \speedometerRadius * \cosSix,
     \speedometerCenterY + 1.00 * \speedometerRadius * \sinSix);
    \draw [line width=0.25mm]
    (\speedometerCenterX + 0.90 * \speedometerRadius * \cosSeven,
     \speedometerCenterY + 0.90 * \speedometerRadius * \sinSeven) --
    (\speedometerCenterX + 1.00 * \speedometerRadius * \cosSeven,
     \speedometerCenterY + 1.00 * \speedometerRadius * \sinSeven);
    \draw [line width=0.25mm]
    (\speedometerCenterX + 0.90 * \speedometerRadius * \cosEight,
     \speedometerCenterY + 0.90 * \speedometerRadius * \sinEight) --
    (\speedometerCenterX + 1.00 * \speedometerRadius * \cosEight,
     \speedometerCenterY + 1.00 * \speedometerRadius * \sinEight);
    \draw [line width=0.25mm]
    (\speedometerCenterX + 0.90 * \speedometerRadius * \cosNine,
     \speedometerCenterY + 0.90 * \speedometerRadius * \sinNine) --
    (\speedometerCenterX + 1.00 * \speedometerRadius * \cosNine,
     \speedometerCenterY + 1.00 * \speedometerRadius * \sinNine);
    \draw [line width=0.25mm]
    (\speedometerCenterX + 0.90 * \speedometerRadius * \cosTen,
     \speedometerCenterY + 0.90 * \speedometerRadius * \sinTen) --
    (\speedometerCenterX + 1.00 * \speedometerRadius * \cosTen,
     \speedometerCenterY + 1.00 * \speedometerRadius * \sinTen);
    \draw [line width=0.35mm]
    (\speedometerCenterX + 0.75 * \speedometerRadius * \cosEleven,
     \speedometerCenterY + 0.75 * \speedometerRadius * \sinEleven) --
    (\speedometerCenterX + 1.00 * \speedometerRadius * \cosEleven,
     \speedometerCenterY + 1.00 * \speedometerRadius * \sinEleven);
    \draw [line width=0.25mm]
    (\speedometerCenterX + 0.90 * \speedometerRadius * \cosTwelve,
     \speedometerCenterY + 0.90 * \speedometerRadius * \sinTwelve) --
    (\speedometerCenterX + 1.00 * \speedometerRadius * \cosTwelve,
     \speedometerCenterY + 1.00 * \speedometerRadius * \sinTwelve);
    \draw [line width=0.25mm]
    (\speedometerCenterX + 0.90 * \speedometerRadius * \cosThirteen,
     \speedometerCenterY + 0.90 * \speedometerRadius * \sinThirteen) --
    (\speedometerCenterX + 1.00 * \speedometerRadius * \cosThirteen,
     \speedometerCenterY + 1.00 * \speedometerRadius * \sinThirteen);
    \draw [line width=0.25mm]
    (\speedometerCenterX + 0.90 * \speedometerRadius * \cosFourteen,
     \speedometerCenterY + 0.90 * \speedometerRadius * \sinFourteen) --
    (\speedometerCenterX + 1.00 * \speedometerRadius * \cosFourteen,
     \speedometerCenterY + 1.00 * \speedometerRadius * \sinFourteen);
    \draw [line width=0.25mm]
    (\speedometerCenterX + 0.90 * \speedometerRadius * \cosFifteen,
     \speedometerCenterY + 0.90 * \speedometerRadius * \sinFifteen) --
    (\speedometerCenterX + 1.00 * \speedometerRadius * \cosFifteen,
     \speedometerCenterY + 1.00 * \speedometerRadius * \sinFifteen);
    \draw [line width=0.35mm]
    (\speedometerCenterX + 0.75 * \speedometerRadius * \cosSixteen,
     \speedometerCenterY + 0.75 * \speedometerRadius * \sinSixteen) --
    (\speedometerCenterX + 1.00 * \speedometerRadius * \cosSixteen,
     \speedometerCenterY + 1.00 * \speedometerRadius * \sinSixteen);
    \draw [line width=0.25mm]
    (\speedometerCenterX + 0.90 * \speedometerRadius * \cosSeventeen,
     \speedometerCenterY + 0.90 * \speedometerRadius * \sinSeventeen) --
    (\speedometerCenterX + 1.00 * \speedometerRadius * \cosSeventeen,
     \speedometerCenterY + 1.00 * \speedometerRadius * \sinSeventeen);
    \draw [line width=0.25mm]
    (\speedometerCenterX + 0.90 * \speedometerRadius * \cosEighteen,
     \speedometerCenterY + 0.90 * \speedometerRadius * \sinEighteen) --
    (\speedometerCenterX + 1.00 * \speedometerRadius * \cosEighteen,
     \speedometerCenterY + 1.00 * \speedometerRadius * \sinEighteen);
    \draw [line width=0.25mm]
    (\speedometerCenterX + 0.90 * \speedometerRadius * \cosNineteen,
     \speedometerCenterY + 0.90 * \speedometerRadius * \sinNineteen) --
    (\speedometerCenterX + 1.00 * \speedometerRadius * \cosNineteen,
     \speedometerCenterY + 1.00 * \speedometerRadius * \sinNineteen);
    \draw [line width=0.25mm]
    (\speedometerCenterX + 0.90 * \speedometerRadius * \cosTwenty,
     \speedometerCenterY + 0.90 * \speedometerRadius * \sinTwenty) --
    (\speedometerCenterX + 1.00 * \speedometerRadius * \cosTwenty,
     \speedometerCenterY + 1.00 * \speedometerRadius * \sinTwenty);
    \draw [line width=0.35mm]
    (\speedometerCenterX + 0.75 * \speedometerRadius * \cosTwentyOne,
     \speedometerCenterY + 0.75 * \speedometerRadius * \sinTwentyOne) --
    (\speedometerCenterX + 1.00 * \speedometerRadius * \cosTwentyOne,
     \speedometerCenterY + 1.00 * \speedometerRadius * \sinTwentyOne);

    \draw [line width=0.5mm, color=blue]
    (\speedometerCenterX + 0.0 * \speedometerRadius * \cosGoal,
     \speedometerCenterY + 0.0 * \speedometerRadius * \sinGoal) --
    (\speedometerCenterX + 1.0 * \speedometerRadius * \cosGoal,
     \speedometerCenterY + 1.0 * \speedometerRadius * \sinGoal);
    \draw 
    (\speedometerCenterX + 1.15 * \speedometerRadius * \cosGoal + 0.3,
     \speedometerCenterY + 1.15 * \speedometerRadius * \sinGoal) node {\small \textcolor{blue}{goal}}; 

    \draw [line width=0.5mm, color=red]
    (\speedometerCenterX + 0.0 * \speedometerRadius * \cosSpeed,
     \speedometerCenterY + 0.0 * \speedometerRadius * \sinSpeed) --
    (\speedometerCenterX + 1.0 * \speedometerRadius * \cosSpeed,
     \speedometerCenterY + 1.0 * \speedometerRadius * \sinSpeed);
    \draw 
    (\speedometerCenterX + 1.15 * \speedometerRadius * \cosSpeed + 1,
     \speedometerCenterY + 1.15 * \speedometerRadius * \sinSpeed) node {\small \textcolor{red}{current speed}}; 

    \filldraw (\speedometerCenterX,\speedometerCenterY) circle (0.05 * \speedometerRadius);

    
    \draw [dashed] 
    (0.788 * \textwidth, 0.0) -- 
    (0.788 * \textwidth, 3.2); 
    \draw [dashed] 
    (0.788 * \textwidth, 3.2) -- 
    (\textwidth, 3.2); 

    \hspace*{0.8\textwidth}\includegraphics[width=0.2\textwidth]{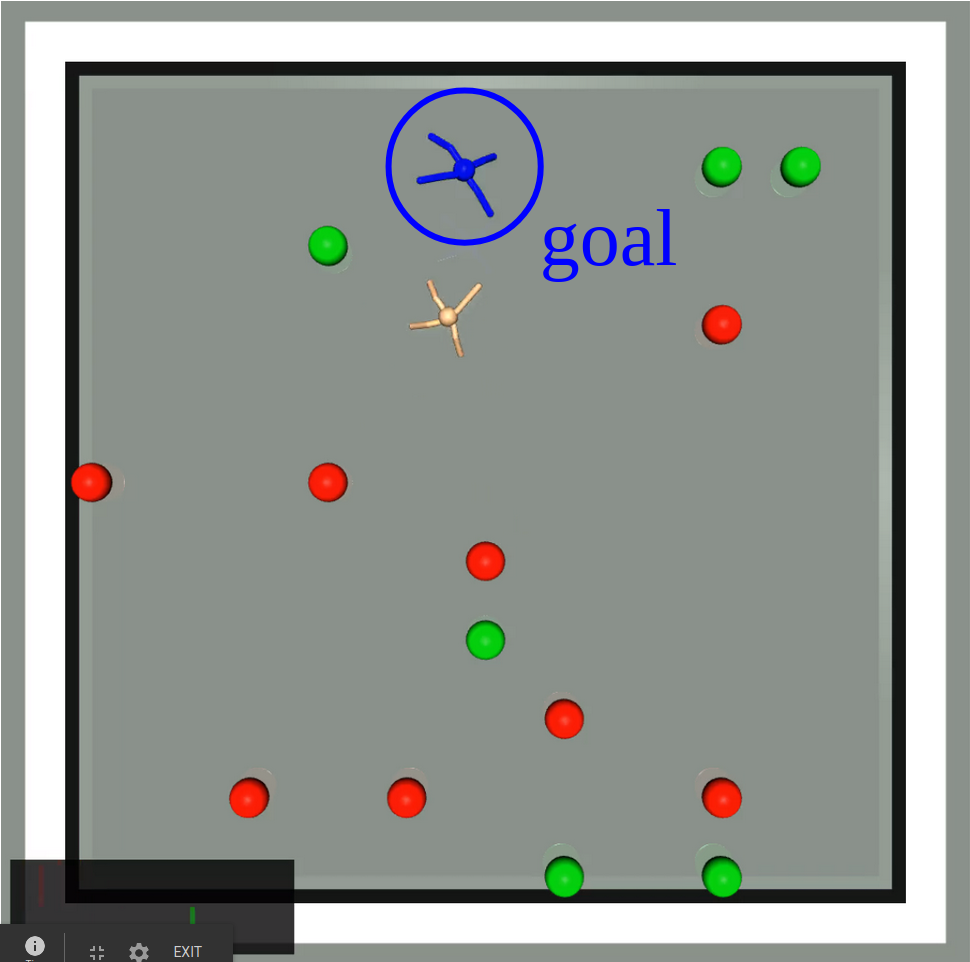}

\end{tikzpicture}
\caption{An illustration of the studied hierarchical model. \textbf{Left:} A manager network $\pi_m$ issues commands (or goals) $g_t$ over $k$ consecutive time steps to a worker $\pi_w$. The worker then performs environment actions $a_t$ to accomplish these goals. \textbf{Top/right:} The commands issued by the manager denote desired states for the worker to traverse. For the AV control tasks, we define the goal as the desired speeds for each AV. Moreover, for agent navigation tasks, the goals are defined as the desired position and joint angles of the agent.}
\label{fig:hrl-model}
\end{figure*}

Several HRL frameworks have been proposed to facilitate and/or encourage the decomposition of decision-making and execution during training~\citep{dayan1993feudal, sutton1999between, parr1998reinforcement, dietterich2000hierarchical}. In this article, we consider a two-level goal-conditioned arrangement~\citep{2017-TOG-deepLoco, vezhnevets2017feudal, nachum2018data, nasiriany2019planning} (see Figure~\ref{fig:hrl-model}). This network consists of a high-level, or manager, policy $\pi_m$ that computes and outputs goals $g_t \sim \pi_m(s_t)$ every $k$ time steps, and a low-level, or worker, policy $\pi_w$ that takes as inputs the current state and the assigned goals and is encouraged to perform actions $a_t \sim \pi_w(s_t, g_t)$ that satisfy these goals via an intrinsic reward function $r_w(s_t, g_t, s_{t+1})$ (see Appendix~\ref{sec:intrinsic-reward}).

\subsection{Hierarchical policy optimization} \label{sec:no-cooperation}

We consider a concurrent training procedure for the manager and worker policies. In this setting, the manager receives a reward based on the original environmental reward function: $r_m(s_t)$. The objective function from the perspective of the manager is then:\\[-3pt]
\begin{equation} \label{eq:manager-objective}
\resizebox{!}{12pt}{$
    J_m = \mathbb{E}_{s\sim p_\pi} \left[ \sum_{t=0}^{T/k} \left[ \gamma^t r_m(s_t) \right] \right]
$}
\end{equation}
Conversely, the worker policy is motivated to follow the goals set by the manager via an intrinsic reward $r_w (s_t, g_t, s_{t+1})$ separate from the external reward. The objective function from the perspective of the worker is then:\\[-6pt]
\begin{equation} \label{eq:worker-objective}
\resizebox{!}{12pt}{$
    J_w = \mathbb{E}_{s\sim p_\pi} \left[ \sum_{t=0}^k \gamma^t r_w(s_t, g_t,\pi_w(s_t,g_t)) \right]
$}
\end{equation}
Notably, no gradients are shared between the manager and worker policies. As discussed in Section~\ref{sec:introduction} and depicted in Figure~\ref{fig:goaldists}, the absence of such feedback often results in the formation of non-cooperative goal-assignment behaviors that strain the learning process.


\section{Cooperative hierarchical reinforcement learning} \label{sec:method}

\methodName promotes cooperation by propagating losses that arise from random or unachievable goal-assignment strategies. As part of this algorithm, we also present a mechanism to optimize the level of cooperation and ground the notion of cooperation in HRL to measurable variables. 

\subsection{Promoting cooperation via loss-sharing} \label{sec:algorithm}

\citet{tampuu2017multiagent} explored the effects of reward (or loss) sharing between agents on the emergence of cooperative and competitive in multiagent two-player games. Their study highlights two potential benefits of loss-sharing in multiagent systems: 1) emerged cooperative behaviors are less aggressive and more likely to emphasize improved interactions with neighboring agents, 2) these interactions reduce overestimation bias of the Q-function from the perspective of each agent. We develop a method to gain similar benefits by using loss-sharing paradigms in goal-conditioned hierarchies. In particular, we focus on the emergence of collaborative behaviors from the perspective of goal-assignment and goal-achieving policies In line with prior work, we promote the emergence of cooperative behaviors by incorporating a weighted form of the worker's expected return to the original manager objective $J_m$. The manager's new expected return is:
\begin{equation} \label{eq:connected-return}
    J_m' = J_m + \lambda J_w
\end{equation}
where $\lambda$ is a weighting term that controls the level of cooperation the manager has with the worker.

In practice, we find that this addition to the objective serves to promote cooperation by aligning goal-assignment actions by the manager with achievable trajectories by the goal-achieving worker. This serves as a soft constraint to the manager policy: when presented with goals that perform similarly, the manager tends towards goals that more closely match the worker states. The degree to which the policy tends toward achievable states is dictated by the $\lambda$ term. The choice of this parameter accordingly can have a significant effect on learned goals. For large or infinite values of $\lambda$, for instance, this cooperative term can eliminate exploration by assigning goals that match the worker's current state and prevent forward movement (this is highlighted in Section~\ref{sec:cg-weight}). In Section~\ref{sec:method-constrained-hrl} we detail how we mitigate this issue by dynamically controlling the level of cooperation online.

\subsection{Cooperative gradients via differentiable communication}

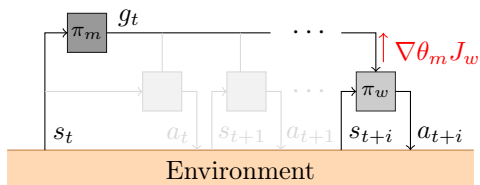
\begin{wrapfigure}{R}{0.42\textwidth}
\centering
\begin{minipage}[b]{0.95\textwidth}
\begin{tikzpicture}
    \newcommand \boxwidth {0.52}
    \newcommand \boxoffset {0.4}
    \newcommand \envwidth {6.2}
    \newcommand \envheight {0.52}
    \newcommand \layerskip {1.12}

    \fill [orange!30!white] (0,0) rectangle (\envwidth, \envheight);
    \draw [brown] (0,0) -- (\envwidth,0);
    \draw [brown] (0, \envheight) -- (\envwidth, \envheight);
    \draw (\envwidth/2, \envheight/2) node {\small Environment};

    \filldraw [black!40!white, draw=black!90!white] 
    (2*\boxoffset, 3.5*\envheight) rectangle 
    (2*\boxoffset + \boxwidth, 3.5*\envheight + \boxwidth);
    \draw 
    (2.0*\boxoffset + 0.5*\boxwidth, 3.5*\envheight + 0.5*\boxwidth) node {\scriptsize $\pi_m$};

    \filldraw [black!5!white, draw=black!15!white] 
    (4.5*\boxoffset, 2*\envheight) rectangle 
    (4.5*\boxoffset + \boxwidth, 2*\envheight + \boxwidth);

    \filldraw [black!5!white, draw=black!15!white] 
    (6.0*\boxoffset + 1.0*\boxwidth, 2*\envheight) rectangle 
    (6.0*\boxoffset + 2.0*\boxwidth, 2*\envheight + \boxwidth);

    \filldraw [black!20!white, draw=black!80!white] 
    (9.0*\boxoffset + 2.0*\boxwidth, 2*\envheight) rectangle 
    (9.0*\boxoffset + 3.0*\boxwidth, 2*\envheight + \boxwidth);
    \draw
    (9.0*\boxoffset + 2.5*\boxwidth, 2*\envheight + 0.5*\boxwidth) node {\scriptsize $\pi_w$};


    \draw [red] [->] 
    (8.6*\boxoffset + 3.0*\boxwidth, 3.25*\envheight) --
    (8.6*\boxoffset + 3.0*\boxwidth, 4.00*\envheight);
    \draw 
    (13*\boxoffset + \boxwidth, 3.5*\envheight) node {\small \color{red}{$\nabla \theta_m J_w$}}; 


    \draw [->] 
    (1.25 * \boxoffset, \envheight) -- (1.25 * \boxoffset, 4 * \envheight) -- (2*\boxoffset, 4 * \envheight);
    \draw [black!15!white] [->] 
    (1.25 * \boxoffset, 2.5*\envheight) -- (4.5*\boxoffset, 2.5*\envheight);
    \draw 
    (1.25 * \boxoffset + 0.25, \envheight + 0.2) node {\small $s_t$}; 

    \draw
    (2*\boxoffset + \boxwidth, 4*\envheight) -- 
    (5.5*\boxoffset + 2.5*\boxwidth, 4*\envheight); 
    \draw [black!15!white]
    (4.5*\boxoffset + 0.5*\boxwidth, 4*\envheight) -- 
    (4.5*\boxoffset + 0.5*\boxwidth, 3*\envheight);
    \draw [black!15!white]
    (4.5*\boxoffset + 2.5*\boxwidth, 4*\envheight) -- 
    (4.5*\boxoffset + 2.5*\boxwidth, 3*\envheight);
    \draw 
    (7.5*\boxoffset + 2.0*\boxwidth, 4.0*\envheight) node {$\cdots$}; 
    \draw [->]
    (8.5*\boxoffset + 2.0*\boxwidth, 4*\envheight) -- 
    (9.0*\boxoffset + 2.5*\boxwidth, 4*\envheight) --
    (9.0*\boxoffset + 2.5*\boxwidth, 3*\envheight); 
    \draw 
    (2*\boxoffset + \boxwidth + 0.3, 4 * \envheight + 0.2) node {\small $g_t$}; 

    \draw [black!15!white] [->] 
    (4.5*\boxoffset + \boxwidth, 2.5*\envheight) --
    (5.0*\boxoffset + \boxwidth, 2.5*\envheight) --
    (5.0*\boxoffset + \boxwidth, 1.0*\envheight);
    \draw
    (5.0*\boxoffset + \boxwidth - 0.25, \envheight + 0.2) node {\small \color{black!15!white}{$a_t$}}; 

    \draw [black!15!white] [->] 
    (5.5*\boxoffset + \boxwidth, 1.0*\envheight) --
    (5.5*\boxoffset + \boxwidth, 2.5*\envheight) --
    (6.0*\boxoffset + \boxwidth, 2.5*\envheight);
    \draw
    (5.5*\boxoffset + \boxwidth + 0.4, \envheight + 0.2) node {\small \color{black!15!white}{$s_{t+1}$}}; 

    \draw [black!15!white] [->] 
    (6.0*\boxoffset + 2.0*\boxwidth, 2.5*\envheight) --
    (6.5*\boxoffset + 2.0*\boxwidth, 2.5*\envheight) --
    (6.5*\boxoffset + 2.0*\boxwidth, 1.0*\envheight);
    \draw
    (6.5*\boxoffset + 2.0*\boxwidth + 0.40, \envheight + 0.2) node {\small \color{black!15!white}{$a_{t+1}$}}; 

    \draw (7.5*\boxoffset + 2.0*\boxwidth, 2.5*\envheight) node {\color{black!15!white}{$\cdots$}}; 

    \draw [->] 
    (8.5*\boxoffset + 2.0*\boxwidth, 1.0*\envheight) --
    (8.5*\boxoffset + 2.0*\boxwidth, 2.5*\envheight) --
    (9.0*\boxoffset + 2.0*\boxwidth, 2.5*\envheight);
    \draw
    (8.5*\boxoffset + 2.0*\boxwidth + 0.40, \envheight + 0.2) node {\small $s_{t+i}$}; 

    \draw [->] 
    (9.0*\boxoffset + 3.0*\boxwidth, 2.5*\envheight) --
    (9.5*\boxoffset + 3.0*\boxwidth, 2.5*\envheight) --
    (9.5*\boxoffset + 3.0*\boxwidth, 1.0*\envheight);
    \draw
    (9.5*\boxoffset + 3.0*\boxwidth + 0.40, \envheight + 0.2) node {\small $a_{t+i}$};
\end{tikzpicture}
\end{minipage}
\caption{An illustration of the policy gradient procedure. To encourage cooperation, we introduce a gradient that propagates the losses of the worker policies through the manager.
}
\label{fig:hrl-connected}
\end{wrapfigure}

To compute the gradient of the additional weighted expected return through the parameters of the manager policy, we take inspiration from similar studies in differentiable communication in MARL. The main insight that enables the derivation of such a gradient is the notion that goal states $g_t$ from the perspective of the worker policy are structurally similar to communication signals in multi-agent systems. As a result, the gradients of the expected intrinsic returns can be computed by replacing the goal term within the reward function of the worker with the direct output from the manager's policy. This is depicted in Figure~\ref{fig:hrl-connected}. The updated gradient is defined in Theorem~\ref{theorem} below.

\begin{theorem} \label{theorem}
Define the goal $g_t$ provided to the input of the worker policy $\pi_w(s_t,g_t)$ as the direct output from the manager policy $g_t$ whose transition function is:
\begin{equation}
    g_t(\theta_m) = 
    \begin{cases}
        \pi_m(s_t) & \text{if } t \text{ mod } k = 0 \\
        h(s_{t-1}, g_{t-1}(\theta_m), s_t) & \text{otherwise}
    \end{cases}
\end{equation}
where $h(\cdot)$ is a fixed goal transition function between meta-periods (see Appendix~\ref{sec:intrinsic-reward}). Under this assumption, the solution to the deterministic policy gradient~\citep{silver2014deterministic} of Eq.~\eqref{eq:connected-return} with respect to the manager's parameters $\theta_m$ is:
\begin{equation} \label{eq:connected-gradient}
    \resizebox{.9\hsize}{!}{$
    \begin{aligned}
        \nabla_{\theta_m} J_m' &= \mathbb{E}_{s\sim p_\pi} \big[ \nabla_a Q_m (s,a)|_{a=\pi_m(s)} \nabla_{\theta_m} \pi_m(s)\big] \\
        &\quad + \lambda \mathbb{E}_{s\sim p_\pi} \bigg[ \nabla_{\theta_m} g_t \nabla_g \big(r_w(s_t,g,\pi_w(s_t,g_t)) + \pi_w (s_t,g) \nabla_a Q_w(s_t,g_t,a)|_{a=\pi_w(s_t,g_t)}\vphantom{\int} \big) \bigg\rvert_{g=g_t} \bigg]
    \end{aligned}
    $}
\end{equation}
where $Q_m(s,a)$ and $Q_w(s,g,a)$ are approximations for the expected environmental and intrinsic returns, respectively.
\end{theorem}

\vspace{-6pt}

\textit{Proof.} See Appendix~\ref{sec:derivation}.

This new gradient consists of three terms. The first and third term computes the gradient of the critic policies $Q_m$ and $Q_w$ for the parameters $\theta_m$, respectively. The second term computes the gradient of the worker-specific reward for the parameters $\theta_m$. This reward is a design feature within the goal-conditioned RL formulation, however, any reward function for which the gradient can be explicitly computed can be used. We describe a practical algorithm for training a cooperative two-level hierarchy implementing this loss function in Algorithm~\ref{alg:training}.

\subsection{Cooperative HRL as constrained optimization}
\label{sec:method-constrained-hrl}

In the previous sections, we introduced a framework for inducing and studying the effects of cooperation between internal agents within a hierarchy. The degree of cooperation is defined through a hyperparameter ($\lambda$), and if properly defined can greatly improve training performance in certain environments. The choice of $\lambda$, however, can be difficult to specify without a priori knowledge of the task it is assigned to. We accordingly wish to ground the choice of $\lambda$ to measurable terms that can be reasoned and adjusted for. To that end, we observe that the cooperative $\lambda$ term acts equivalently as a Lagrangian term in constrained optimization~\citep{bertsekas2014constrained} with the expected return for the lower-level policy serving as the constraint. 
Our formulation of the HRL problem can similarly be framed as a constrained optimization problem, denoted as:
\begin{equation}
    \begin{aligned}
        &\max_{\pi_m} \left[ J_m + \min_{\lambda\geq 0} \left( \lambda \delta - \lambda \min_{\pi_w} J_w \right) \right]
    \end{aligned}
\end{equation}
where $\delta$ is the desired expected discounted \emph{intrinsic} returns. The derivation of this equation and practical implementations are provided in Appendix~\ref{sec:constrained-hrl}.

In practice, this updated form of the objective provides two meaningful benefits: 1) As discussed in Section~\ref{sec:cg-weight}, it introduces bounds for appropriate values of $\delta$ that can then be explored and tuned, and 2) for the more complex and previously unsolvable tasks, we find that this approach results in more stable learning and better performing policies.


\section{Related Work} \label{sec:related-work}

The topic explored in this article takes inspiration in part from studies of communication in multiagent reinforcement learning (MARL)~\citep{thomas2011conjugate, thomas2011policy, DBLP:journals/corr/SukhbaatarSF16, DBLP:journals/corr/FoersterAFW16a}. In MARL, communication channels are often shared among agents as a means of coordinating and influencing neighboring agents. Challenges emerge, however, as a result of the ambiguity of communication signals in the early staging of training, with agents forced to coordinate between sending and interpreting messages~\citep{mordatch2018emergence, eccles2019biases}. Similar communication channels are present in the HRL domain, with higher-level policies communicating one-sided signals in the form of goals to lower-level policies. The difficulties associated with cooperation, accordingly, likely (and as we find here in fact do) persist under this setting. The work presented here serves to make connections between these two fields, and will hopefully motivate future work on unifying the challenges experienced in each.

Our work is most motivated by the Differentiable Inter-Agent Learning (DIAL) method proposed by~\citet{DBLP:journals/corr/FoersterAFW16a}. Our work does not aim to learn an explicit communication channel; however, it is motivated by a similar principle - that letting gradients flow across agents results in richer feedback. Furthermore, we differ in the fact that we structure the problem as a hierarchical reinforcement learning problem in which agents are designed to solve dissimilar tasks. This disparity forces a more constrained and directed objective in which varying degrees of cooperation can be defined. This insight, we find, is important for ensuring that shared gradients can provide meaningful benefits to the hierarchical paradigm.

Another prominent challenge in MARL is the notion of \emph{non-stationarity}~\citep{busoniu2006multi, weinberg2004best, foerster2017stabilising}, whereby the continually changing nature of decision-making policies serve to destabilize training. This has been identified in previous studies on HRL, with techniques such as off-policy sample relabeling~\citep{nachum2018data} and hindsight~\citep{levy2017hierarchical} providing considerable improvements to training performance. Similarly, the method presented in this article focuses on non-stationary in HRL, with the manager constraining its search within the region of achievable goals by the worker. Unlike these methods, however, our approach additionally accounts for ambiguities in the credit assignment problem from the perspective of the manager. As we demonstrate in Section~\ref{sec:experiments}, this cooperation improves learning across a collection of complex and partially observable tasks with a high degree of stochasticity. These results suggest that multifaceted approaches to hierarchical learning, like the one proposed in this article, that account for features such as information-sharing and cooperation in addition to non-stationarity are necessary for stable and generalizable HRL algorithms.


\begin{figure*}
\centering
\begin{subfigure}[b]{\textwidth}
    \centering
    \includegraphics[width=\textwidth]{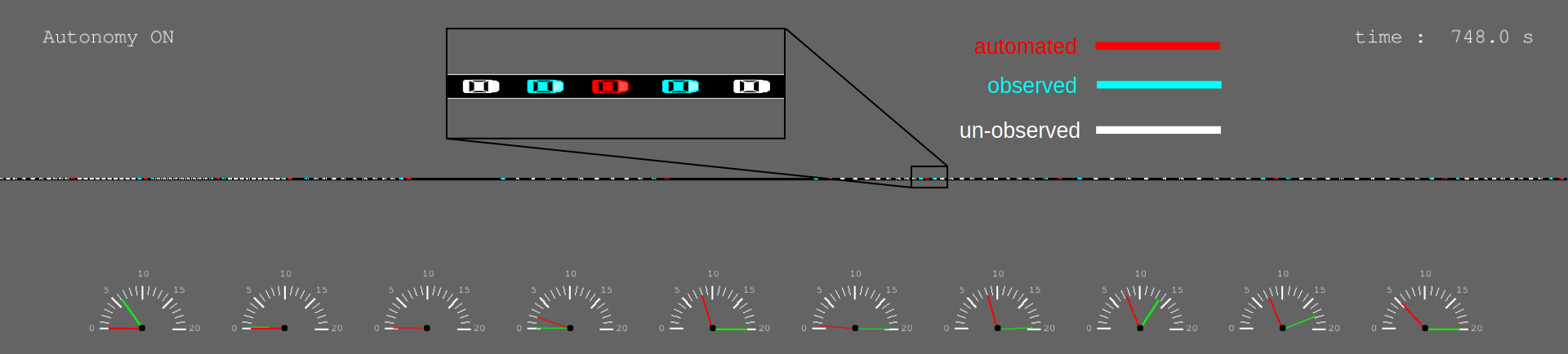}
    \caption{\footnotesize \highwaysingle}
    \label{fig:highwaysingle-env}
\end{subfigure} \\ \vspace{0.4cm}
\begin{subfigure}[b]{0.28\textwidth}
    \centering
    \includegraphics[height=2.7cm]{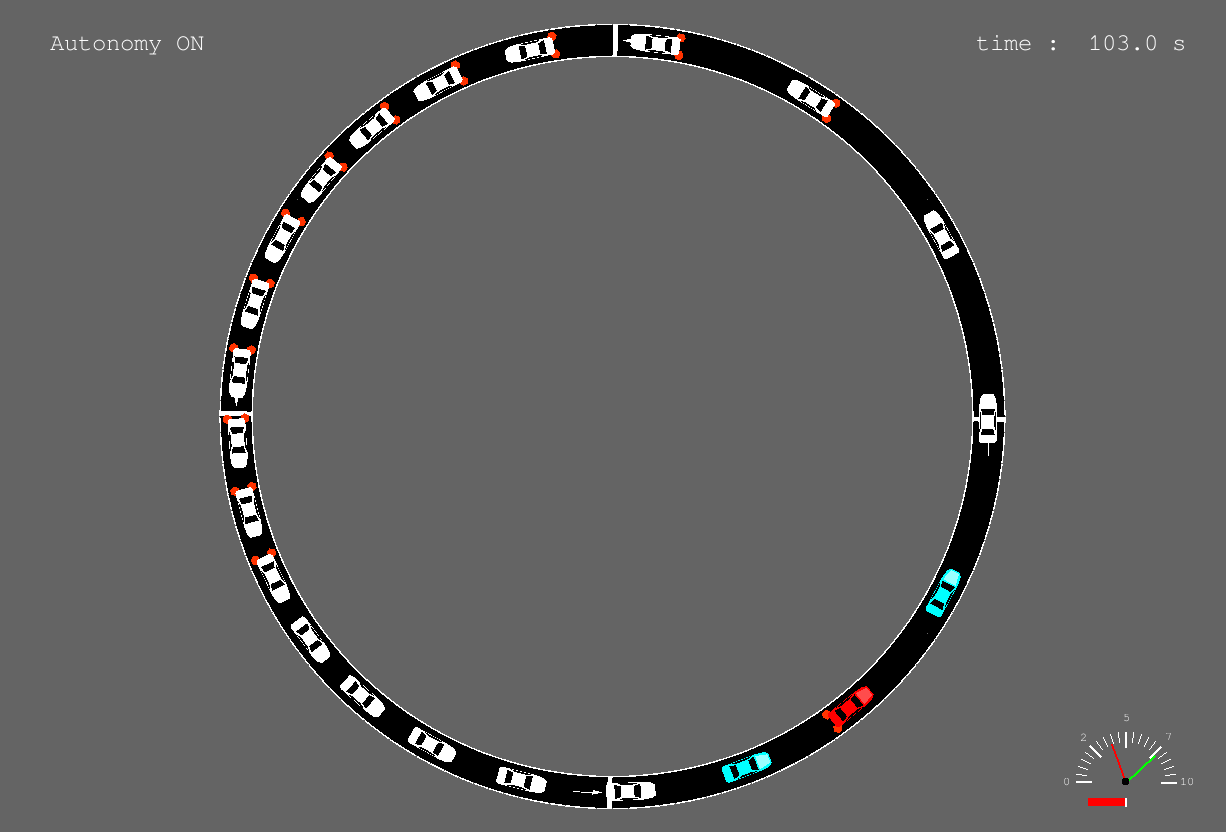}
    \caption{\label{fig:ring-env} \footnotesize \ringroad}
\end{subfigure}
\hfill
\begin{subfigure}[b]{0.2\textwidth}
    \centering
    \includegraphics[height=2.7cm]{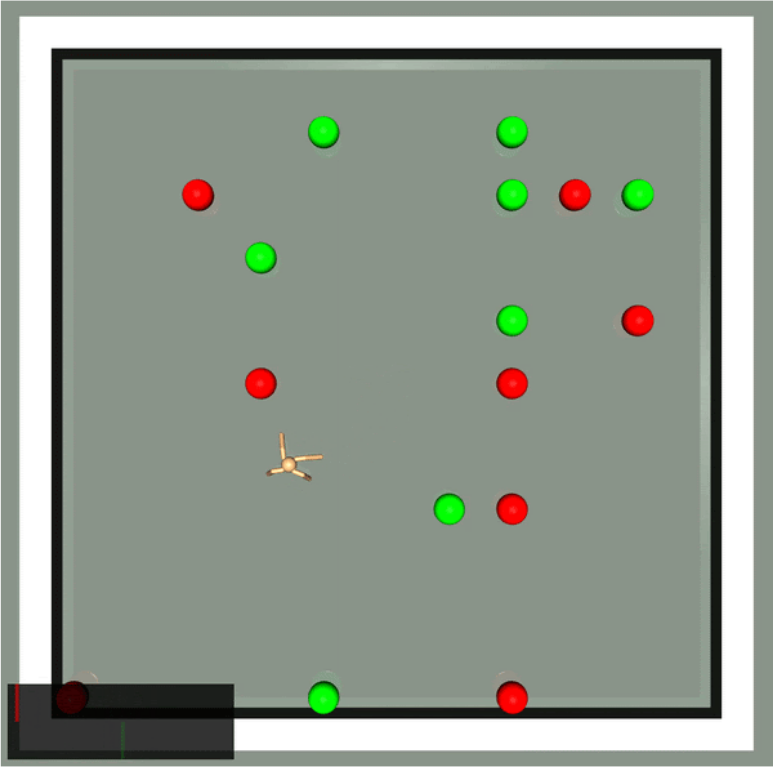}
    \caption{\label{fig:antgather-env} \footnotesize \antgather}
\end{subfigure}
\hfill
\begin{subfigure}[b]{0.2\textwidth}
    \centering
    \includegraphics[height=2.7cm]{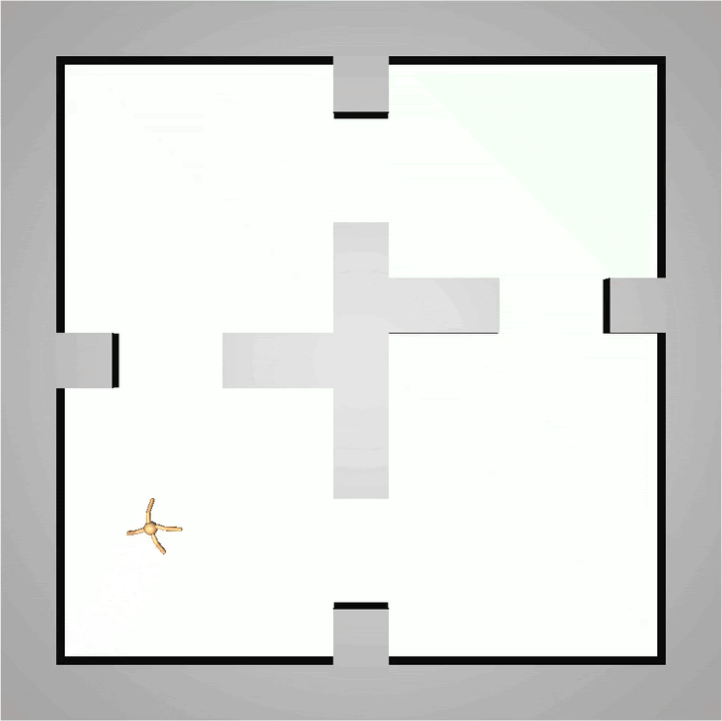}
    \caption{\label{fig:antfourrooms-env} \footnotesize \antfourrooms}
\end{subfigure}
\hfill
\begin{subfigure}[b]{0.2\textwidth}
    \centering
    \includegraphics[height=2.7cm]{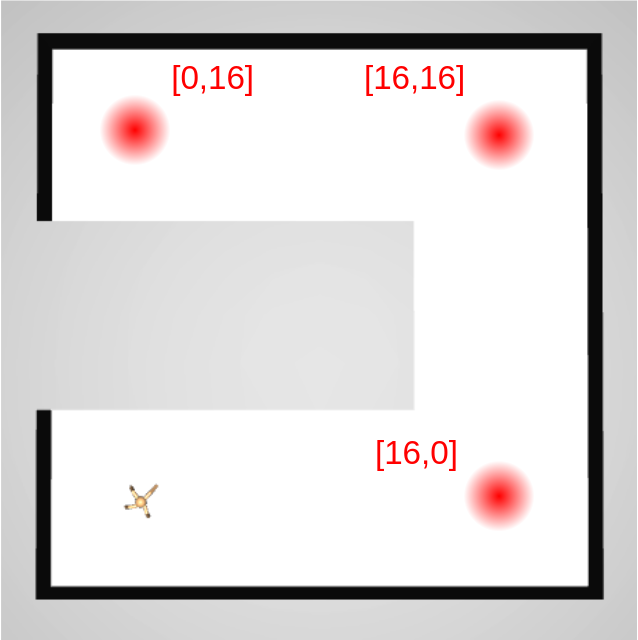}
    \caption{\label{fig:antmaze-sim} \footnotesize \antmaze}
\end{subfigure}
\caption{
Training environments explored within this article. We compare the performance of various HRL algorithms on two mixed-autonomy traffic control task (a,b) and three ant navigation tasks (c,d,e). A description of each of these environments is provided in Section~\ref{sec:environments}.
}
\label{fig:envs}
\end{figure*}

\section{Experiments}
\label{sec:experiments}

In this section, we detail the experimental setup and training procedure and present the performance of our method over various continuous control tasks. These experiments aim to analyze three aspects of HRL training: 
    (1) How does cooperation in HRL improve the development of goal-assignment strategies and the learning performance?
    (2) What impact does automatically varying the degree of cooperation between learned higher and lower level behaviors have?
    (3) Does the use of communication results in a more structured goal condition lower-level policy that transfers better to other tasks?\\[-20pt]

\subsection{Environments}

We explore the use of hierarchical reinforcement learning on a variety of difficult long time horizon tasks, see Figure~\ref{fig:envs}. These environments vary from agent navigation tasks (Figures~\ref{fig:antgather-env}),
in which a robotic agent tries to achieve certain long-term goals, to two mixed-autonomy traffic control tasks (Figures~\ref{fig:ring-env}~to~\ref{fig:highwaysingle-env}), in which a subset of vehicles (seen in red) are treated as automated vehicles and attempt to reduce oscillations in vehicle speeds known as stop-and-go traffic. Further details are available in Appendix~\ref{sec:environments}.

The environments presented here pose a difficulty to standard RL methods for a variety of reasons. We broadly group the most significant of these challenges into two categories: temporal reasoning and delayed feedback.

\noindent \textbf{Temporal reasoning.} \ \
In the agent navigation tasks, the agent must learn to perform multiple tasks at varying levels of granularity. At the level of individual timesteps, the agent must learn to navigate its surroundings, moving up, down, left, or right without falling or dying prematurely. More macroscopically, however, the agent must exploit these action primitives to achieve high-level goals that may be sparsely defined or require exploration across multiple timesteps. For even state-of-the-art RL algorithms, the absence of hierarchies in these settings result in poor performing policies in which the agent stands still or follows sub-optimal greedy behaviors.

\noindent \textbf{Delayed feedback.} \ \
In the mixed autonomy traffic tasks, meaningful events occur over large periods of time, as oscillations in vehicle speeds propagate slowly through a network. As a result, actions often have a delayed effect on metrics of improvement or success, making reasoning on whether a certain action improved the state of traffic a particularly difficult task. The delayed nature of this feedback prevents standard RL techniques from generating meaningful policies without relying on reward shaping techniques which produce undesirable behaviors such as creating large gaps between vehicles~\citep{wu2017flow}.

\begin{figure*}







\begin{subfigure}[b]{0.22\textwidth}
    \centering
    \includegraphics[width=\linewidth]{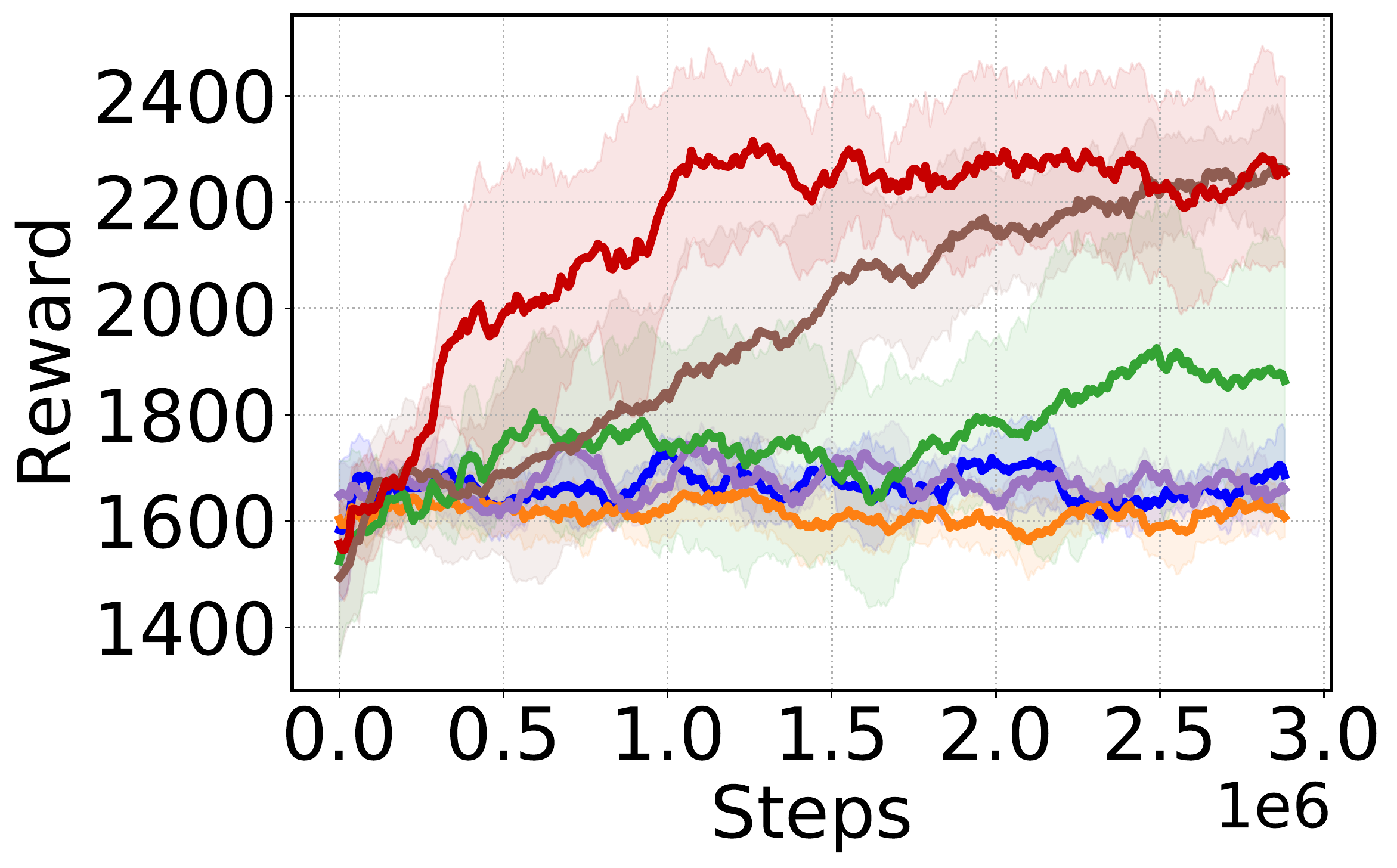}
    \caption{\footnotesize \ringroad}
\end{subfigure}
\hfill
\begin{subfigure}[b]{0.22\textwidth}
    \centering
    \includegraphics[width=\linewidth]{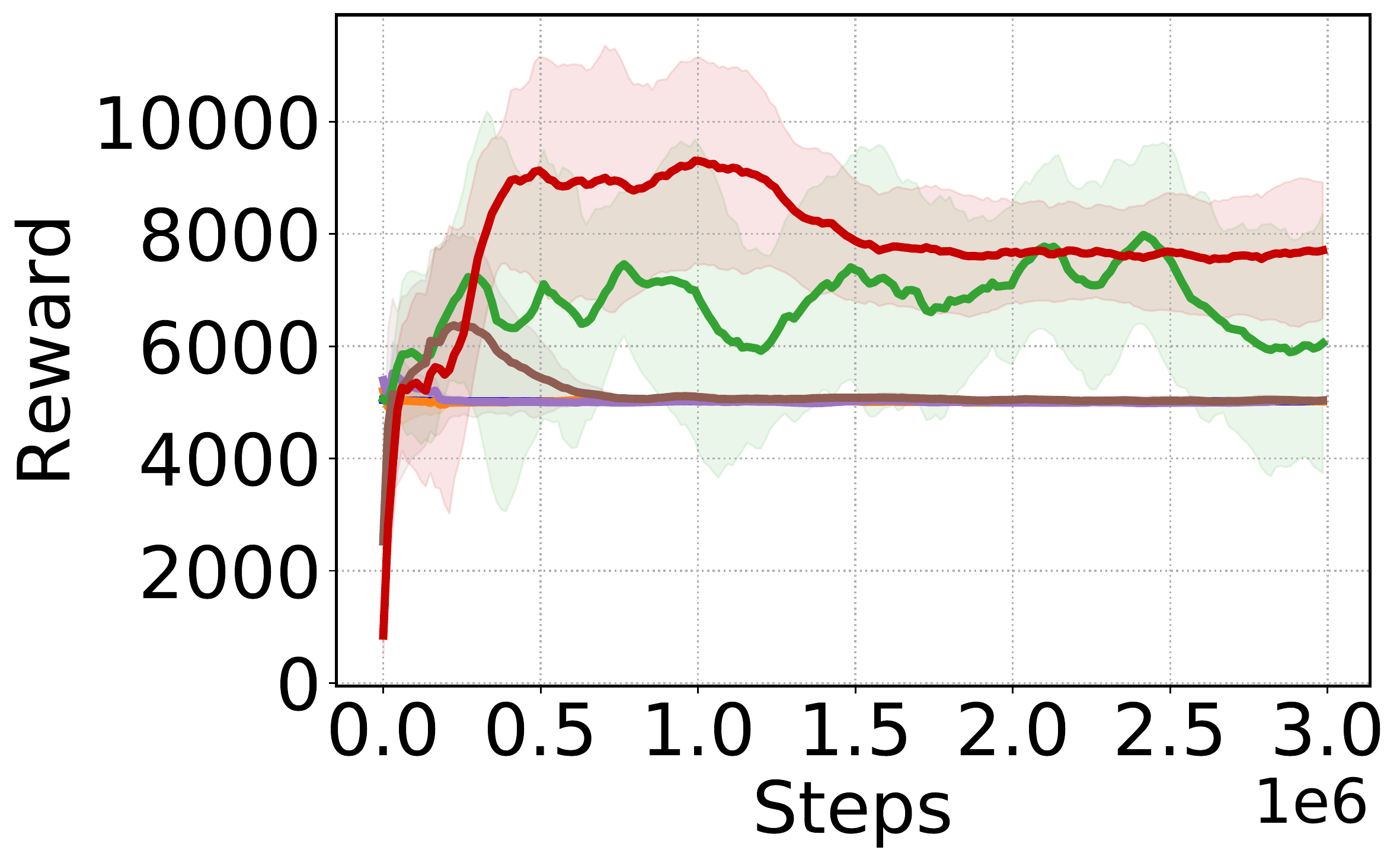}
    \caption{\footnotesize \highwaysingle}
\end{subfigure}
\hfill
\begin{subfigure}[b]{0.205\textwidth}
    \centering
    \includegraphics[width=\linewidth]{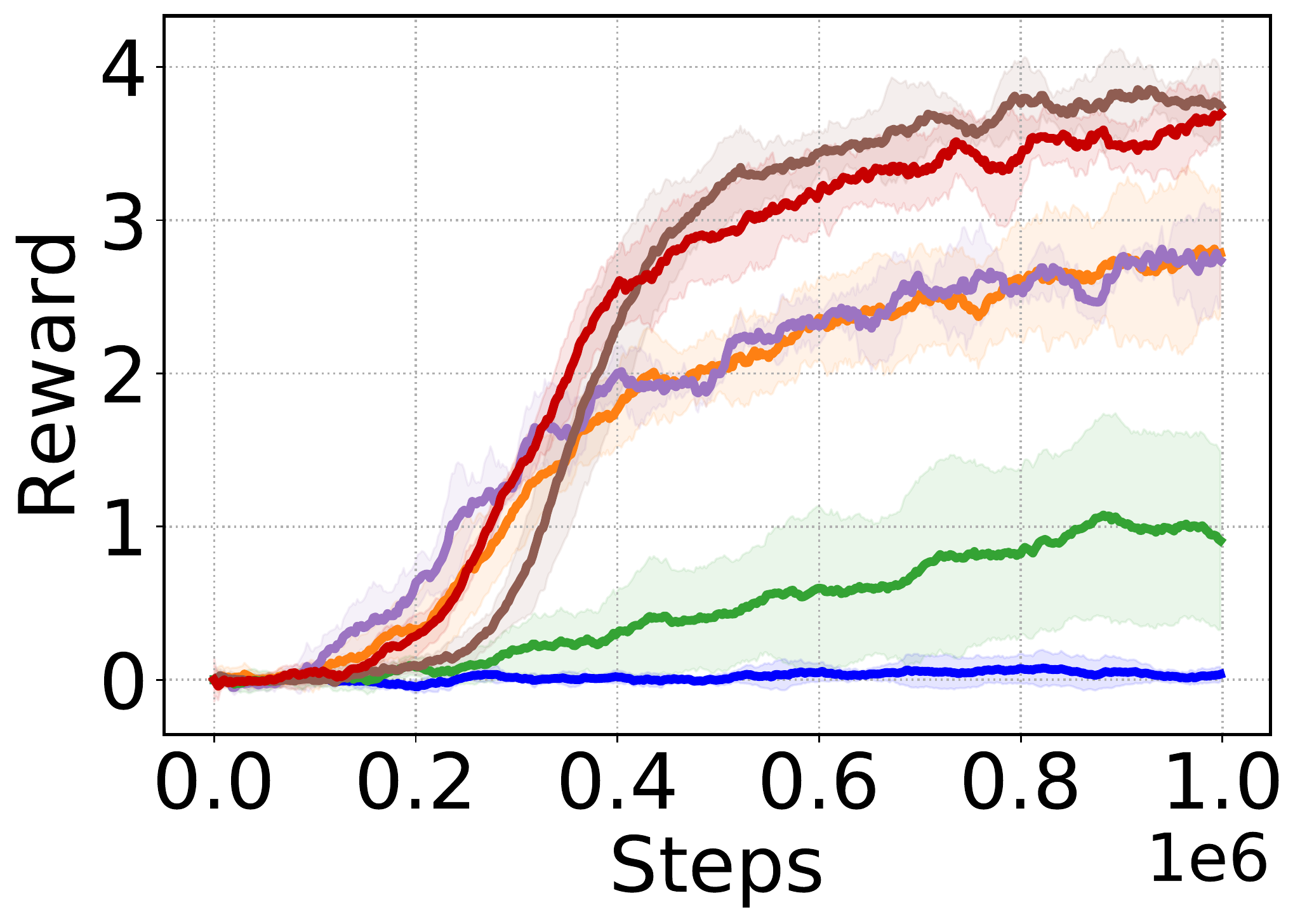}
    \caption{\footnotesize \antgather}
    \label{fig:antgather-rewards}
\end{subfigure}
\hfill
\begin{subfigure}[b]{0.22\textwidth}
    \centering
    \begin{tikzpicture}
        \newcommand \boxwidth {1.0}
        \newcommand \boxheight {\textwidth}
        \newcommand \boxoffset {0}
    
        \draw (0.25, -2.0) node {};
    
        \draw [color=color1]
        (0.25, \boxwidth + \boxoffset - 0.5) -- 
        (1.00, \boxwidth + \boxoffset - 0.5); 
        \draw [anchor=west] 
        (1.05, \boxwidth + \boxoffset - 0.5) node {\scriptsize TD3};
    
        \draw [color=color2]
        (0.25, \boxwidth + \boxoffset - 0.9) -- 
        (1.00, \boxwidth + \boxoffset - 0.9); 
        \draw [anchor=west] 
        (1.05, \boxwidth + \boxoffset - 0.9) node {\scriptsize HRL};
    
        \draw [color=color3] 
        (0.25, \boxwidth + \boxoffset - 1.3) -- 
        (1.00, \boxwidth + \boxoffset - 1.3); 
        \draw [anchor=west] 
        (1.05, \boxwidth + \boxoffset - 1.3) node {\scriptsize HIRO};
    
        \draw [color=color4] 
        (0.25, \boxwidth + \boxoffset - 1.7) -- 
        (1.00, \boxwidth + \boxoffset - 1.7); 
        \draw [anchor=west] 
        (1.05, \boxwidth + \boxoffset - 1.7) node {\scriptsize HAC};
    
        \draw [color=color5] 
        (0.25, \boxwidth + \boxoffset - 2.1) -- 
        (1.00, \boxwidth + \boxoffset - 2.1); 
        \draw [anchor=west] 
        (1.05, \boxwidth + \boxoffset - 2.1) node {\scriptsize \methodName (fixed)};
    
        \draw [color=color6] 
        (0.25, \boxwidth + \boxoffset - 2.5) -- 
        (1.00, \boxwidth + \boxoffset - 2.5); 
        \draw [anchor=west] 
        (1.05, \boxwidth + \boxoffset - 2.5) node {\scriptsize \methodName (dynamic)};
    \end{tikzpicture}
\end{subfigure} \\ \vspace{0.2cm}
\begin{subfigure}[b]{0.22\textwidth}
    \centering
    \includegraphics[width=\linewidth]{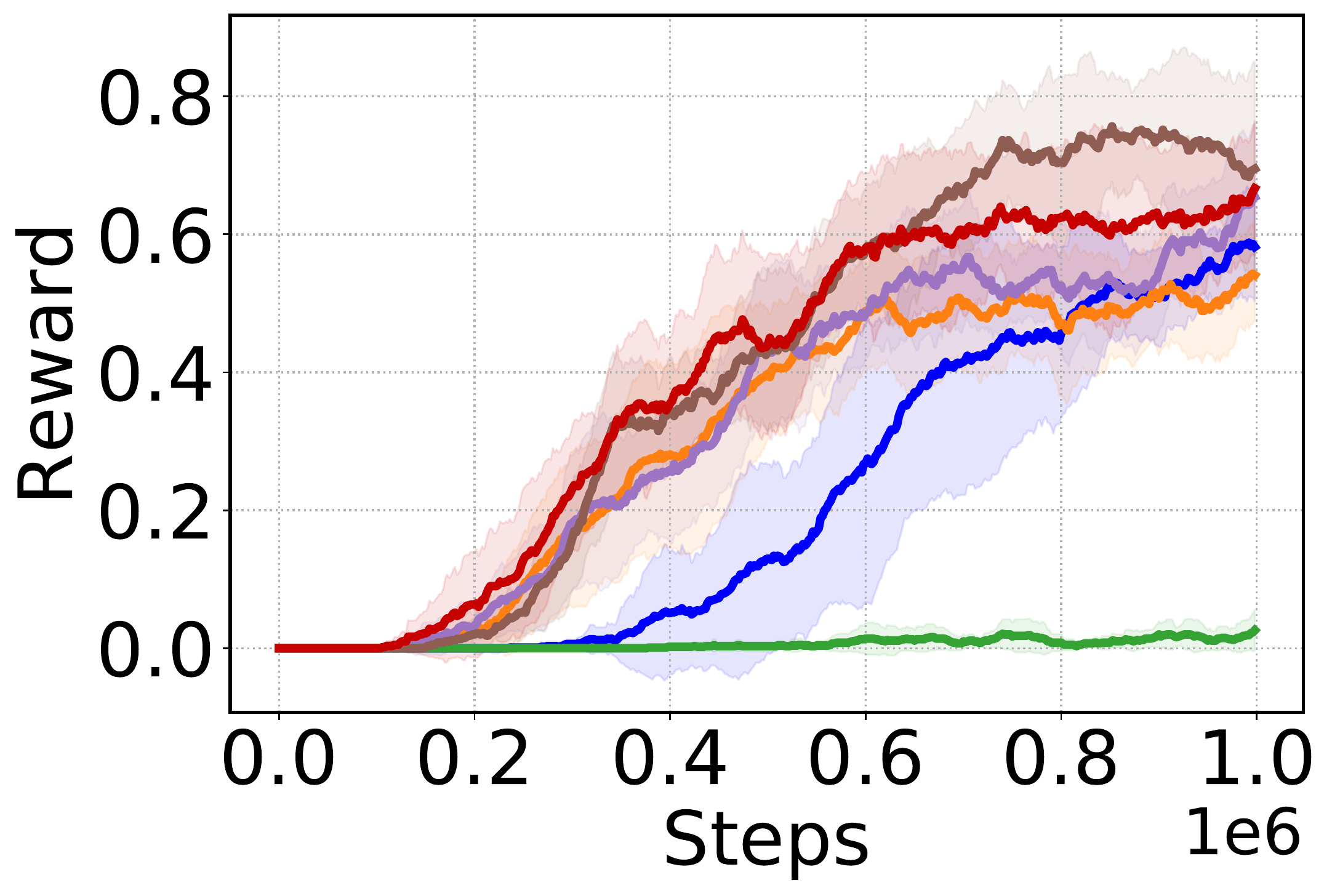}
    \caption{\footnotesize \antfourrooms}
\end{subfigure}
\hfill
\begin{subfigure}[b]{0.22\textwidth}
    \centering
    \includegraphics[width=\linewidth]{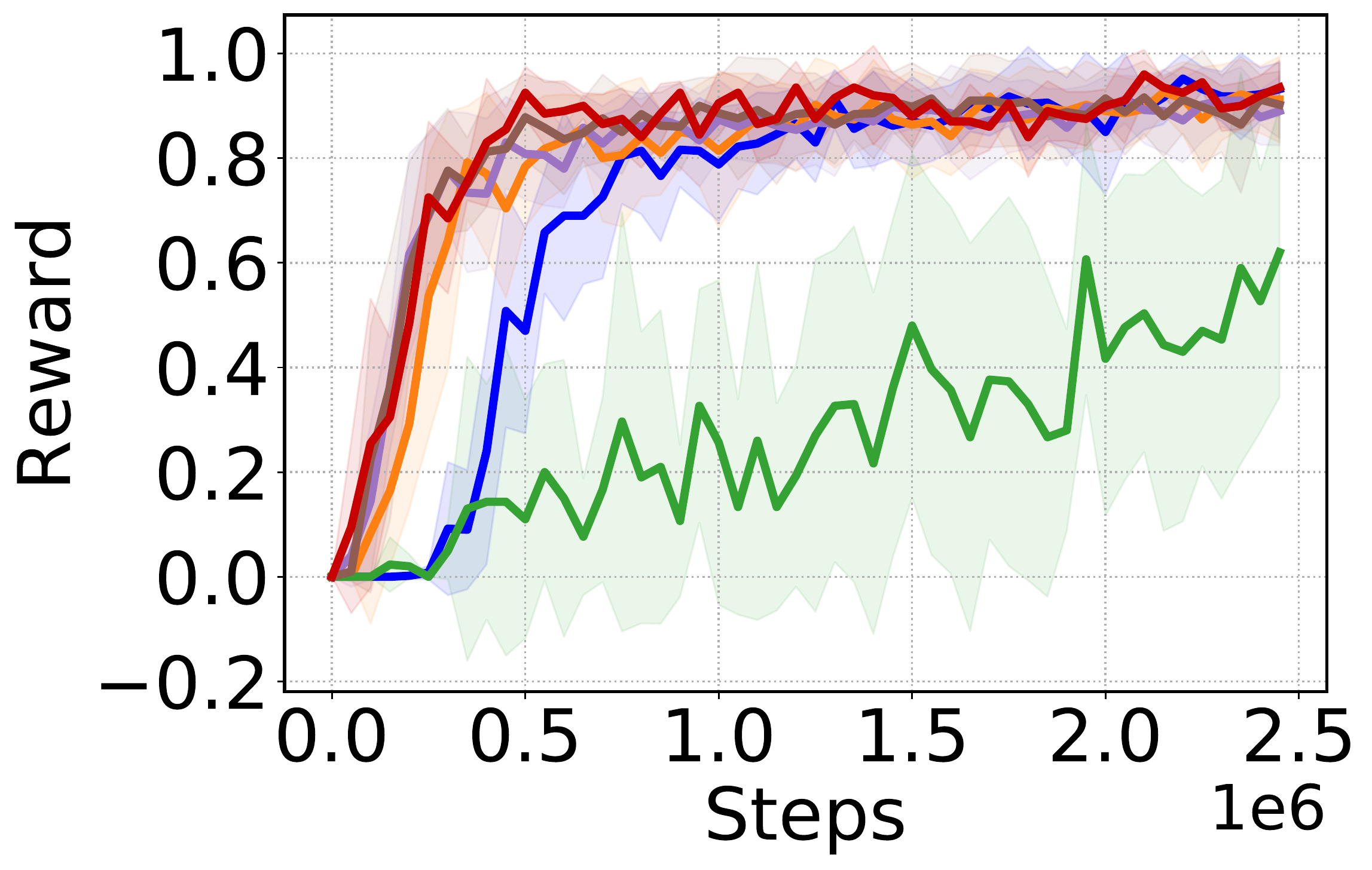}
    \caption{\footnotesize \antmaze [16,0]}
\end{subfigure}
\hfill
\begin{subfigure}[b]{0.22\textwidth}
    \centering
    \includegraphics[width=\linewidth]{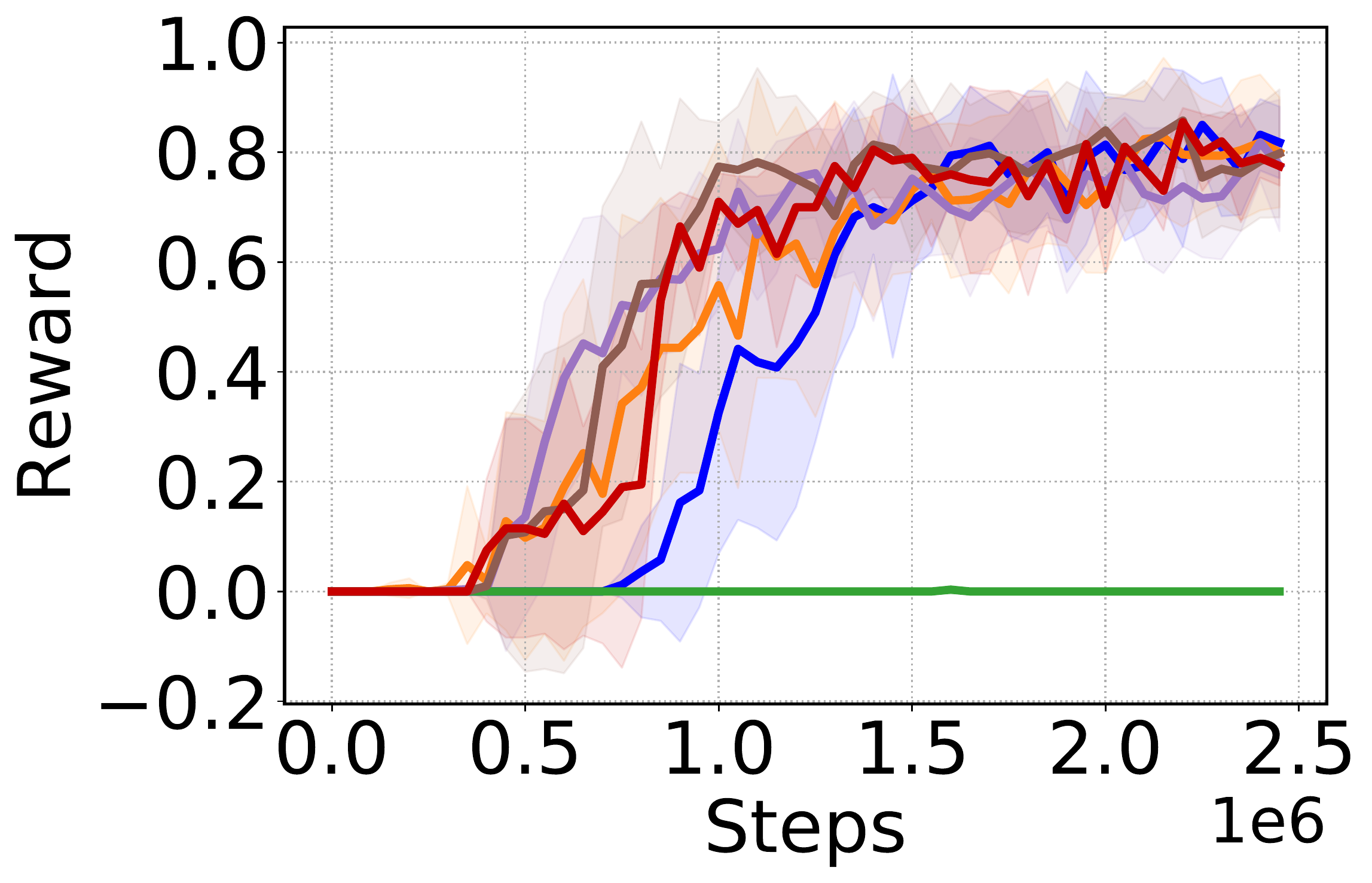}
    \caption{\footnotesize \antmaze [16,16]}
\end{subfigure}
\hfill
\begin{subfigure}[b]{0.22\textwidth}
    \centering
    \includegraphics[width=\linewidth]{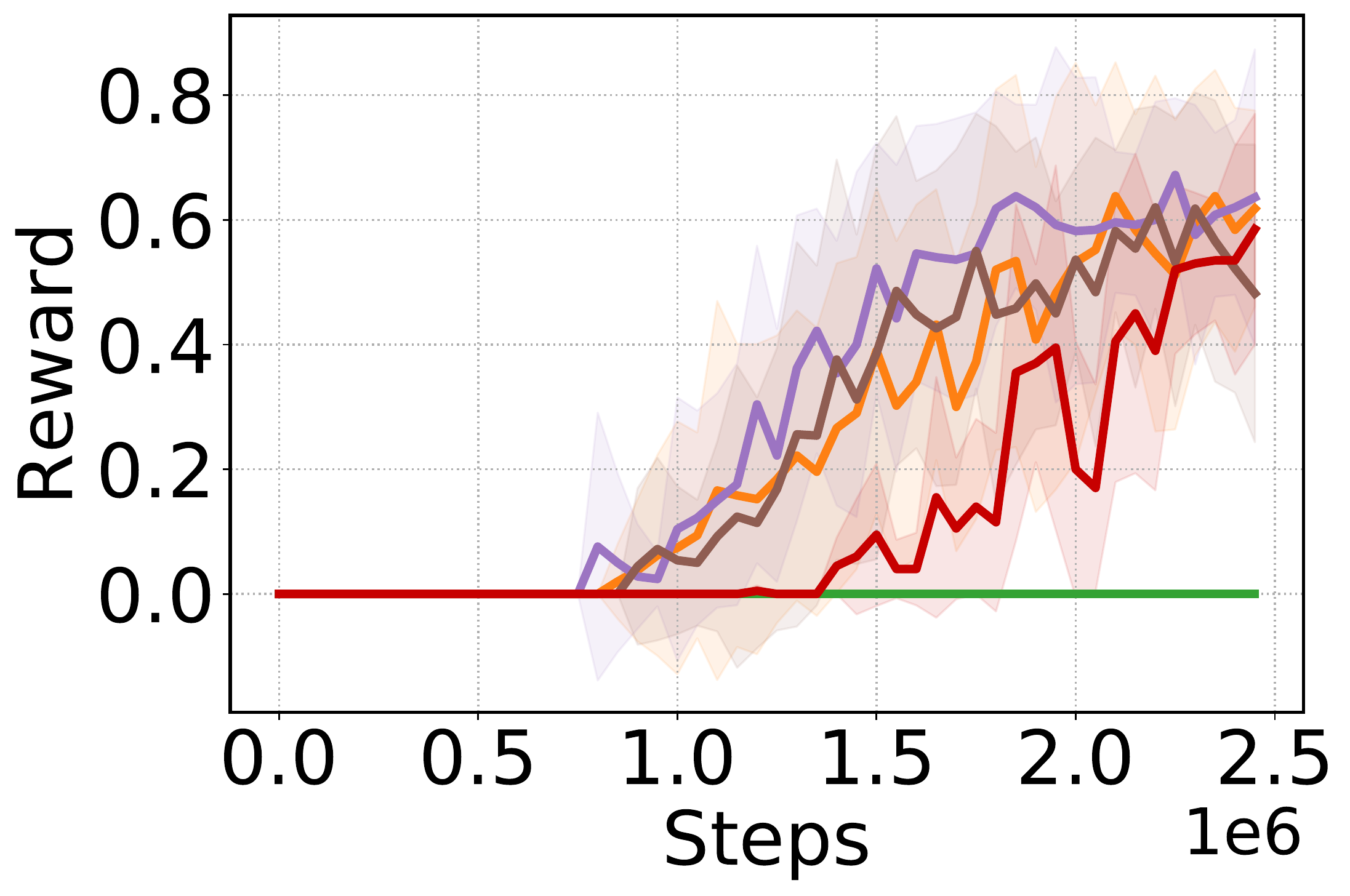}
    \caption{\footnotesize \antmaze [0,16]}
\end{subfigure}
\caption{Training performance of the different algorithms on various environments. All results are reported over $10$ random seeds. Details on the choice of hyperparameters are provided in Appendix~\ref{sec:hyperparams}.
}
\label{fig:learning-curves}
\end{figure*}

\subsection{Baseline algorithms}

We evaluate our proposed algorithm against the following baseline methods:\\[-15pt]
\begin{itemize}[leftmargin=*,noitemsep]
    \item \textit{TD3}: To validate the need for goal-conditioned hierarchies to solve our tasks, we compare all results against a fully connected network with otherwise similar network and training configurations.
    \item \textit{HRL}: This baseline consists of the naive formulation of the hierarchical reinforcement learning optimization scheme (see Section~\ref{sec:no-cooperation}). This algorithm is analogous to the \methodName algorithm with $\lambda$ set to $0$.
    \item \textit{HIRO}: Presented by~\citet{nachum2018data}, this method addresses the non-stationarity effects between the manager and worker policies by relabeling the manager's actions (or goals) to render the observed action sequence more likely to have occurred by the current instantiation of the worker policy. The details of this method are discussed in Appendix~\ref{sec:hiro-reproducibility}.
    \item \textit{HAC}: The HAC algorithm~\citep{levy2017hierarchical} attempts to address non-stationarity in off-policy learning by relabeling sampled data via hindsight action and goal transitions as well as subgoal testing transitions to prevent learning from a restricted set of subgoal states. Implementation details are provided in Appendix~\ref{sec:hac-egocentric}.\\[-15pt]
\end{itemize}

\begin{figure*}
\centering
\begin{subfigure}[b]{\textwidth}
\begin{tikzpicture}
    \newcommand \boxwidth {1.0}
    \newcommand \boxheight {\textwidth}
    \newcommand \boxoffset {0}

    \draw (0, -0.15) node {};

    \filldraw [blue] (0.615 * \textwidth, \boxwidth + \boxoffset - 0.5) circle (0.125);
    \draw [anchor=west]
    (0.63 * \textwidth, \boxwidth + \boxoffset - 0.5) node {\scriptsize High-level goal};

    \draw [dashed] 
    (0.790 * \textwidth, \boxwidth + \boxoffset - 0.5) -- 
    (0.825 * \textwidth, \boxwidth + \boxoffset - 0.5); 
    \draw [anchor=west]
    (0.83 * \textwidth, \boxwidth + \boxoffset - 0.5) node {\scriptsize Agent trajectory};
\end{tikzpicture}
\end{subfigure} \\ \vspace{-0.4cm}
\begin{subfigure}[b]{0.08\textwidth}
    \small{Standard\\ HRL} \\[10pt]
    
    \textcolor{white}{.}
\end{subfigure}
\hfill
\begin{subfigure}[b]{0.42\textwidth}
    \centering
    \includegraphics[width=0.48\linewidth]{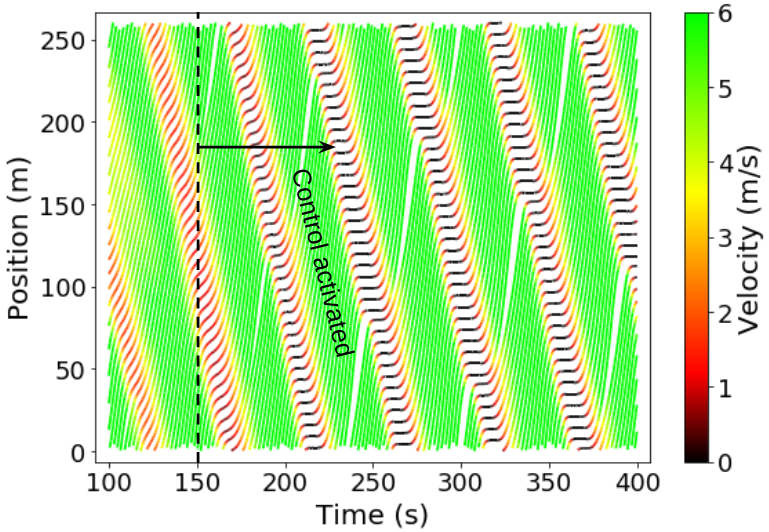}
    \hfill
    \includegraphics[width=0.48\linewidth]{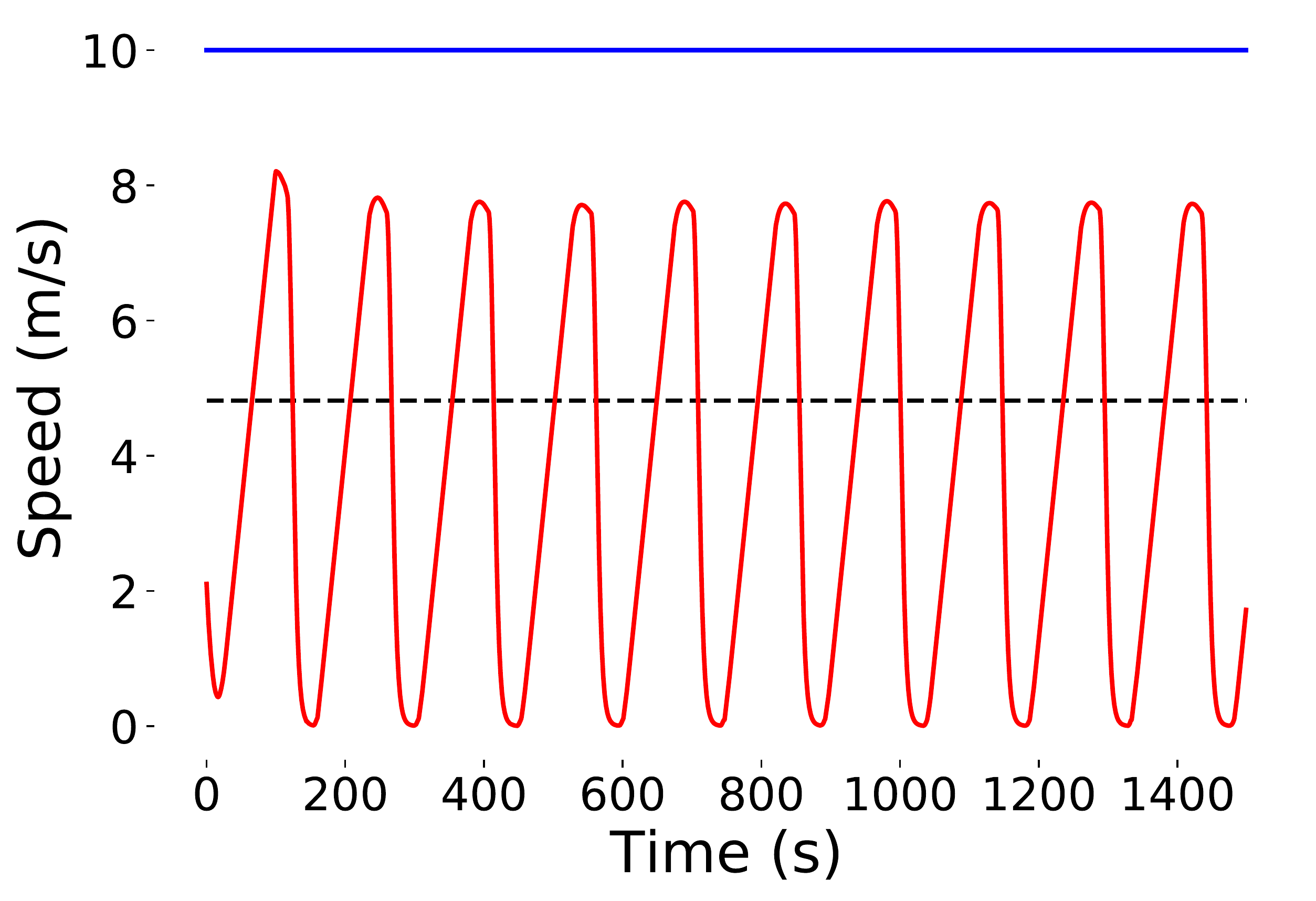}
\end{subfigure}
\hfill
\begin{subfigure}[b]{0.15\textwidth}
    \centering
    \includegraphics[width=\linewidth]{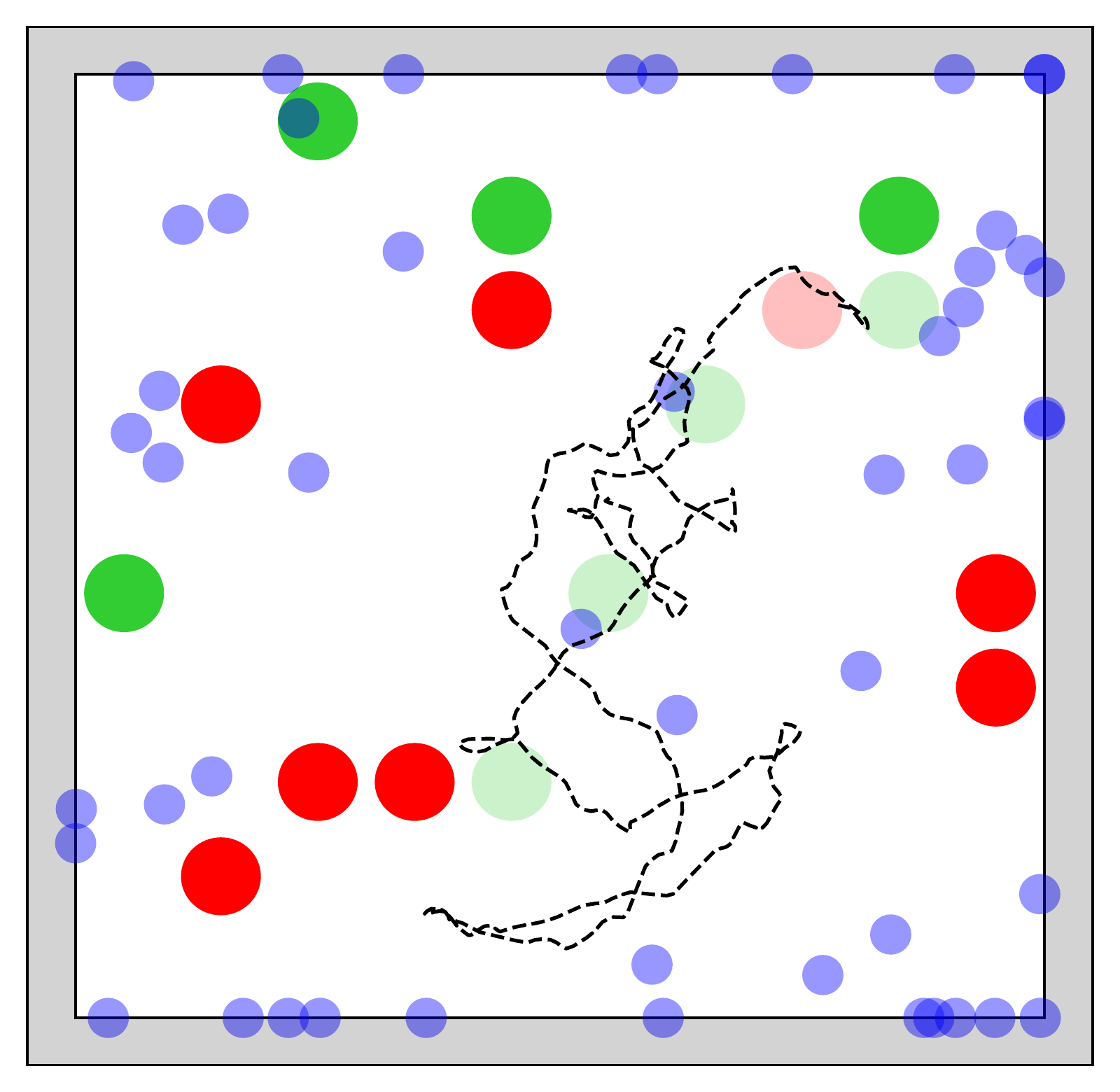}
\end{subfigure}
\hfill
\begin{subfigure}[b]{0.15\textwidth}
    \includegraphics[width=\linewidth]{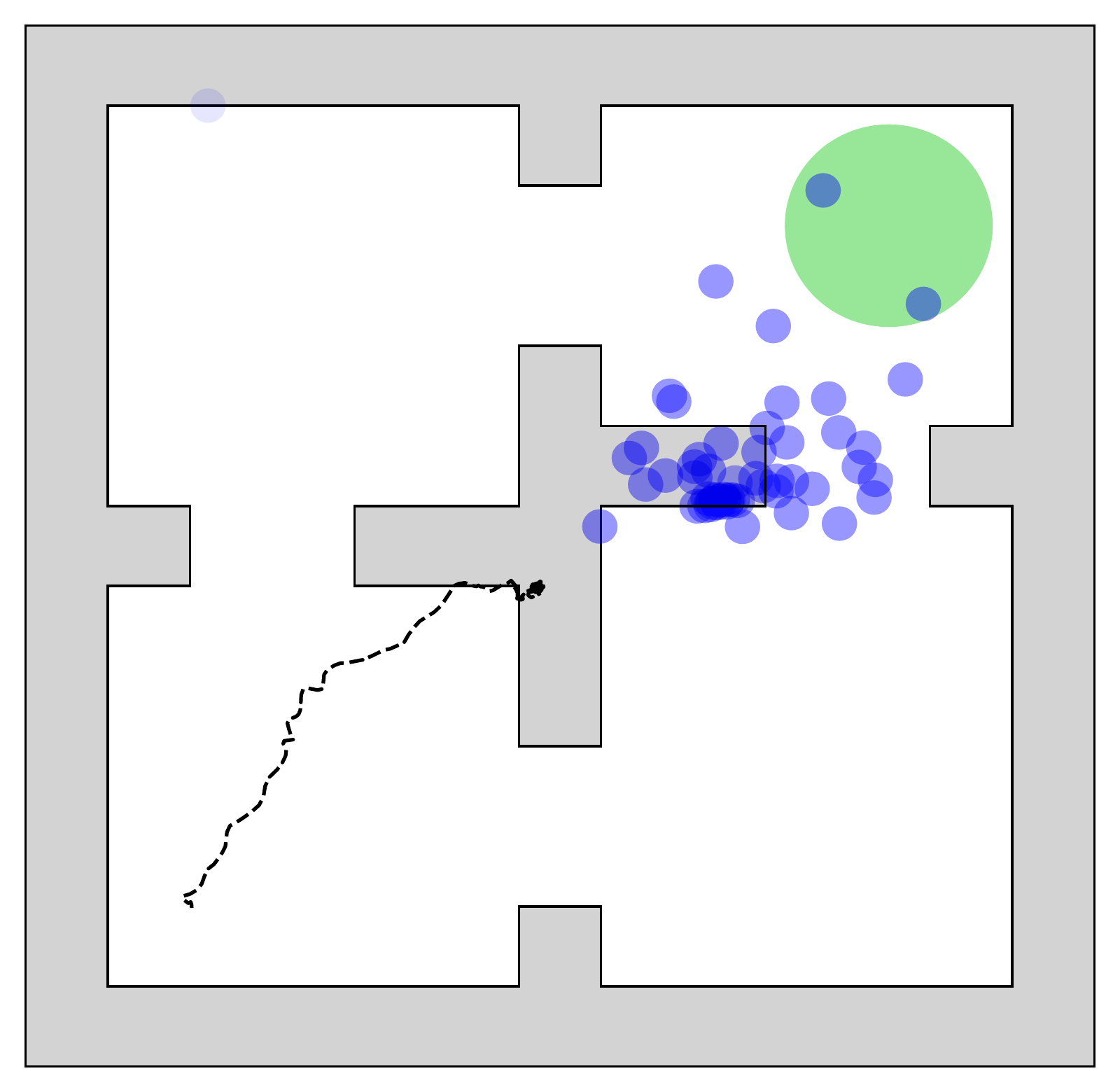}
\end{subfigure}
\hfill
\begin{subfigure}[b]{0.15\textwidth}
    \centering
    \includegraphics[width=\linewidth]{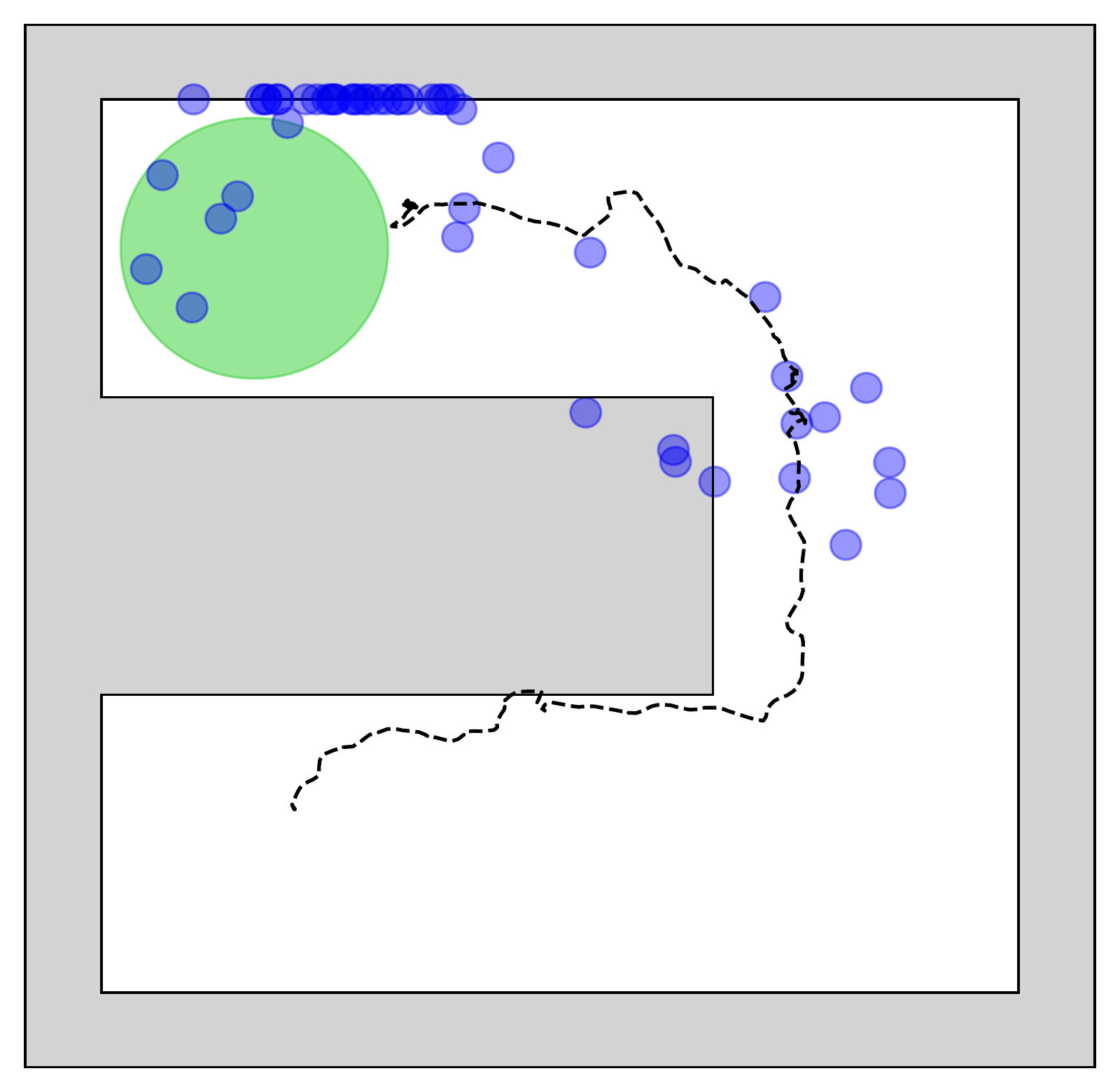}
\end{subfigure}
\\[5pt]
\begin{subfigure}[b]{0.08\textwidth}
    \small{\methodName\\ (ours)} \\[25pt]
    
    \textcolor{white}{.}
\end{subfigure}
\hfill
\begin{subfigure}[b]{0.42\textwidth}
    \centering
    \includegraphics[width=0.48\linewidth]{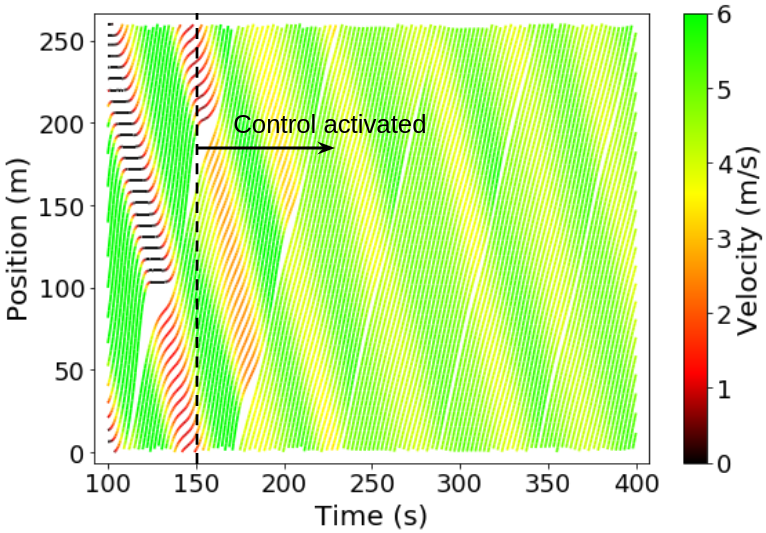}
    \hfill
    \includegraphics[width=0.48\linewidth]{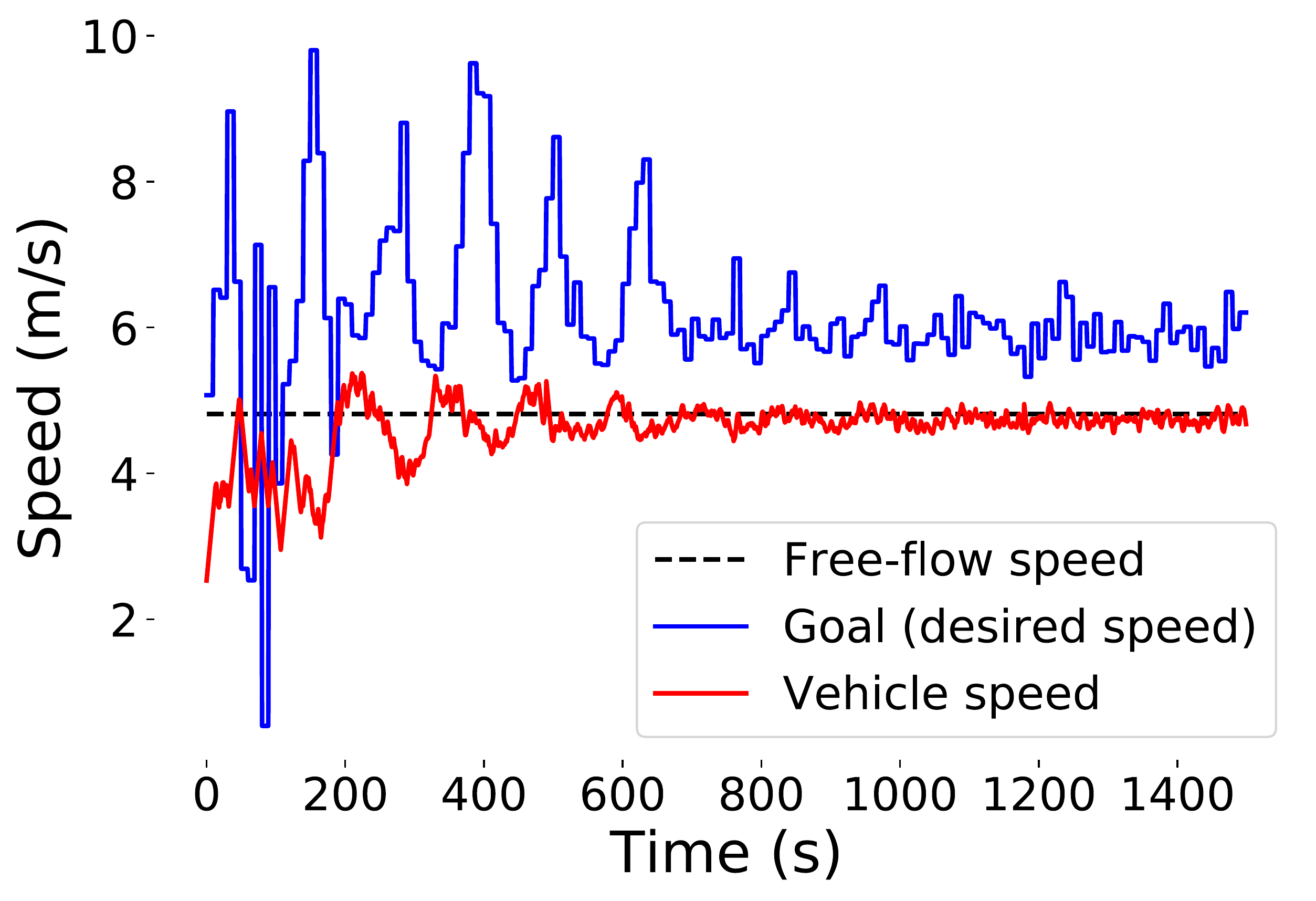}
    \caption{\ringroad}
    \label{fig:ring-traj}
\end{subfigure}
\hfill
\begin{subfigure}[b]{0.15\textwidth}
    \centering
    \includegraphics[width=\linewidth]{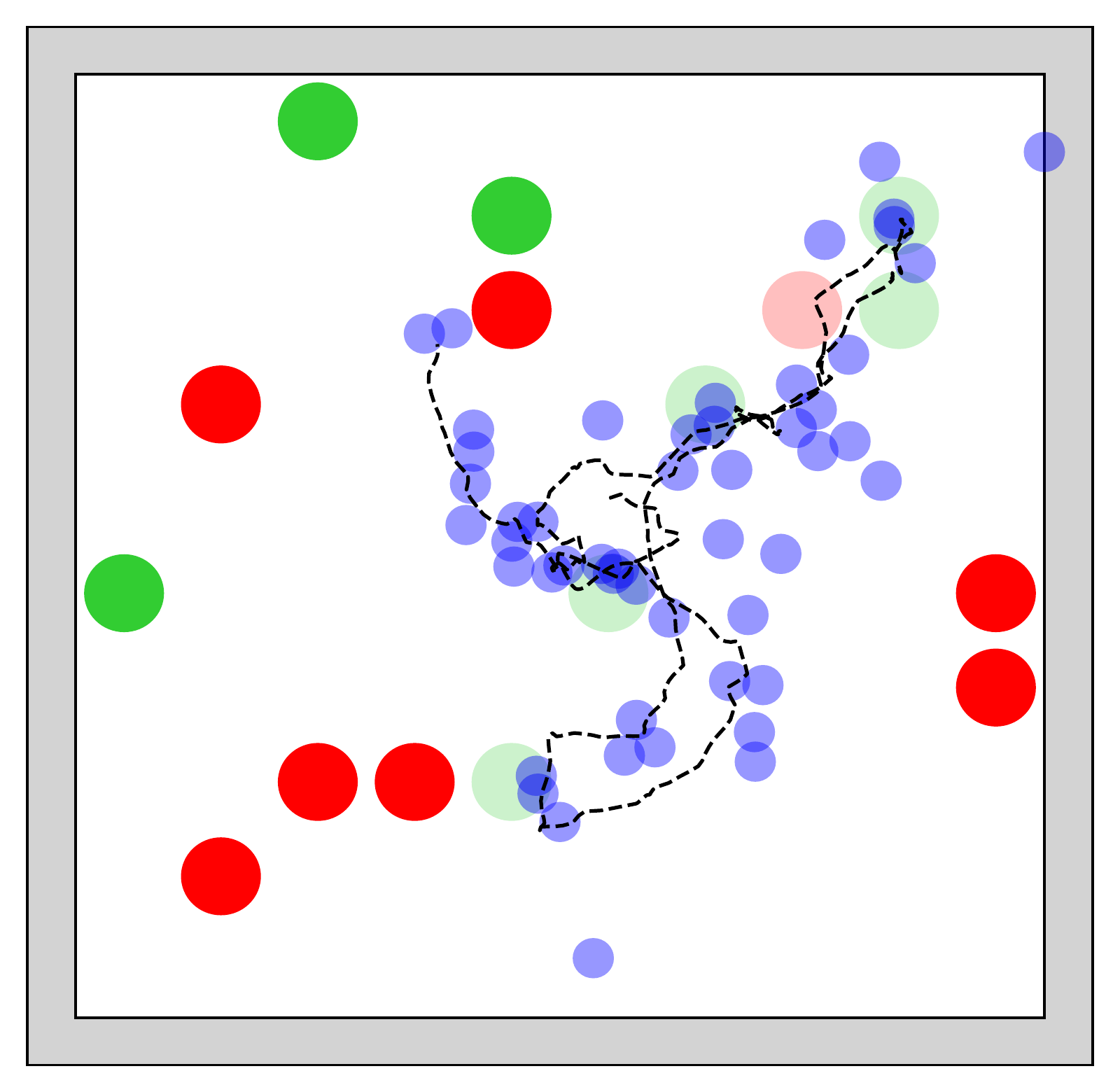}
    \caption{Gather}
    \label{fig:antgather-traj}
\end{subfigure}
\hfill
\begin{subfigure}[b]{0.15\textwidth}
    \includegraphics[width=\linewidth]{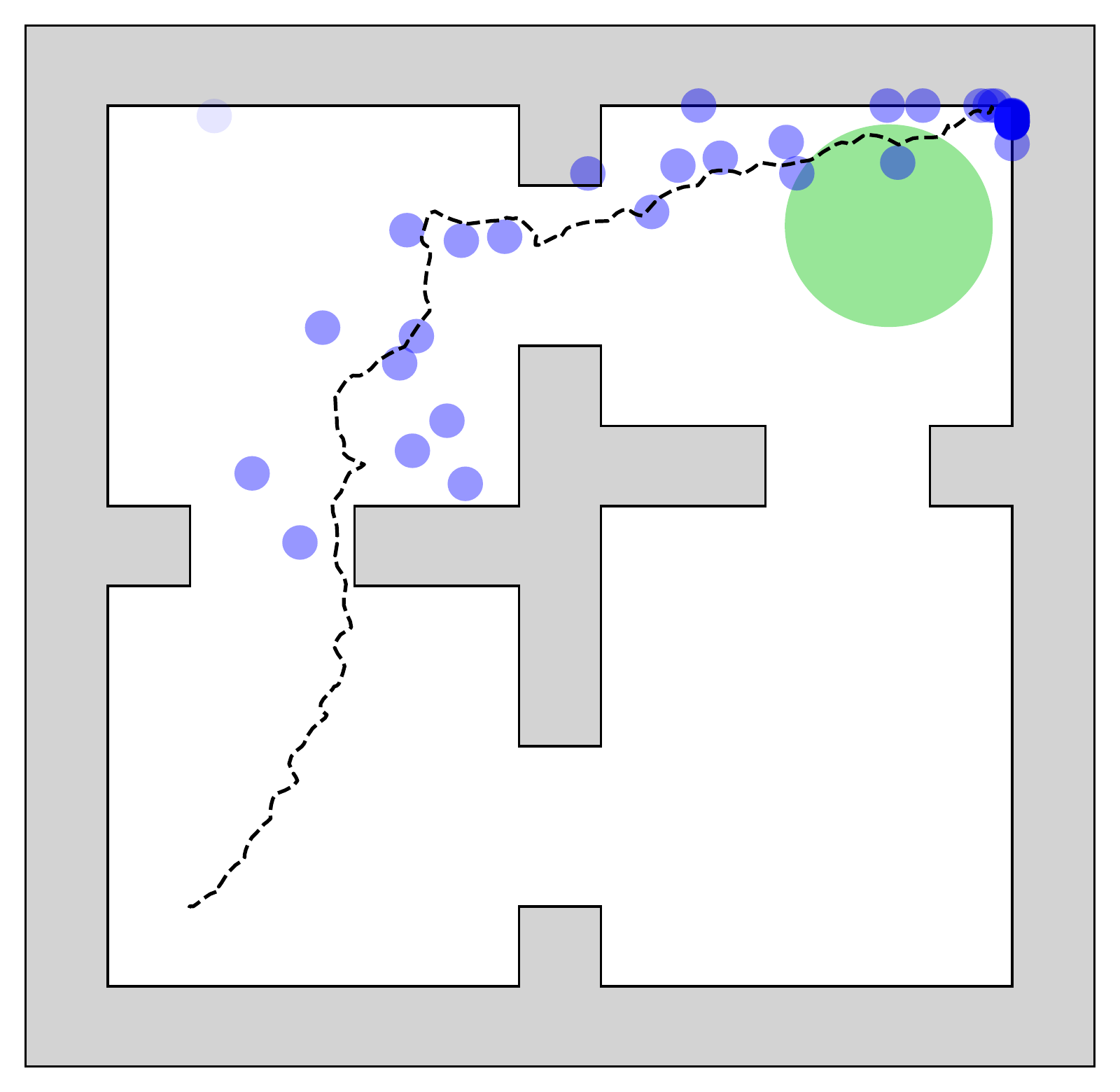}
    \caption{Four Rooms}
    \label{fig:antfourrooms-traj}
\end{subfigure}
\hfill
\begin{subfigure}[b]{0.15\textwidth}
    \centering
    \includegraphics[width=\linewidth]{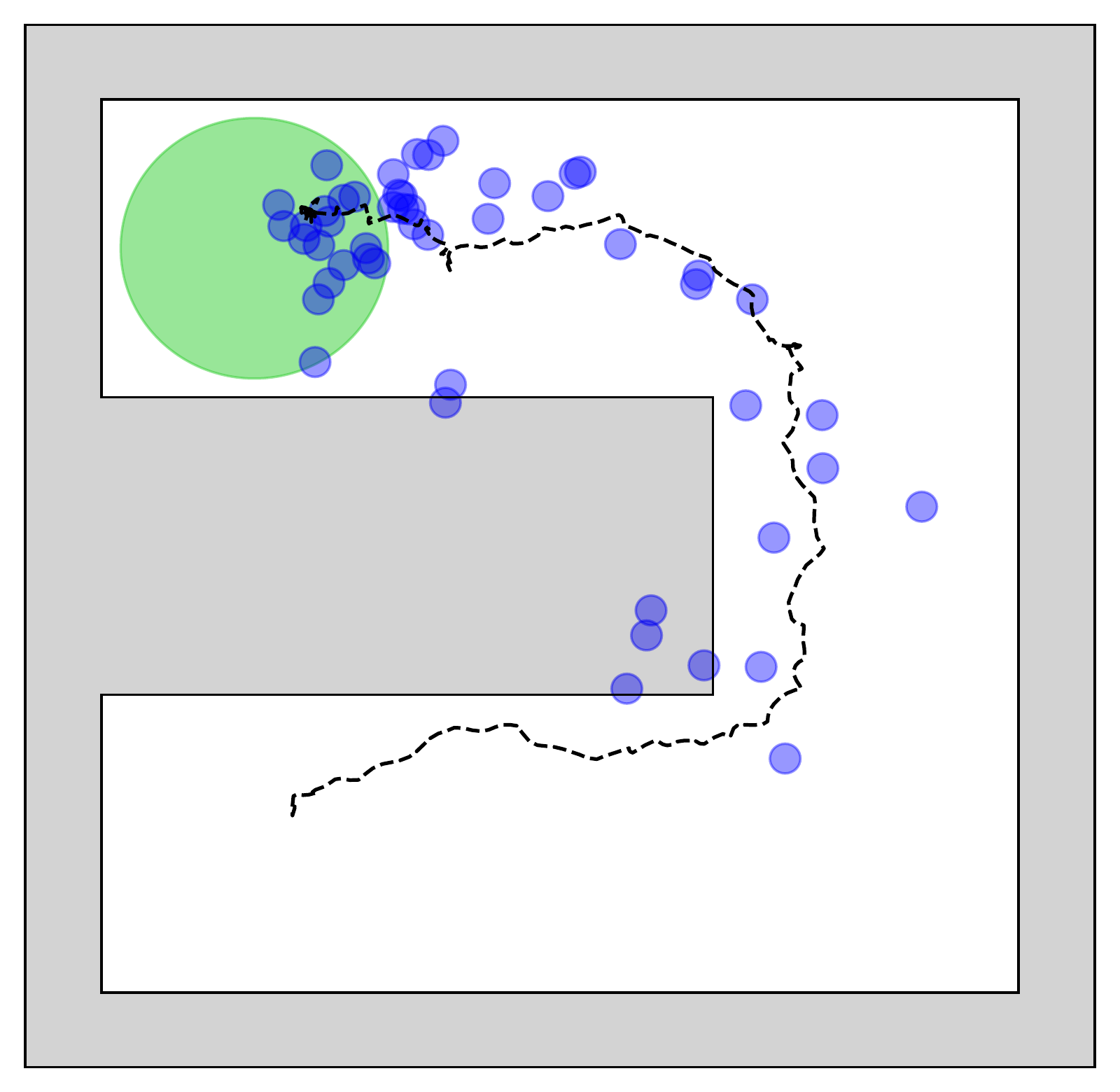}
    \caption{\footnotesize Maze}
    \label{fig:antmaze-traj}
\end{subfigure}
\caption{Illustration of the agent and goal trajectories for some of the environments studied here. The high-level goals learned via \methodName more closely match the agents' trajectories as they traverse the various environments. This results in improved dynamical behaviors by the agent in a majority of the tasks.}
\label{fig:goal-trajectories}
\end{figure*}
\subsection{Comparative analysis}

Figure~\ref{fig:learning-curves} depicts the training performance of each of the above algorithms and \methodName on the studied tasks. We find that \methodName performs comparably or outperforms all other studied algorithms for the provided tasks. Interestingly, \methodName particularly outperforms other algorithms in highly stochastic and partially observable tasks. For the \antgather and traffic control tasks, in particular, the objective (be it the positions of apples and bombs or the density and driving behaviors within a traffic network) varies significantly at the start of a new rollout and is not fully observable to the agent from its local observation. As such, the improved performance by \methodName within these settings suggests that it is more robust than previous methods to learning policies in noisy and unstable environments.

The improvements presented above emerge in part from more informative goal-assignment behaviors by \methodName. Figure~\ref{fig:goal-trajectories} depicts these behaviors for a large number of tasks. We describe some of these behaviors and the performance of the policy below.

For the agent navigation tasks, the \methodName algorithm produces goals that more closely match the agent's trajectory, providing the agent with a more defined path to follow to achieve certain goals. This results in faster and more efficient learning for settings such as \antfourrooms, and in stronger overall policies for tasks such as \antgather. An interesting corollary that appears to emerge as a result of this cooperative approach is more stable movements and actuation commands by the worker policy. For settings in which the agent can fall prematurely, this appears in the form of fewer early terminations as a result of agents attempting to achieve difficult or highly random goals. The absence of frequent early terminations allows the policy to explore further into its environment during training; this is a benefit in settings where long-term reasoning is crucial.

In the \ringroad environment, we find standard HRL techniques fail to learn goal-assignment strategies that yield meaningful performative benefits. Instead, as seen in Figure~\ref{fig:ring-traj} (top), the manager overestimates its worker's ability to perform certain tasks and assign maximum desired speeds. This strategy prevents the policy from dissipating the continued propagation of vehicle oscillations in the traffic, as the worker policy is forced to assign large acceleration to match the desired speeds thereby contributing to the formation of stop-and-go traffic, seen as the red diagonal streaks in Figure \ref{fig:ring-traj}, top-left. 
For these tasks, we find that inducing inter-level cooperation serves to alleviate the challenge of overestimating certain goals. In the \ringroad environment, for instance, \methodName succeeds in assigning meaningful desired speeds that encourage the worker policy to reduce the magnitude of accelerations assigned while near the free-flow, or optimal, speed of the network. This serves to eliminate the formation of stop-and-go traffic both from the perspective of the automated and human-driven vehicles, as seen in Figure \ref{fig:ring-traj}, bottom. We also note that when compared to previous studies of a similar task~\citep{wu2017flow}, our approach succeeds in finding a solution that does not rely on the generation of large or undesirable gaps between the AV and its leader. Similar results for the \highwaysingle environment are provided in Appendix~\ref{appendix:highway-results}.

\begin{figure*}
\centering
\begin{subfigure}[b]{0.25\textwidth}
    \centering
    \includegraphics[width=\linewidth]{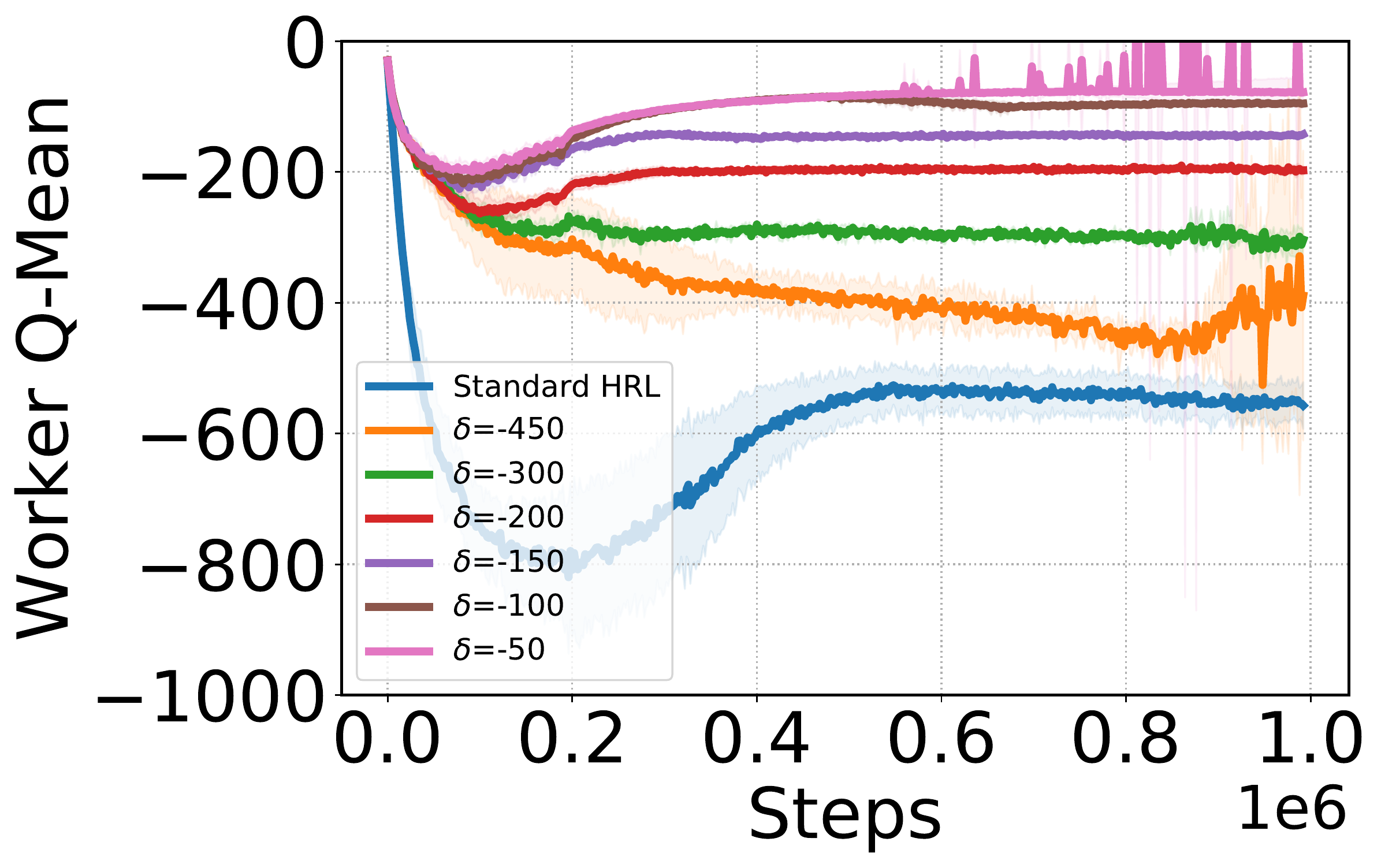}
    \caption{Worker expected returns}
    \label{fig:cg-delta-worker-qvals}
\end{subfigure}
\qquad
\begin{subfigure}[b]{0.25\textwidth}
    \centering
    \includegraphics[width=\linewidth]{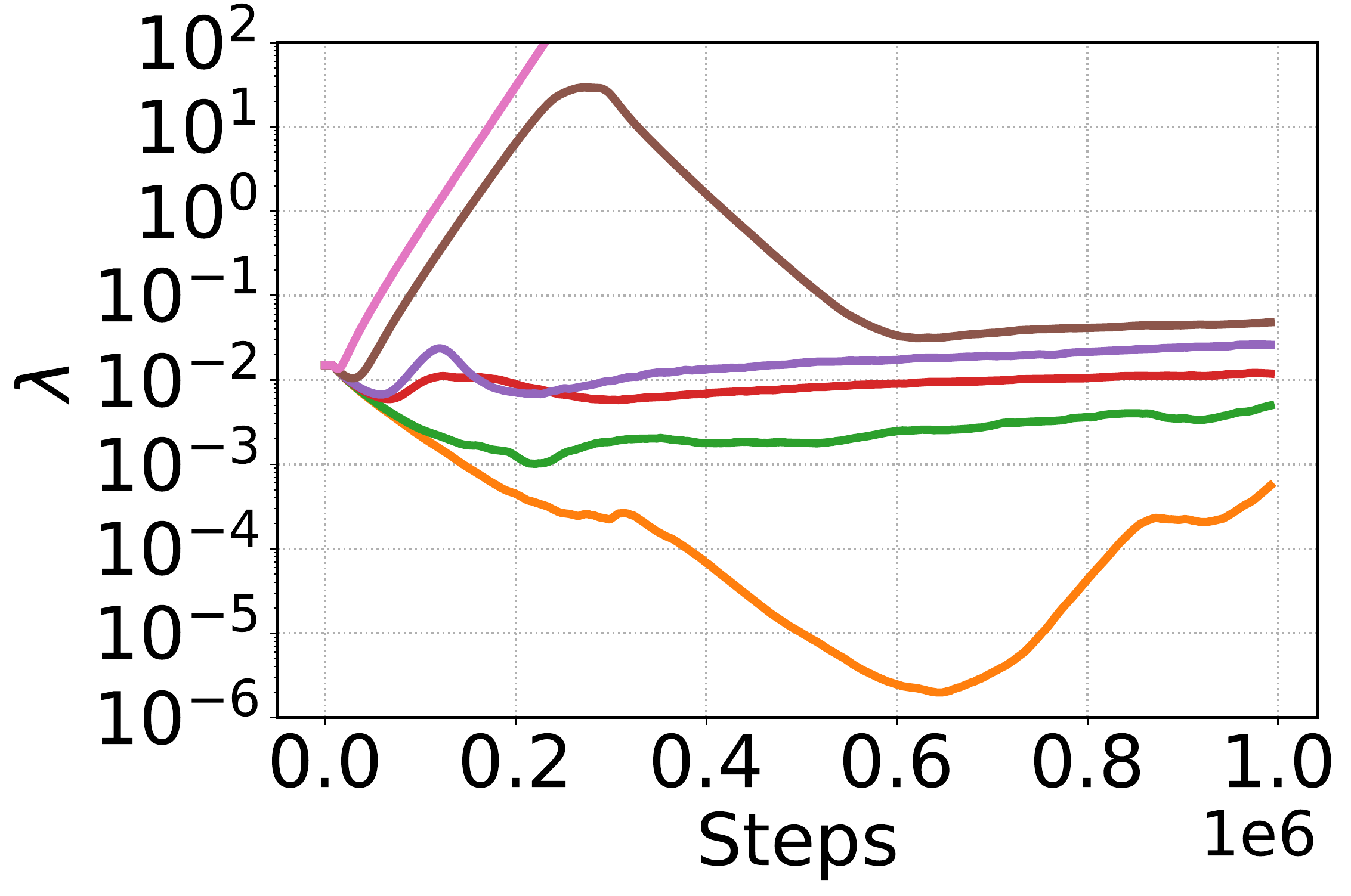}
    \caption{Evolution of dynamic $\lambda$}
    \label{fig:cg-delta-lambda}
\end{subfigure}
\qquad
\begin{subfigure}[b]{0.25\textwidth}
    \centering
    \includegraphics[width=\linewidth]{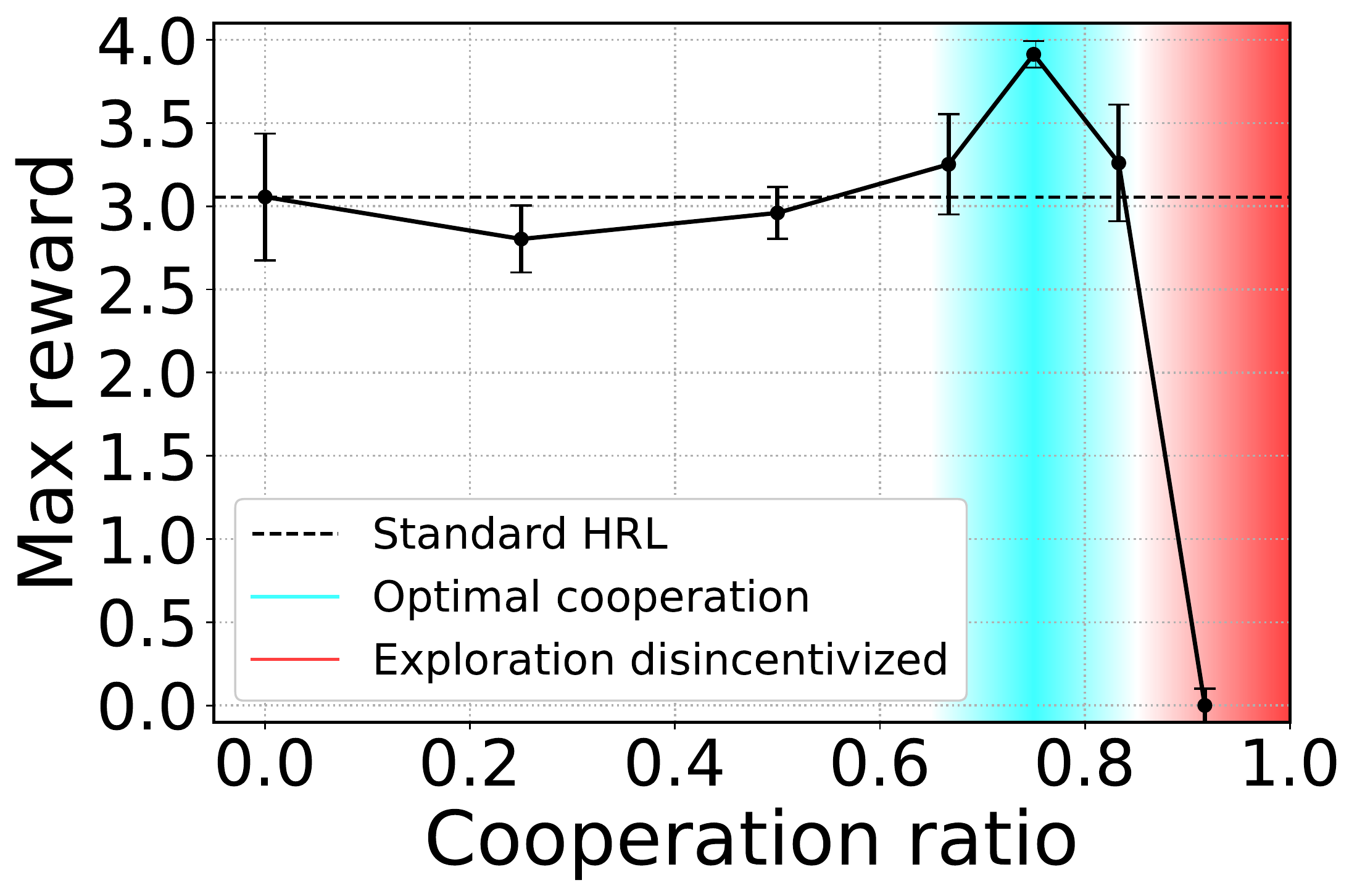}
    \caption{Effects of cooperation}
    \label{fig:cg-delta-rewards}
\end{subfigure}\\ \vspace{0.5cm}
\begin{subfigure}[b]{\textwidth}
\centering
\begin{tikzpicture}
    \newcommand \boxwidth {1.0}
    \newcommand \boxheight {\textwidth}
    \newcommand \boxoffset {0}

    \draw (0, -0.15) node {};

    \filldraw [limegreen] (0.305 * \textwidth,\boxwidth + \boxoffset - 0.5) circle (0.125);
    \draw [anchor=west]
    (0.32 * \textwidth, \boxwidth + \boxoffset - 0.5) node {\scriptsize Apple ($+1$)};

    \filldraw [red] (0.46 * \textwidth,\boxwidth + \boxoffset - 0.5) circle (0.125);
    \draw [anchor=west]
    (0.475 * \textwidth, \boxwidth + \boxoffset - 0.5) node {\scriptsize Bomb ($-1$)};

    \filldraw [blue] (0.615 * \textwidth, \boxwidth + \boxoffset - 0.5) circle (0.125);
    \draw [anchor=west]
    (0.63 * \textwidth, \boxwidth + \boxoffset - 0.5) node {\scriptsize High-level goal};

    \draw [dashed] 
    (0.790 * \textwidth, \boxwidth + \boxoffset - 0.5) -- 
    (0.825 * \textwidth, \boxwidth + \boxoffset - 0.5); 
    \draw [anchor=west]
    (0.83 * \textwidth, \boxwidth + \boxoffset - 0.5) node {\scriptsize Agent trajectory};
\end{tikzpicture}
\end{subfigure} \\ \vspace{-0.4cm}
\begin{subfigure}[b]{0.15\textwidth}
    \centering
    \includegraphics[width=\linewidth]{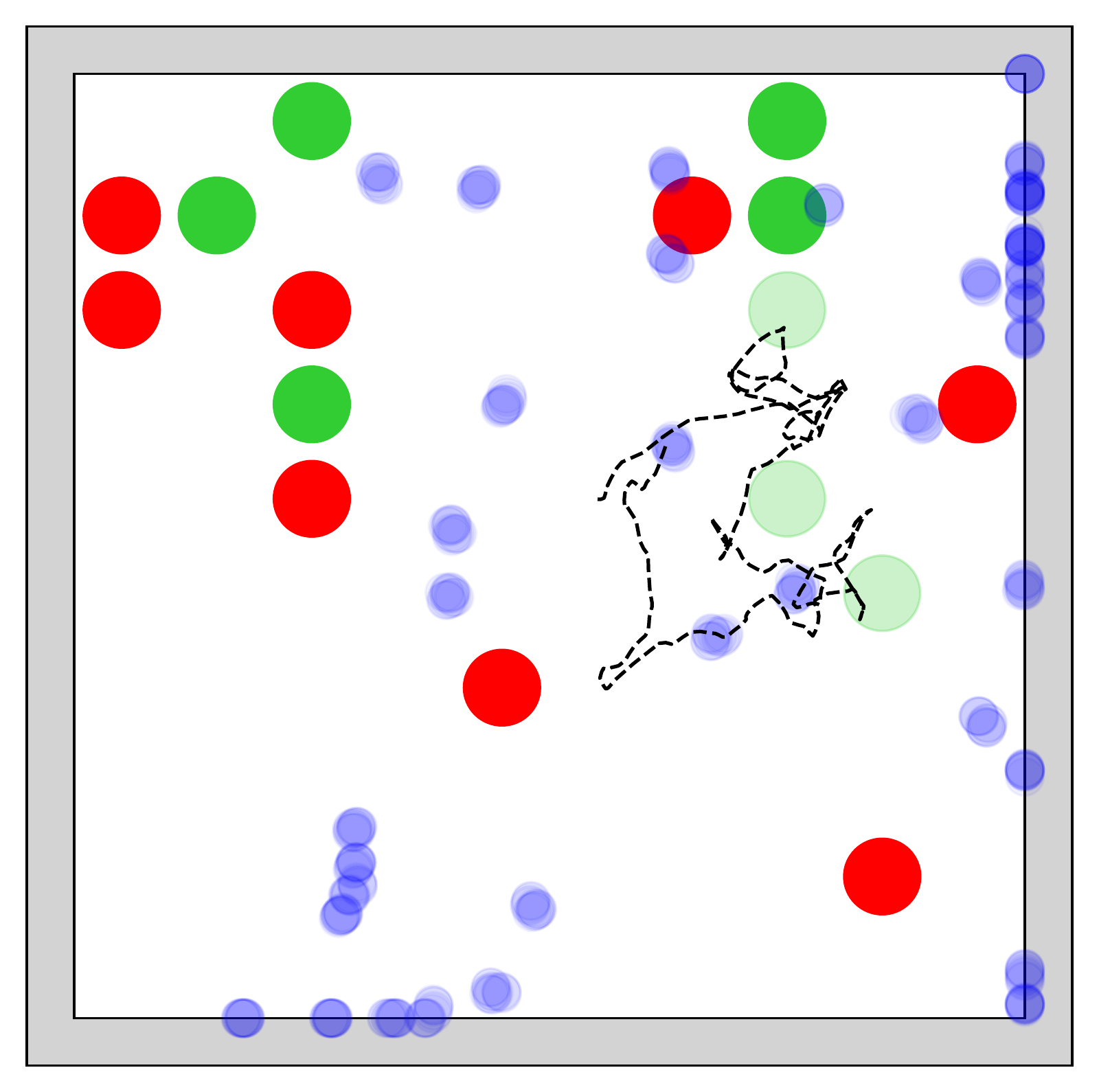}
    \caption{0\%} \label{fig:antgather-0p}
\end{subfigure}
\hfill
\begin{subfigure}[b]{0.15\textwidth}
    \centering
    \includegraphics[width=\linewidth]{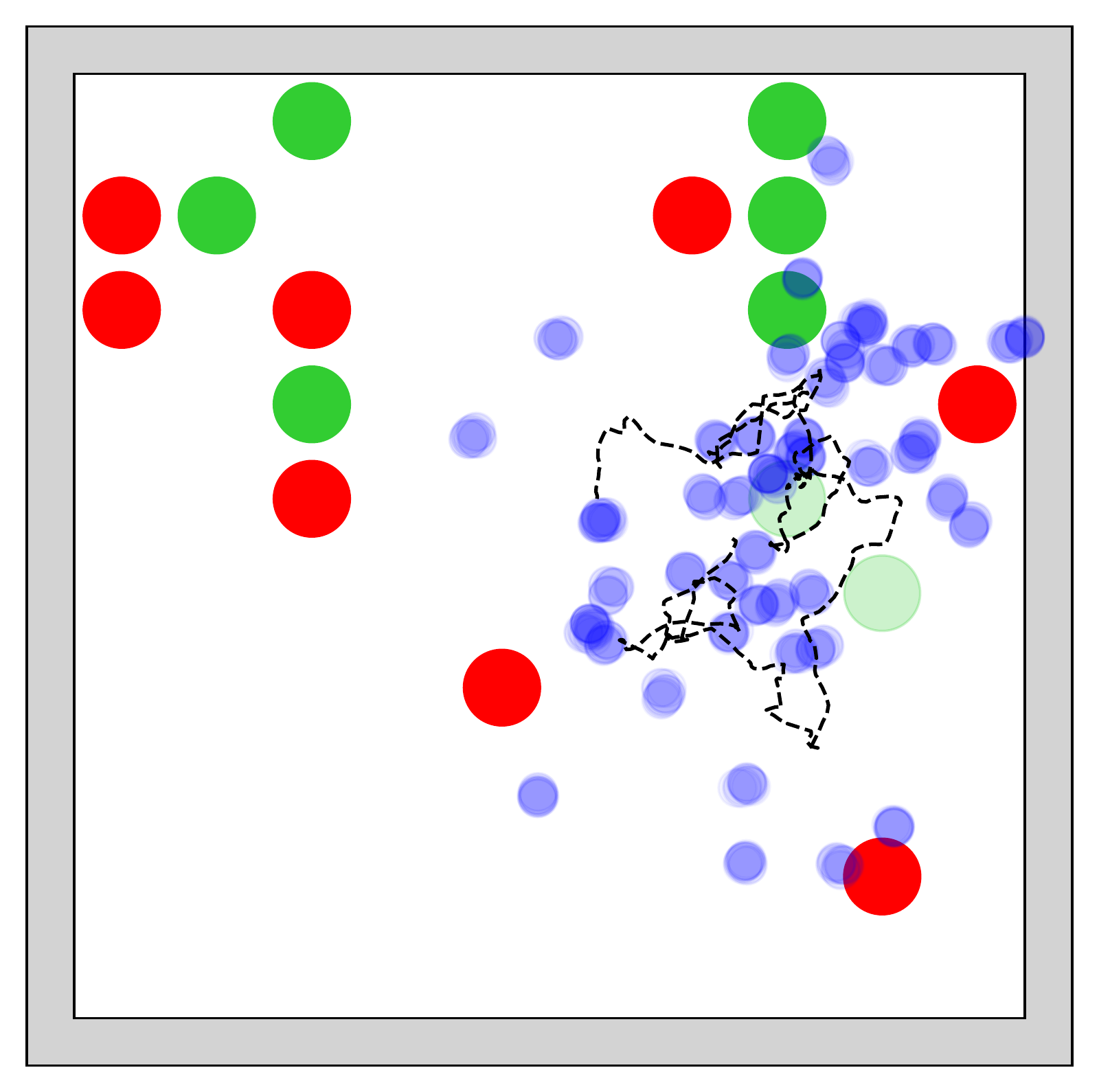}
    \caption{50\%} \label{fig:antgather-50p}
\end{subfigure}
\hfill
\begin{subfigure}[b]{0.15\textwidth}
    \centering
    \includegraphics[width=\linewidth]{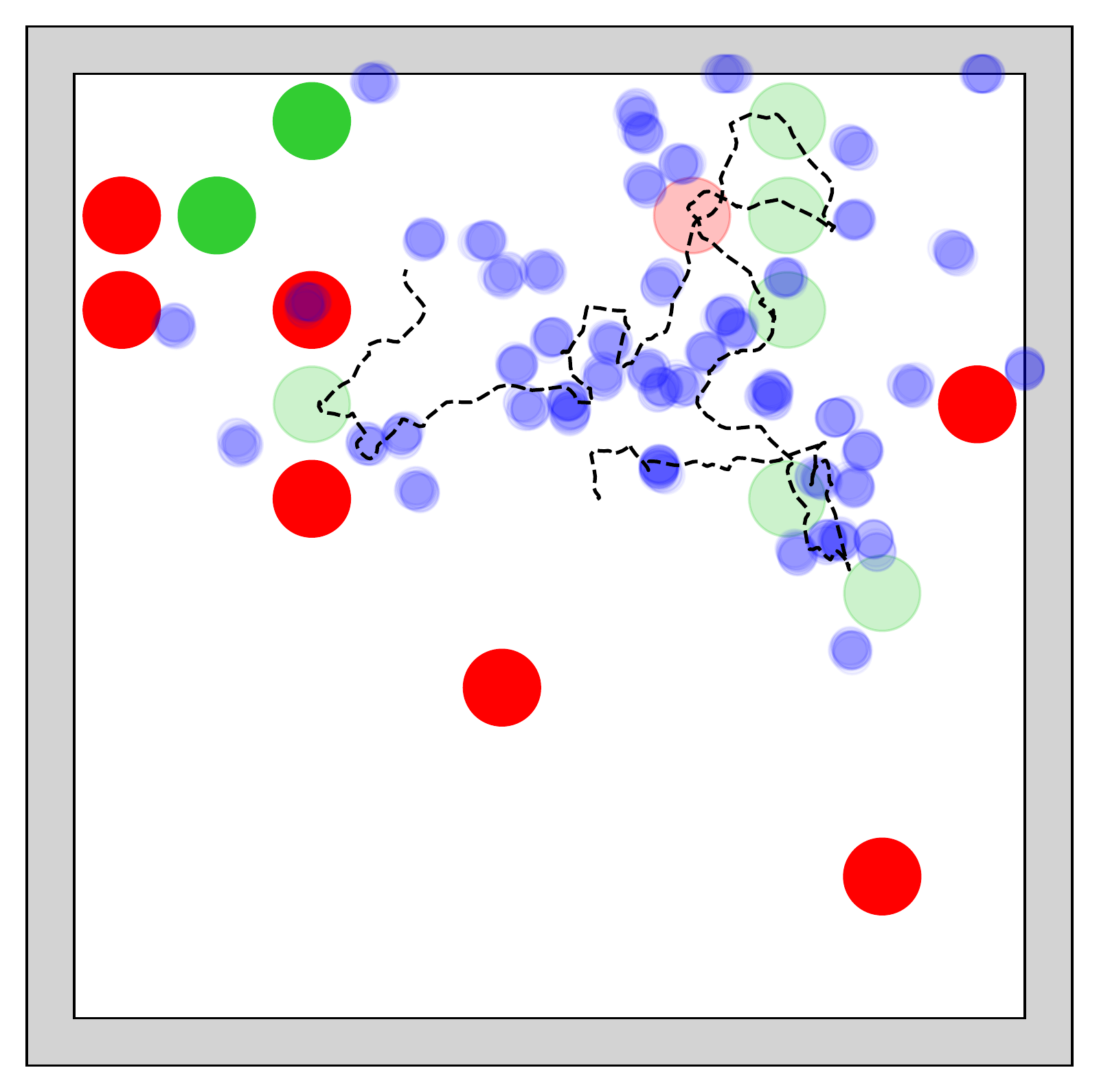}
    \caption{66.7\%} \label{fig:antgather-66.7p}
\end{subfigure}
\hfill
\begin{subfigure}[b]{0.15\textwidth}
    \centering
    \includegraphics[width=\linewidth]{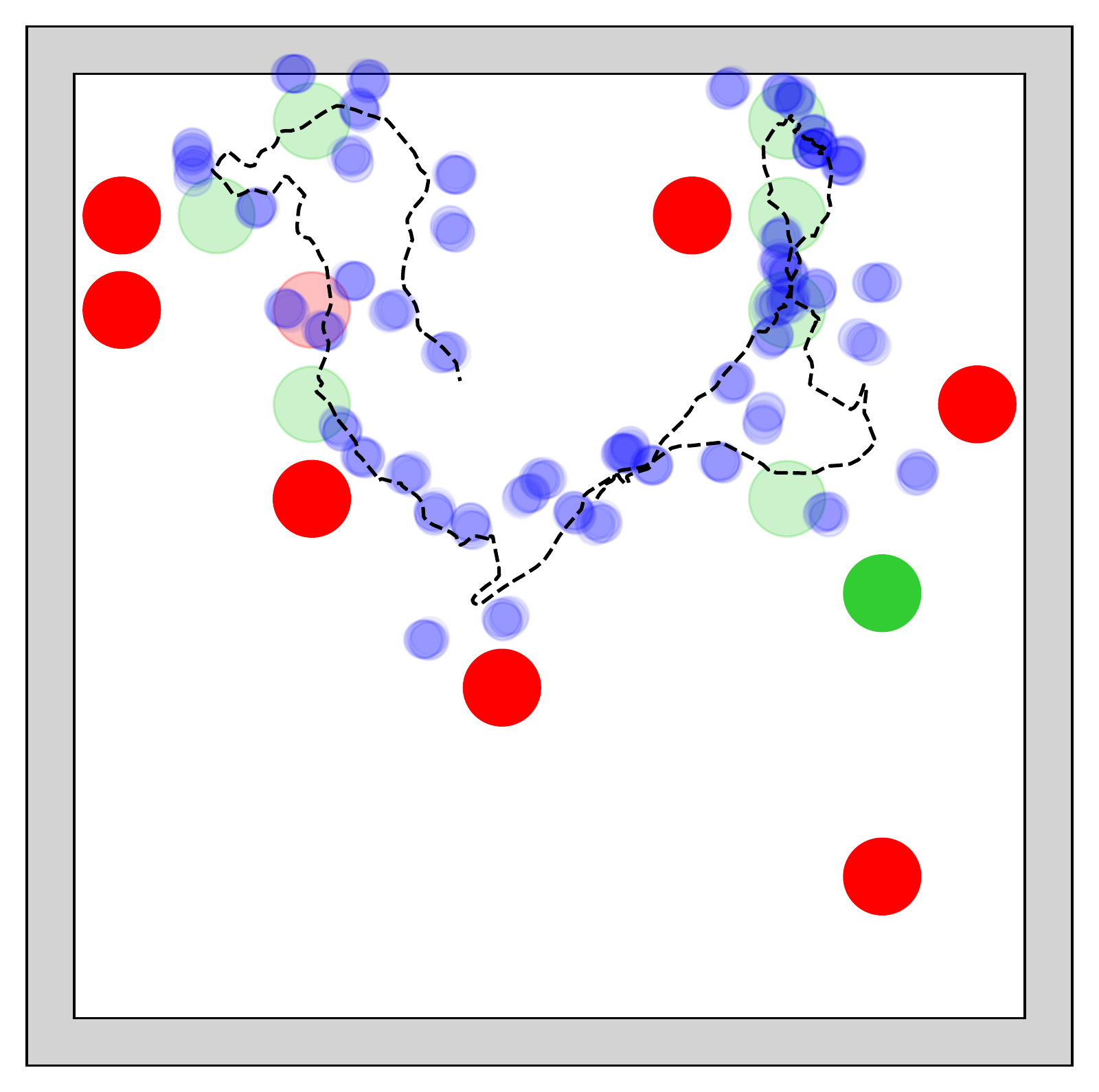}
    \caption{75\%} \label{fig:antgather-75p}
\end{subfigure}
\hfill
\begin{subfigure}[b]{0.15\textwidth}
    \centering
    \includegraphics[width=\linewidth]{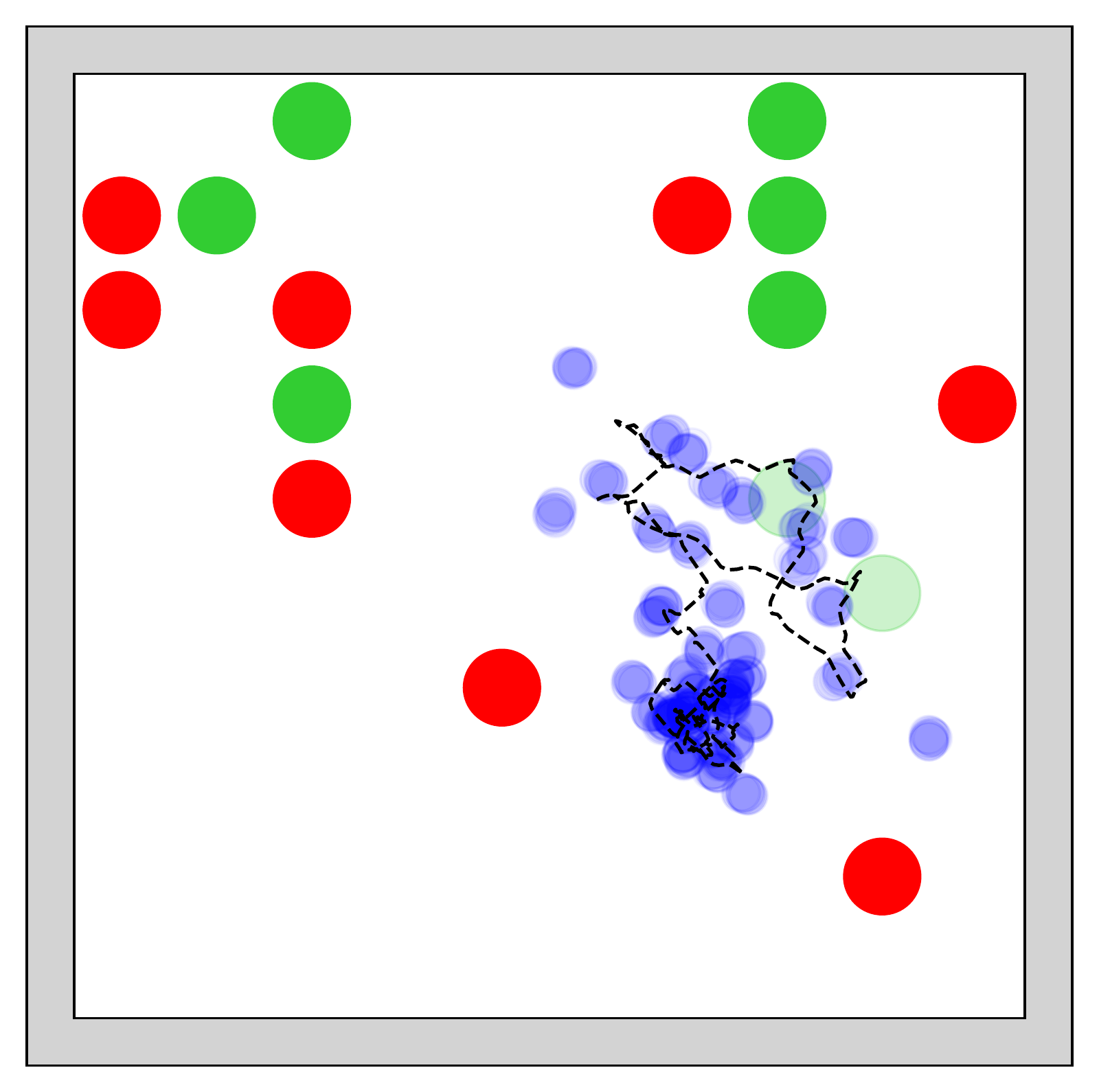}
    \caption{83.3\%} \label{fig:antgather-83.3p}
\end{subfigure}
\hfill
\begin{subfigure}[b]{0.15\textwidth}
    \centering
    \includegraphics[width=\linewidth]{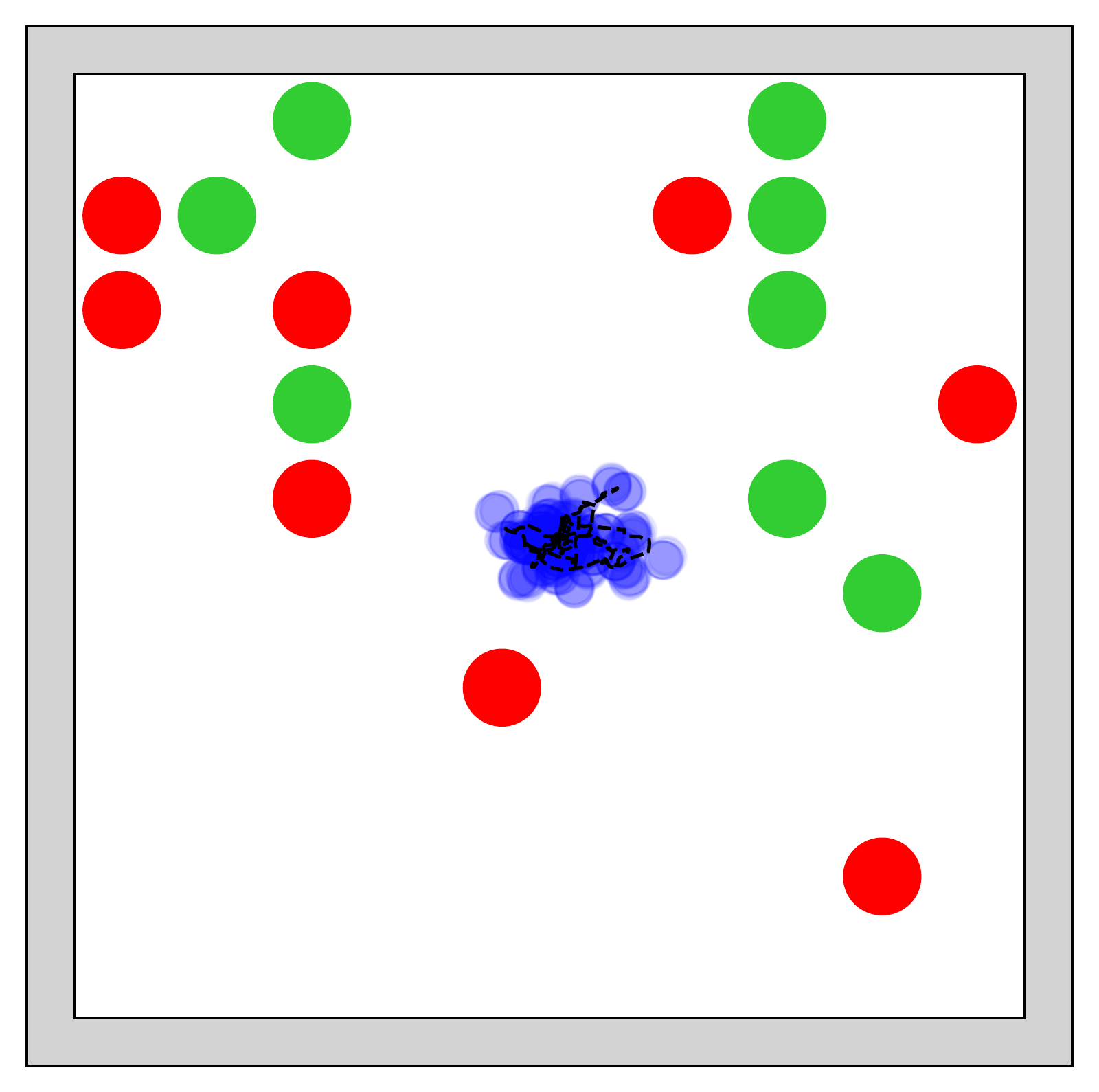}
    \caption{91.7\%} \label{fig:antgather-91.7p}
\end{subfigure}
\caption{Effect of varying degrees of cooperation for the \antgather environment. We plot the performance of the optimal policy and sample trajectories for various cooperation ratios as defined in Section~\ref{sec:cg-weight} (see the subcaptions for d-i). Increasing degrees of cooperation improve the agent's ability to assign desired goals that match the agent's trajectory, which subsequently improves the performance of the policy. Large degrees of cooperation, however, begin to disincentivize the agent from moving.}
\label{fig:effect-of-cg-weights}
\end{figure*}

\subsection{Cooperation tradeoff}  
\label{sec:cg-weight}
In this section, we explore the effects of varying degrees of cooperation on the performance of the resulting policy. Figure~\ref{fig:effect-of-cg-weights} depicts the effect of increasing cooperative penalties on the resulting policy in the \antgather environment. 
%
In 
this 
task, we notice that expected intrinsic returns for the worker in the standard HRL approach converge to a value of about $-600$ (see Figure~\ref{fig:cg-delta-worker-qvals}). As a result, we promote cooperation by assigning values of $\delta$
that are progressively larger than $-600$ to determine what level of cooperation leads to the best performance.

Figures~\ref{fig:cg-delta-rewards}~to~\ref{fig:antgather-91.7p} depict the effect of varying levels of cooperation on the performance of the policy\footnote{We define the cooperative ratio in these figures as the ratio of the assigned $\delta$ constraint between the standard hierarchical approach and the maximum expected return. The intrinsic rewards used here are non-positive meaning that the largest expected return is $0$. For example, a $\delta$ value of $-450$ is equated to a cooperation ratio of $(-600 + 450) / (-600 - 0) = 0.25$, or $25\%$.}.
As expected, we see that as the level of cooperation increases, the goal-assignment behaviors increasing consolidate near the path of the agent. This produces optimal behaviors in this setting for cooperation levels in the vicinity of 75\% (see Figures~\ref{fig:cg-delta-rewards}~and~\ref{fig:antgather-75p}). For levels of cooperation nearing 100\%, however, this consolidation begins to disincentivize forward movement, and subsequently exploration, by assignment goals that align with the current position of the agent (Figure~\ref{fig:antgather-91.7p}), thereby deteriorating the overall performance of the policy.

Figures~\ref{fig:cg-delta-worker-qvals}~and~\ref{fig:cg-delta-lambda} depict the agent's ability to achieve the assigned $\delta$ constraint and the dynamic $\lambda$ terms assigned to achieve these constraints, respectively. For most choices of $\delta$, \methodName succeeds in defining dynamic $\lambda$ values that match the constraint, demonstrating the efficacy of the designed optimization procedure. For very large constraints, in this case for $\delta=-50$, no choice of $\lambda$ can be assigned to match the desired constraint, thereby causing the value of $\lambda$ to grow exponentially. This explosion in the relevance of the constraint term likely obfuscates the relevance of the environment expected return to the manager gradients, resulting in the aforementioned disincentive for forward movement.

\begin{figure}
    \centering
    \begin{subfigure}[b]{\textwidth}
    \begin{tikzpicture}
        \newcommand \boxwidth {1.0}
        \newcommand \boxheight {\textwidth}
        \newcommand \boxoffset {0}
    
        \draw (0, -0.15) node {};
    
        \draw [color2] 
        (0.33 * \textwidth, \boxwidth + \boxoffset - 0.5) -- 
        (0.39 * \textwidth, \boxwidth + \boxoffset - 0.5);
        \draw [anchor=west]
        (0.40 * \textwidth, \boxwidth + \boxoffset - 0.5) node {\scriptsize HRL worker};

        \draw [color5] 
        (0.56 * \textwidth, \boxwidth + \boxoffset - 0.5) -- 
        (0.62 * \textwidth, \boxwidth + \boxoffset - 0.5);
        \draw [anchor=west]
        (0.63 * \textwidth, \boxwidth + \boxoffset - 0.5) node {\scriptsize \methodName worker};
    
        \draw [color3] 
        (0.79 * \textwidth, \boxwidth + \boxoffset - 0.5) -- 
        (0.85 * \textwidth, \boxwidth + \boxoffset - 0.5); 
        \draw [anchor=west]
        (0.86 * \textwidth, \boxwidth + \boxoffset - 0.5) node {\scriptsize HIRO worker};
    \end{tikzpicture}
    \end{subfigure} \\ \vspace{-0.5cm}    \begin{subfigure}[c]{0.36\textwidth}
        \begin{subfigure}[c]{0.31\textwidth}
            \includegraphics[height=1.9cm]{figures/AntGather-env.png}
        \end{subfigure}
        \begin{subfigure}[c]{0.31\textwidth}
            \begin{tikzpicture}
                \draw (1, 0.15) node {\scriptsize Transfer $\pi_w$};
                \draw (1, -0.15) node {\scriptsize Retrain $\pi_m$};
                \draw [->] (0.4, -0.5) -- (1.75, -1.1);
                \draw [->] (0.4, 0.5) -- (1.75, 1.1);
            \end{tikzpicture}
        \end{subfigure}
        \begin{subfigure}[c]{0.32\textwidth}
            \includegraphics[height=1.9cm]{figures/AntMaze-env.png} \\[10pt]
            \includegraphics[height=1.9cm]{figures/AntFourRooms-env.png}
        \end{subfigure}
    \end{subfigure}
    \ \
    \begin{subfigure}[c]{0.17\textwidth}
        \caption*{\ \ \ \ \ \ \ [16,0]}
        \includegraphics[height=2cm]{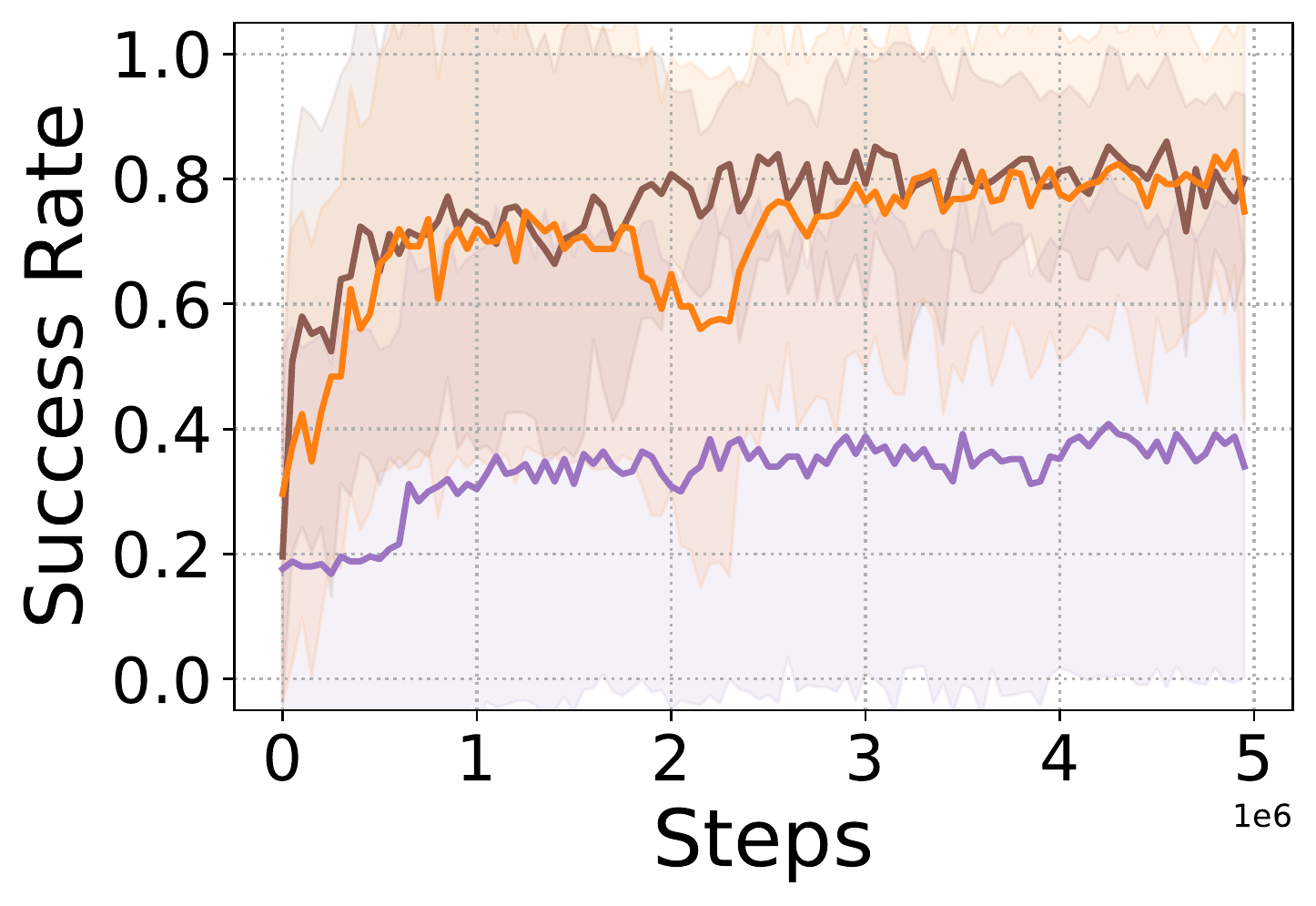}
        \caption*{\ \ \ \ \ \ \ [20,0]}
        \includegraphics[height=2cm]{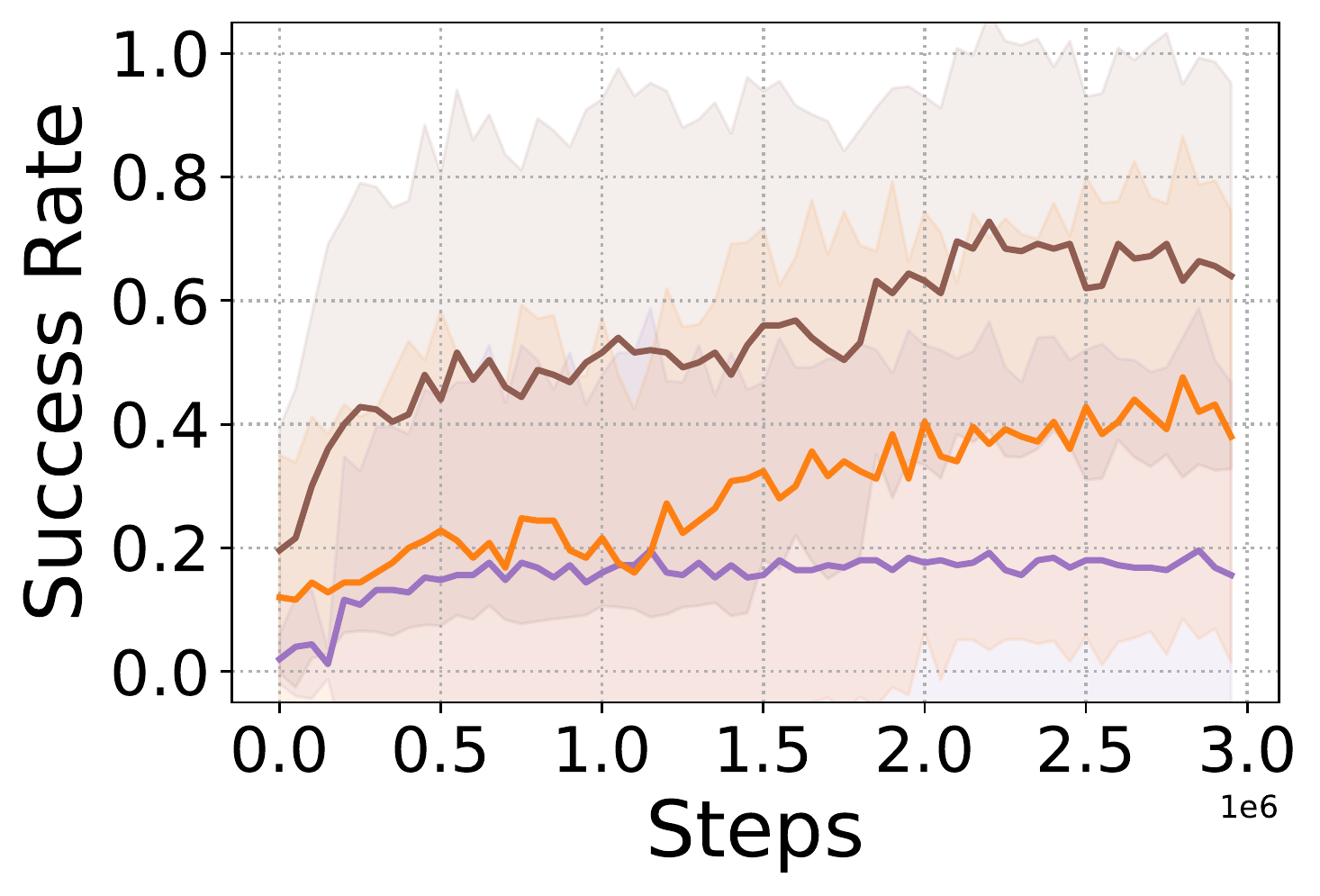}
    \end{subfigure}
    \quad
    \begin{subfigure}[c]{0.17\textwidth}
        \caption*{\ \ \ \ \ \ \ [16,16]}
        \includegraphics[height=2cm]{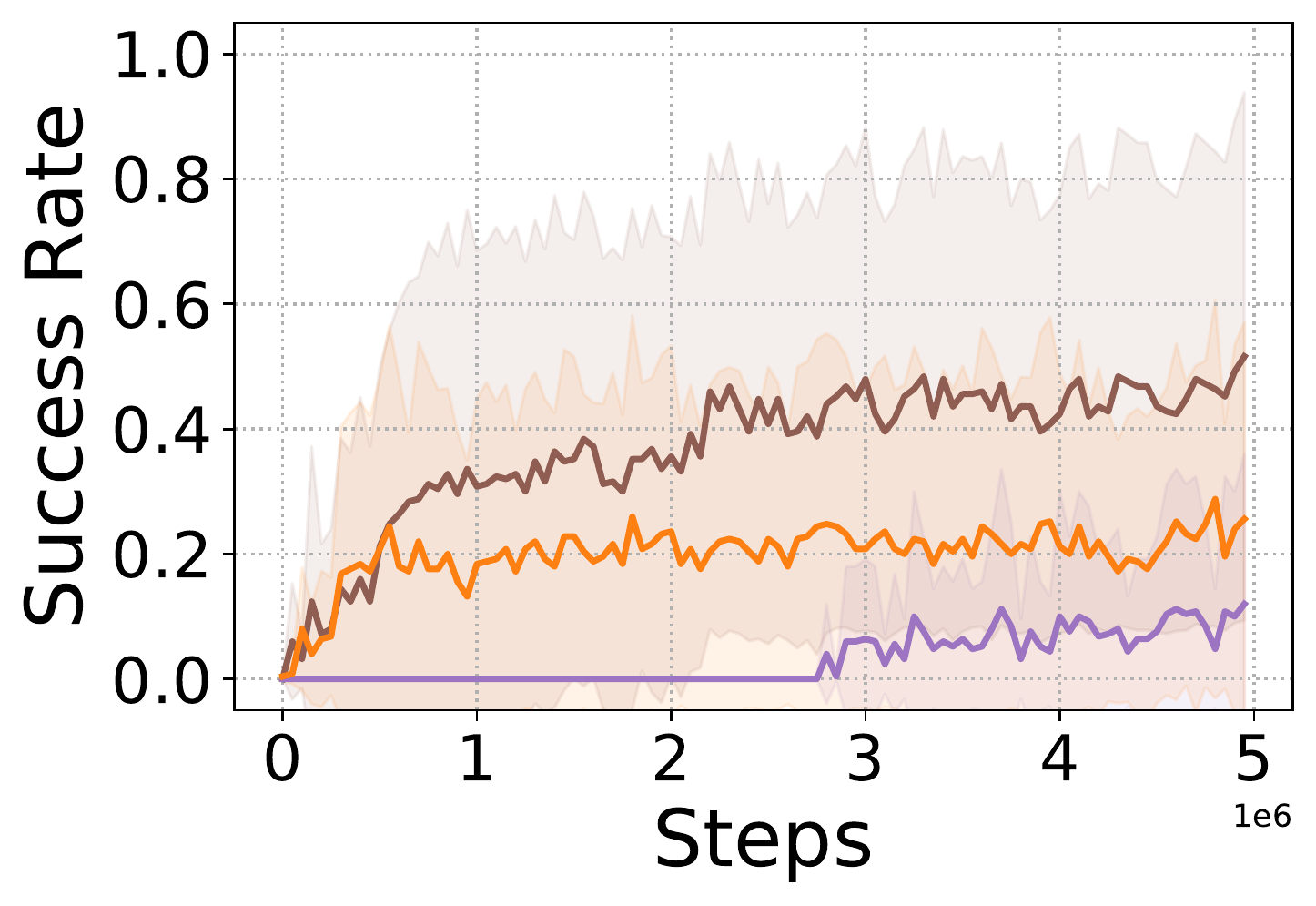}
        \caption*{\ \ \ \ \ \ \ [0,20]}
        \includegraphics[height=2cm]{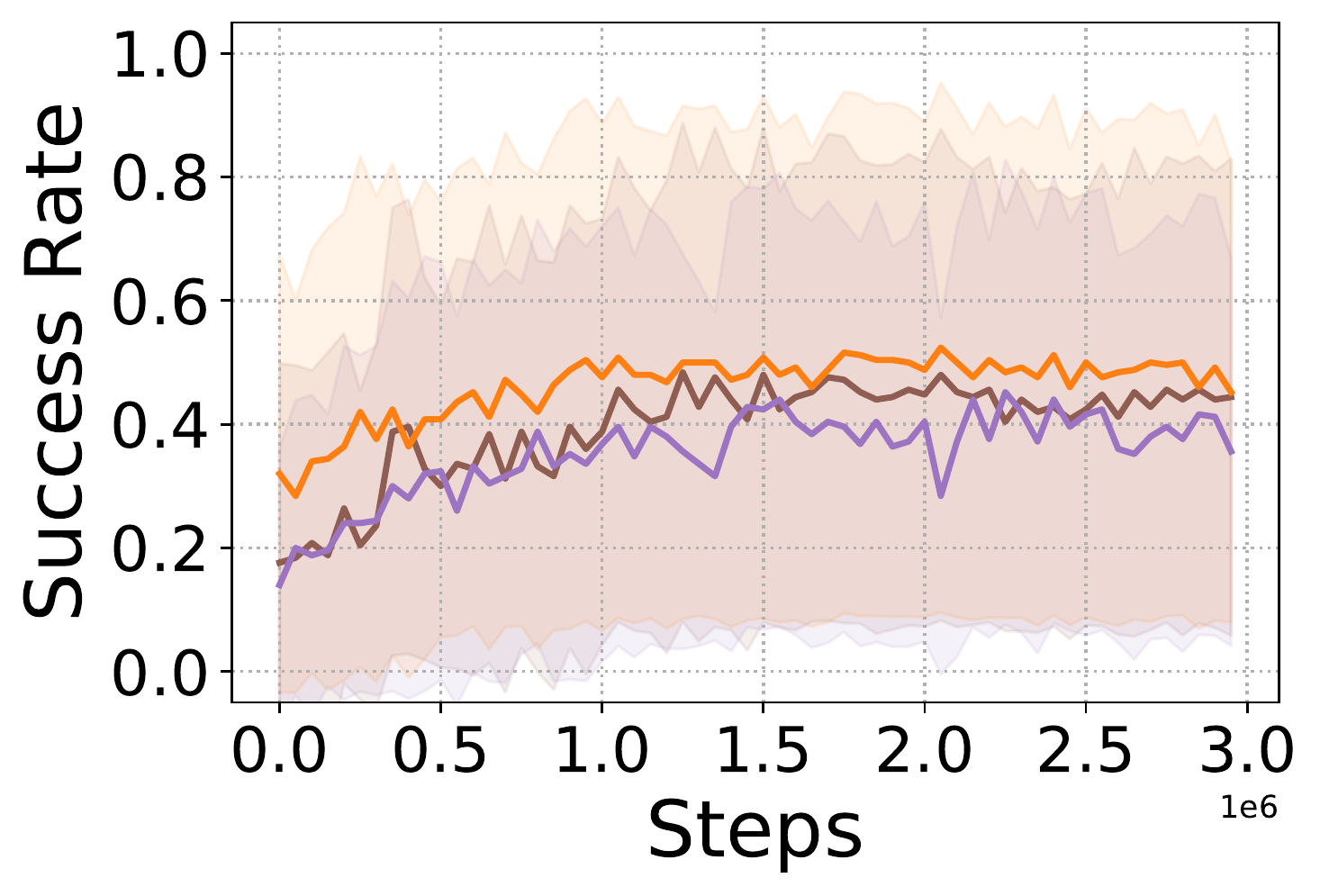}
    \end{subfigure}
    \quad
    \begin{subfigure}[c]{0.17\textwidth}
        \caption*{\ \ \ \ \ \ \ [0,16]}
        \includegraphics[height=2cm]{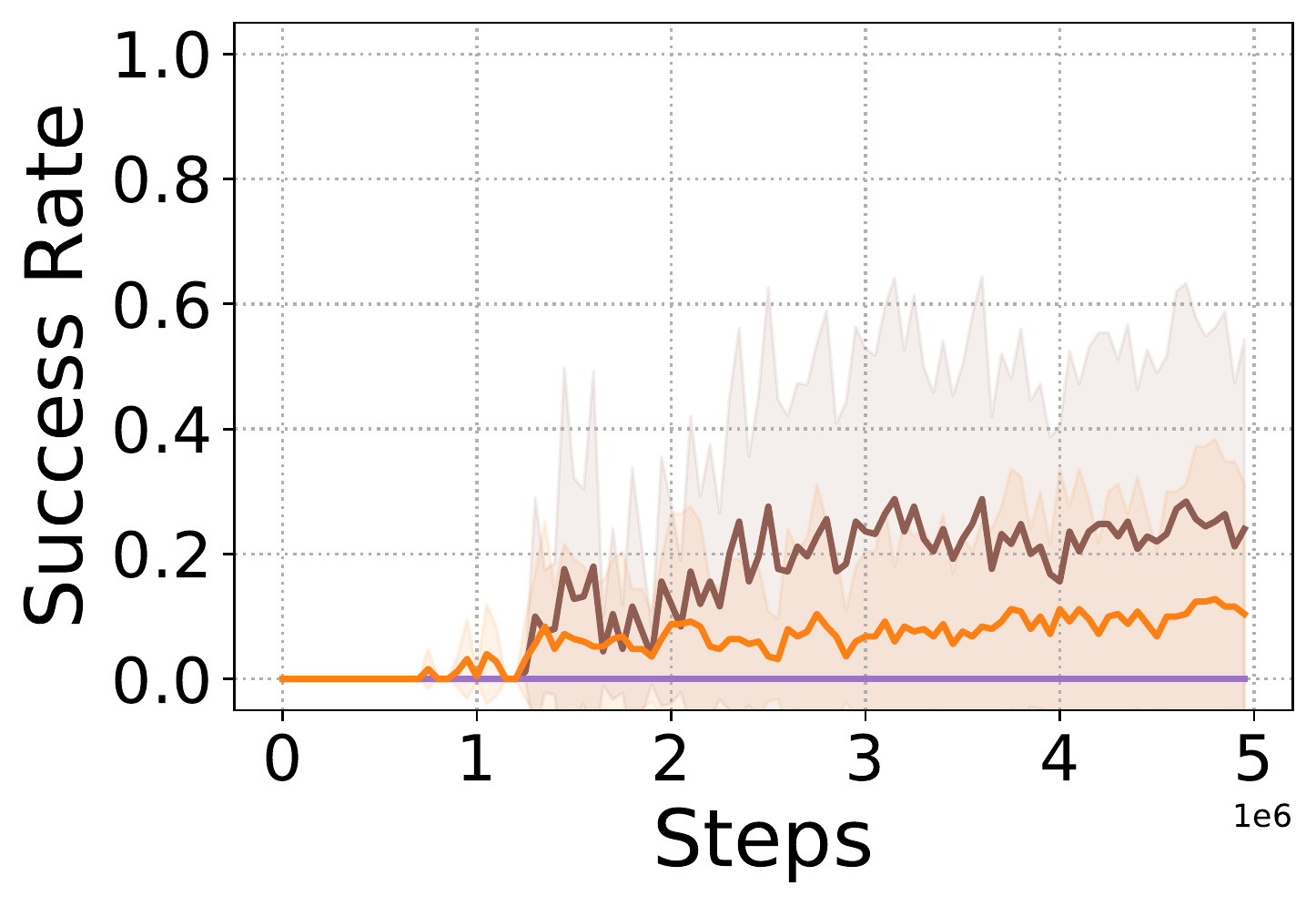}
        \caption*{\ \ \ \ \ \ \ [20,20]}
        \includegraphics[height=2cm]{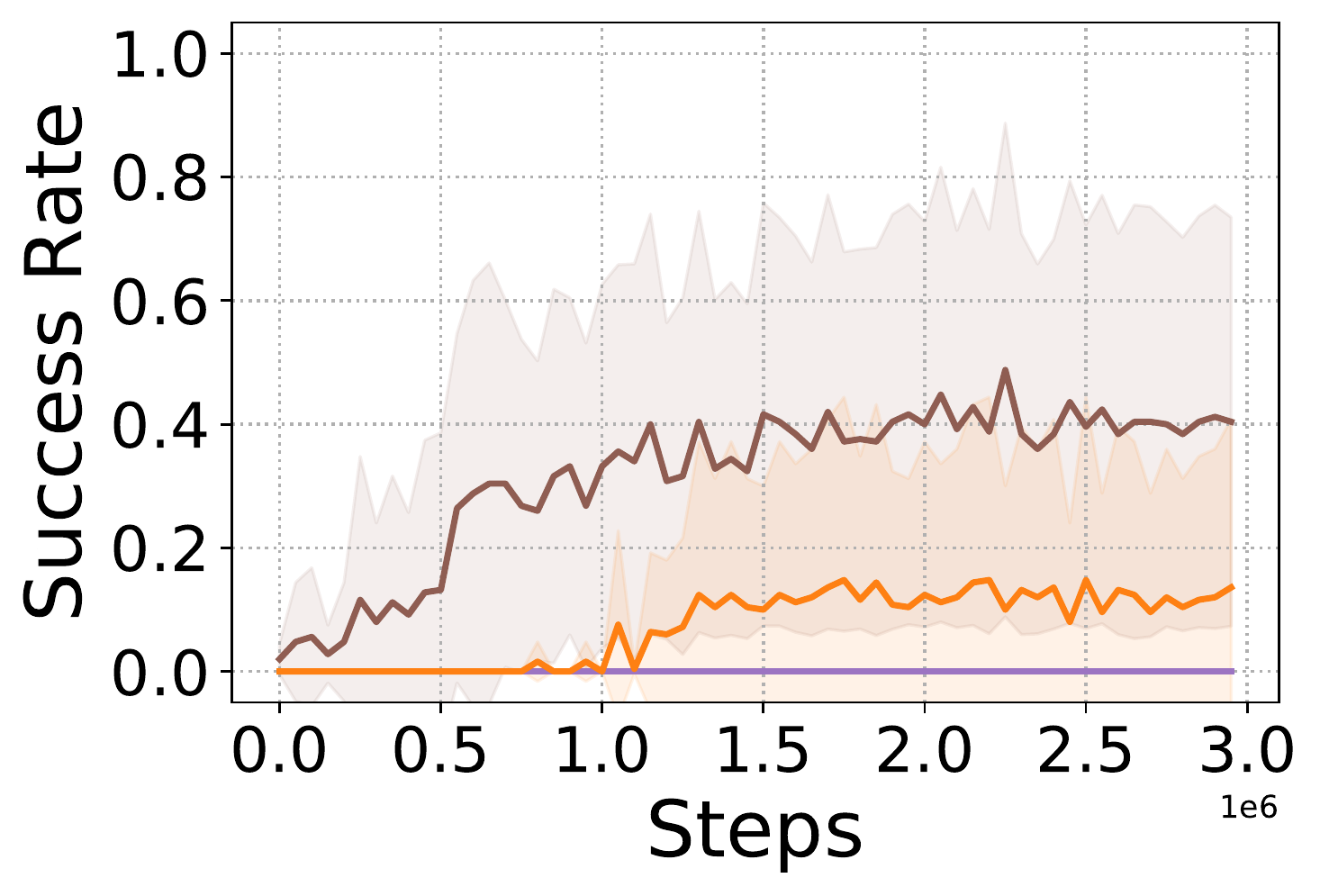}
    \end{subfigure}
    \caption{An illustration of the transferability of policies learned via \methodName. A policy is trained in the \antgather environment and the worker policy is frozen and transferred to the \antmaze and \antfourrooms environments. The policies learned when utilizing the \methodName worker policy significantly outperform the HIRO policy for a wide variety of evaluation points, highlighting the benefit of \methodName in learning a more informative and generalizable policy representations.}
    \label{fig:transfer}
\end{figure}

\subsection{Transferability of policies between tasks}

Finally, we explore the effects of promoting inter-level cooperation on the transferability of learned policies to different tasks. To study this, we look to the Ant environments in Figure~\ref{fig:envs} and choose to learn a policy in one environment (\antgather) and transfer the learned worker policy to two separate environments (\antmaze and \antfourrooms). The initial policy within the \antgather environment is trained for 1 million samples as in Figure~\ref{fig:antgather-rewards} utilizing either the HRL, HIRO, or \methodName algorithms. To highlight the transferability of the learned policy, we fix the weights of the worker policy, and instead attempt to learn a new manager policy for the given task; this allows us to identify whether the original policy generalizes better in the zero-shot setting. We still, however, must learn a new higher-level policy as the observations and objectives for the manager differ between problems.

Figure~\ref{fig:transfer} depicts the transfer setup and performance of the policy for a set number of evaluation points. We find that worker policies learned via \methodName significantly improve the efficacy of the overall agent when exposed to new tasks. This suggests that the policies learned via more structured and informative goal-assignment procedures result in policies that are more robust to varying goal-assignment strategies, and as such allow the learned behaviors to be more task-agnostic.


\section{Conclusions and future work} \label{sec:conclusions}

In this work, we propose connections between multi-agent and hierarchical reinforcement learning that motivates our novel method of inducing cooperation in hierarchies. We provide a derivation of the gradient of a manager policy with respect to its workers for an actor-critic formulation as well as introducing a $\lambda$ weighting term for this gradient which controls the level of cooperation.
We find that using \methodName results in consistently better-performing policies, that have lower empirical non-stationarity than prior work, particularly for more difficult tasks.

Next, we find that policies learned with a fixed $\lambda$ term are at times highly sensitive to the choice of value, and accordingly derive a dynamic variant of the cooperative gradients that automatically updates the value of $\lambda$, balancing goal exploration. We demonstrate that this dynamic variant further expands the scope of solvable tasks, in particular allowing us to generate highly effective mixed-autonomy driving behaviors in a very sample-efficient manner.

For future work, we would like to apply this method to discrete action environments and additional hierarchical models such as the options framework. Potential future work also includes extending the cooperative HRL formulation to multi-level hierarchies, in which the multi-agent nature of hierarchical training is likely to be increasingly detrimental to training stability.




\bibliography{cooperative-hrl}


\newpage

\appendix

\section{Derivation of cooperative manager gradients} \label{sec:derivation}

In this section, we derive an analytic expression of the gradient of the manager policy in a two-level goal-conditioned hierarchy with respect to both the losses associated with the high level and low level policies. In mathematical terms, we are trying to derive an expression for the weighted summation of the derivation of both losses, expressed as follows:
\begin{equation}
    \nabla_{\theta_m} J_m' = \nabla_{\theta_m} \left( J_m + \lambda J_w \right) = \nabla_{\theta_m} J_m + \lambda \nabla_{\theta_m} J_w
\end{equation}
where $\lambda$ is a weighting term and $J_m$ and $J_w$ are the expected returns assigned to the manager and worker policies, respectively. More specifically, these two terms are:
\begin{equation}
    \resizebox{!}{12pt}{$
    J_m = \mathbb{E}_{s\sim p_\pi} \left[ \sum_{t=0}^{T/k} \gamma^t r_m(s_{kt}) \right] = \int_{\mathcal{S}} \rho_0(s_t) V_m(s_t) ds_t
    $}
\end{equation}\\[-25pt]
\begin{equation}
    \resizebox{!}{12pt}{$
    J_w = \mathbb{E}_{s\sim p_\pi} \left[ \sum_{t=0}^k \gamma^t r_w(s_t, g_t,\pi_w(s_t,g_t)) \right] = \int_{\mathcal{S}} \rho_0(s_t) V_w(s_t, g_t) ds_t 
    $}
\end{equation}
Here, under the actor-critic formulation we replace the expected return under a given starting state with the value functions $V_m$ and $V_w$ This is integrated over the distribution of initial states $\rho_0(\cdot)$.

Following the results by \citet{silver2014deterministic}, we can express the first term in Eq.~\eqref{eq:connected-gradient} as:
\begin{equation}
    \nabla_{\theta_m} J_m = \mathbb{E}_{s\sim p_\pi} \left[ \nabla_a Q_m (s,a)|_{a=\pi_m(s)}\nabla_{\theta_m} \pi_m(s) \right]
\end{equation}

We now expand the second term of the gradient into a function of the manager and worker actor ($\pi_m$, $\pi_w$) and critic ($Q_m$, $Q_w$) policies and their trainable parameters. In order to propagate the loss associated with the worker through the policy parameters of the manager, we assume that the goals assigned to the worker $g_t$ are not fixed variables, but rather temporally abstracted outputs from the manager policy $\pi_m$, and may be updated in between decisions by the manager via a transition function $h$. Mathematically, the goal transition is defined as: 
\begin{equation}
    g_t(\theta_m) = 
    \begin{cases}
        \pi_m(s_t) & \text{if } t \text{ mod } k = 0 \\
        h(s_{t-1}, g_{t-1}(\theta_m), s_t) & \text{otherwise}
    \end{cases}
\end{equation}
For the purposes of simplicity, we express the manager output term as $g_t$ from now on.

We begin by computing the partial derivative of the worker value function with respect to the parameters of the manager:
\begin{equation}
    \resizebox{.9\hsize}{!}{$
    \begin{aligned}
        \nabla_{\theta_m} V_w(s_t, g_t) &=\nabla_{\theta_m} Q_w (s_t, g_t, \pi_w(s_t, g_t)) \\
        &= \nabla_{\theta_m} \bigg( r_w(s_t, g_t, \pi_w(s_t,g_t)) +\int_{\mathcal{G}} \int_{\mathcal{S}} \gamma p_w(s', g'| s_t,g_t, \pi_w(s_t,g_t)) V_w(s',g')ds'dg' \bigg) \\
        &= \nabla_{\theta_m} r_w(s_t,g_t,\pi_w(s_t,g_t)) + \gamma \nabla_{\theta_m} \int_{\mathcal{G}}\int_{\mathcal{S}} p_w(s',g'| s_t, g_t, \pi_w(s_t,g_t)) V_w(s',g')ds'dg'
    \end{aligned}
    $}
    \label{eq:gradient_p1}
\end{equation}
where $\mathcal{G}$ and $\mathcal{S}$ are the goal and environment state spaces, respectively, and $p_w(\cdot, \cdot | \cdot, \cdot, \cdot)$ is the probability distribution of the next state from the perspective of the worker given the current state and action.

Expanding the latter term, we get:
\begin{equation}
    \begin{aligned}
    &p_w(s',g'|s_t,g_t,\pi_w(s_t,g_t)) = p_{w,1} (g'| s', s_t,g_t,\pi_w(s_t,g_t)) p_{w,2} (s'| s_t,g_t,\pi_w(s_t,g_t))
    \end{aligned}
    \label{eq:pw_decompose}
\end{equation}
The first element, $p_{w1}$, is the probability distribution of the next goal, and is deterministic with respect to the conditional variables. Specifically:
\begin{equation}
    p_{w,1} (g'| s_t,g_t,\pi_w(s_t,g_t)) = 
    \begin{cases}
        1 & \text{if } g' = g_{t+1} \\
        0 & \text{otherwise}
    \end{cases}
    \label{eq:pw1}
\end{equation}

The second element, $p_{w,2}$, is the state transition probability from the MDP formulation of the task, i.e.
\begin{equation}
    p_{w,2}(s'| s_t,g_t,\pi_w(s_t,g_t)) = p (s'| s_t,\pi_w(s_t,g_t))
    \label{eq:pw2}
\end{equation}

Combining Eq.~\eqref{eq:pw_decompose}-\eqref{eq:pw2} into Eq.~\eqref{eq:gradient_p1}, we get:
\begin{equation} \label{eq:simplified-next-step-value}
    \resizebox{.9\hsize}{!}{$
    \begin{aligned}
        \nabla_{\theta_m} V_w(s_t,g_t) &=\nabla_{\theta_m} r_w(s_t,g_t,\pi_w(s_t,g_t)) \\
        &\quad + \gamma \nabla_{\theta_m} \int_{\mathcal{G}}\int_{\mathcal{S}}\bigg( p_{w,1} (g'| s', s_t,g_t,\pi_w(s_t,g_t)) p_{w,2} (s'| s_t,g_t,\pi_w(s_t,g_t)) V_w(s',g') ds'dg'\bigg) \\
        &= \nabla_{\theta_m} r_w(s_t,g_t,\pi_w(s_t,g_t)) \\
        &\quad + \gamma \nabla_{\theta_m} \int_{\mathcal{G}\cap \{g_{t+1}\}}\int_{\mathcal{S}} 1 \cdot p (s'| s_t,\pi_w(s_t,g_t)) V_w(s',g') ds'dg' \\
        &\quad + \gamma \nabla_{\theta_m} \int_{(\mathcal{G}\cap \{g_{t+1}\})^c}\int_{\mathcal{S}} 0 \cdot p (s'| s_t,\pi_w(s_t,g_t)) V_w(s',g') ds'dg' \\
        &= \nabla_{\theta_m} r_w(s_t,g_t,\pi_w(s_t,g_t)) + \gamma \nabla_{\theta_m} \int_{\mathcal{S}} p(s'| s_t,\pi_w(s_t,g_t)) V_w(s',g_{t+1})ds'
    \end{aligned}
    $}
\end{equation}

Continuing the derivation of $\nabla_{\theta_m}V_w$ from Eq.~\eqref{eq:simplified-next-step-value}, we get,
\begin{equation} \label{eq:continue-derivatione}
    \resizebox{.9\hsize}{!}{$
    \begin{aligned}
        \nabla_{\theta_m} V_w(s_t,g_t) &= \nabla_{\theta_m} r_w(s_t,g_t,\pi_w(s_t,g_t)) +\gamma \nabla_{\theta_m} \int_{\mathcal{S}} p(s'| s_t,\pi_w(s_t,g_t)) V_w(g_{t+1}, s')ds' \\
        &= \nabla_{\theta_m} r_w(s_t,g_t,\pi_w(s_t,g_t)) +\gamma \int_{\mathcal{S}} \nabla_{\theta_m} p(s'| s_t,\pi_w(s_t,g_t)) V_w(g_{t+1}, s')ds' \\
        &= \nabla_{\theta_m} g_t \nabla_g r_w(s_t,g,\pi_w(s_t,g_t))|_{g=g_t} \\
        &\quad + \nabla_{\theta_m}g_t \nabla_g \pi_w (s_t,g)|_{g=g_t} \nabla_a r_w(s_t,g_t,a)|_{a=\pi_w(s_t,g_t)} \\
        &\quad +\gamma\int_\mathcal{S} \bigg(V_w(s',g_{t+1})\nabla_{\theta_m} g_t \nabla_g \pi_w(s_t,g)|_{g=g_t} \nabla_a p(s'\vert s_t,a)|_{a=\pi_w(s_t,g_t)}ds'\bigg)\\
        &\quad +\gamma\int_\mathcal{S}p(s'\vert s_t,\pi_w(s_t,g_t))\nabla_{\theta_m} V_w(s',g_{t+1}) ds'\\
        &= \nabla_{\theta_m} g_t \nabla_g \bigg(r_w(s_t,g,\pi_w(s_t,g_t)) \\
        &\quad \quad \quad \quad \quad \quad + \pi_w (s_t,g) \nabla_a r_w(s_t,g_t,a)|_{a=\pi_w(s_t,g_t)} \vphantom{\int} \\
        &\quad \quad \quad \quad \quad \quad + \gamma\int_\mathcal{S} V_w(s',g_{t+1}) \pi_w(s_t,g) \nabla_a p(s'\vert s_t,a)|_{a=\pi_w(s_t,g_t)}ds' \bigg) \bigg\rvert_{g=g_t}\\
        &\quad +\gamma\int_\mathcal{S}p(s'\vert s_t,\pi_w(s_t,g_t))\nabla_{\theta_m} V_w(s',g_{t+1}) ds'\\
        &= \nabla_{\theta_m} g_t \nabla_g \bigg(r_w(s_t,g,\pi_w(s_t,g_t)) \\
        &\quad \quad \quad \quad \quad \quad + \pi_w (s_t,g) \nabla_a \bigg( r_w(s_t,g_t,a) + \gamma\int_\mathcal{S} V_w(s',g_{t+1}) p(s'\vert s_t,a)ds' \bigg)\bigg\rvert_{a=\pi_w(s_t,g_t)} \bigg) \bigg\rvert_{g=g_t}\\
        &\quad + \gamma\int_\mathcal{S}p(s'\vert s_t,\pi_w(g_t, s_t))\nabla_{\theta_m} V_w(s',g_{t+1}) ds'\\
        &= \nabla_{\theta_m} g_t \nabla_g \bigg(r_w(s_t,g,\pi_w(s_t,g_t)) + \pi_w (s_t,g) \nabla_a Q_w(s_t,g_t,a)|_{a=\pi_w(s_t,g_t)}\vphantom{\int} \bigg) \bigg\rvert_{g=g_t}
        \\
        &\quad + \gamma\int_\mathcal{S}p(s'\vert s_t,\pi_w(s_t,g_t))\nabla_{\theta_m} V_w(s',g_{t+1}) ds'
    \end{aligned}
    $}
\end{equation} 

Iterating this formula, we have,
\begin{equation}
    \resizebox{.9\hsize}{!}{$
     \begin{aligned}
        \nabla_{\theta_m} V_w(s_t,g_t) &= \nabla_{\theta_m} g_t \nabla_g \bigg(r_w(s_t,g,\pi_w(s_t,g_t)) + \pi_w (s_t,g) \nabla_a Q_w(s_t,g_t,a)|_{a=\pi_w(s_t,g_t)}\vphantom{\int} \bigg) \bigg\rvert_{g=g_t}\\
        &\quad +\gamma\int_\mathcal{S}p(s_{t+1}\vert s_t,\pi_w(s_t,g_t))\nabla_{\theta_m} V_w(s_{t+1},g_{t+1}) ds_{t+1} \\
        &= \nabla_{\theta_m} g_t \nabla_g \bigg(r_w(s_t,g,\pi_w(s_t,g_t)) + \pi_w (s_t,g) \nabla_a Q_w(s_t,g_t,a)|_{a=\pi_w(s_t,g_t)}\vphantom{\int} \bigg) \bigg\rvert_{g=g_t}
        \quad \\
        &\quad +\gamma\int_\mathcal{S}p(s_{t+1}\vert s_t,\pi_w(s_t,g_t)) \nabla_{\theta_m} g_{t+1} \nabla_g \bigg(r_w(s_{t+1},g,\pi_w(s_{t+1},g_{t+1})) \vphantom{\int} \\
        &\quad \quad \quad \quad \quad + \pi_w (s_{t+1},g) \nabla_a Q_w(s_{t+1},g_{t+1},a)|_{a=\pi_w(s_{t+1},g_{t+1})}\vphantom{\int} \bigg) \bigg\rvert_{g=g_{t+1}}ds_{t+1} \\
        & \quad +\gamma^2 \int_\mathcal{S}\int_\mathcal{S} \bigg( p(s_{t+1}\vert s_t,\pi_w(s_t,g_t)) p(s_{t+2}\vert s_{t+1},\pi_w(g_{t+1}, s_{t+1}))\\
        &\quad \quad \quad \quad \quad \quad \quad \nabla_{\theta_m} V_w(s_{t+2},g_{t+2}) ds_{t+2}   ds_{t+1} \bigg)\\
        & \hspace{45mm} \vdots\\
        &= \sum_{n=0}^{\infty} \gamma^n \underbrace{\int_\mathcal{S} \cdots \int_\mathcal{S}}_{n \text{ times}} \left(\prod_{k=0}^{n-1} p(s_{t+k+1}|s_{t+k},\pi_w(s_{t+k},g_{t+k})) \right) \\
        &\quad \quad \quad \quad \times \nabla_{\theta_m} g_{t+n} 
        \nabla_g \bigg(r_w(s_{t+n},g,\pi_w(s_{t+n},g_{t+n})) \\
        & \quad \quad \quad +\pi_w (s_{t+n},g) \nabla_a Q_w(s_{t+n},g_{t+n},a)|_{a=\pi_w(s_{t+n},g_{t+n})}\bigg)\vphantom{\int} \bigg) \bigg\rvert_{g=g_{t+n}} ds_{t+n}\cdots ds_{t+1}
    \end{aligned}
    $}
\end{equation}

Taking the gradient of the expected worker value function, we get,
\begin{small}
\begin{equation}
    \resizebox{.9\hsize}{!}{$
    \begin{aligned}
        \nabla_{\theta_m} J_w &= \nabla_{\theta_m} \int_{\mathcal{S}} \rho_0(s_0) V_w(s_0, g_0) ds_0 \\
        &= \int_{\mathcal{S}} \rho_0(s_0) \nabla_{\theta_m} V_w(s_0, g_0) ds_0 \\
        &= \int_{\mathcal{S}} \rho_0(s_0) \sum_{n=0}^{\infty} \gamma^n \underbrace{\int_\mathcal{S} \cdots \int_\mathcal{S}}_{n \text{ times}} \Bigg[\left(\prod_{k=0}^{n-1} p(s_{k+1}|s_k,\pi_w(s_k,g_k)) \right) \nabla_{\theta_m} g_n \\
        &\quad \quad \quad \quad \times \nabla_g \bigg(r_w(s_n,g,\pi_w(s_n,g_n))\vphantom{\int} + \pi_w (s_n,g) \nabla_a Q_w(s_n,g_n,a)|_{a=\pi_w(s_n,g_n)}\vphantom{\int} \bigg)\Bigg] \bigg\rvert_{g=g_n} ds_n\cdots ds_0 \\
        &= \sum_{n=0}^{\infty} \underbrace{\int_\mathcal{S} \cdots \int_\mathcal{S}}_{n+1 \text{ times}} \gamma^n p_{\theta_m, \theta_w, n}(\tau) \nabla_{\theta_m} g_n
        \nabla_g \bigg(r_w(s_n,g,\pi_w(s_n,g_n))\vphantom{\int}\\
        &\quad \quad \quad \quad + \pi_w (s_n,g) \nabla_a Q_w(s_n,g_n,a)|_{a=\pi_w(s_n,g_n)}\vphantom{\int} \bigg) \bigg\rvert_{g=g_n} ds_n\cdots ds_0 \\
        &= \mathbb{E}_{\tau \sim p_{\theta_m, \theta_w}(\tau)} \bigg[ \nabla_{\theta_m} g_t \nabla_g \bigg(r_w(s_t,g,\pi_w(s_t,g_t)) + \pi_w (s_t,g) \nabla_a Q_w(s_t,g_t,a)|_{a=\pi_w(s_t,g_t)}\vphantom{\int} \bigg) \bigg\rvert_{g=g_t} \bigg]
    \end{aligned}
    $}
\end{equation}
\end{small}
where $\tau=(s_0, a_0, s_1, a_1, \dots, s_n)$ is a trajectory and $p_{\theta_m, \theta_w, n}(\tau)$ is the (improper) discounted probability of witnessing a trajectory a set of policy parameters $\theta_m$ and $\theta_w$.

The final representation of the connected gradient formulation is then:
\begin{equation}
    \resizebox{.9\hsize}{!}{$
    \begin{aligned}
        \nabla_{\theta_m} J_m' &= \mathbb{E}_{s\sim p_\pi} \left[ \nabla_a Q_m (s,a)|_{a=\pi_m(s)}\nabla_{\theta_m} \pi_m(s) \right] \\
        & \quad + \mathbb{E}_{\tau \sim p_{\theta_m, \theta_w}(\tau)} \bigg[ \nabla_{\theta_m} g_t \nabla_g \bigg(r_w(s_t,g,\pi_w(s_t,g_t)) + \pi_w (s_t,g) \nabla_a Q_w(s_t,g_t,a)|_{a=\pi_w(s_t,g_t)}\vphantom{\int} \bigg) \bigg\rvert_{g=g_t} \bigg]
    \end{aligned}
    $}
\end{equation}

\section{Cooperative HRL as goal-constrained optimization}
\label{sec:constrained-hrl}

In this section we will derive a constrained optimization problem that motivates cooperation between a meta policy $\pi$ and a worker policy $\omega$. We will derive an update rule for the finite horizon reinforcement learning setting, and then approximate the derivation for stationary policies by dropping the time dependencies from the meta policy, worker policy, and the cooperative $\lambda$. Our goal is to find a hierarchy of policies $\pi$ and $\omega$ with maximal expected return subject to a constraint on minimum expected distance from goals proposed by $\pi$. Put formally, 
\begin{gather}
    \max_{\pi_{0:T}, \omega_{0:T}} \sum_{t = 0}^{T} \mathbb{E} \left[ r (s_{t}, a_{t}) \right] \;\text{s.t.}\; \sum_{i = t}^{T} \mathbb{E} \left[ \left\| s_{i + 1} - g_{i} \right\|_{p} \right] \leq \delta \; \forall t
\end{gather}

where $\delta$ is the desired minimum expected distance from goals proposed by $\pi$. The optimal worker policy $\omega$ without the constraint need not be goal-reaching, and so we expect the constraint to be tight in practice---this seems to be true in our experiments in this article. The hierarchy of policies at iteration $t$ may only affect the future, and so we can use approximate dynamic programming to solve for the optimal hierarchy at the last timestep, and proceed backwards in time. We write the optimization problem as iterated maximization,
\begin{gather}
    \max_{\pi_{0}, \omega_{0}} \mathbb{E} \left[ r (s_{0}, a_{0}) + \max_{\pi_{1}, \omega_{1}} \mathbb{E} \left[ \cdots + \max_{\pi_{T}, \omega_{T}} \mathbb{E} \left[ r (s_{T}, a_{T}) \right]  \right] \right]
\end{gather}

subject to a constraint on the minimum expected distance from goals proposed by $\pi$. Starting from the last time step, we convert the primal problem into a dual problem. Subject to the original constraint on minimum expected distance from goals proposed by $\pi_{T}$ at the last timestep,
\begin{gather}
    \max_{\pi_{T}, \omega_{T}} \mathbb{E} \left[ r (s_{T}, a_{T}) \right] = \min_{\lambda_{T} \geq 0} \max_{\pi_{T}, \omega_{T}} \mathbb{E} \left[ r (s_{T}, a_{T}) \right] + \lambda_{T} \delta - \lambda_{T} \sum_{i = T}^{T} \mathbb{E} \left[ \left\| s_{i + 1} - g_{i} \right\|_{p} \right]
\end{gather}

where $\lambda_{T}$ is a Lagrange multiplier for time step $T$, representing the extent of the cooperation bonus between the meta policy $\pi_{T}$ and the worker policy $\omega_{T}$ at the last time step. In the last step we applied strong duality, because the objective and constraint are linear functions of $\pi_{T}$ and $\omega_{T}$. Solving the dual problem corresponds to CHER, which trains a meta policy $\pi_{T}$ with a cooperative goal-reaching bonus weighted by $\lambda_{T}$. The optimal cooperative bonus can be found by performing minimization over a simplified objective using the optimal meta and worker policies.
\begin{gather}
    \min_{\lambda_{T}\geq 0} \lambda_{T} \delta - \lambda_{T} \sum_{i = T}^{T} \mathbb{E}_{g_{i} \sim \pi^{*}_{T} (g_{i} | s_{i}; \lambda_{T}), a_{i} \sim \omega^{*}_{T} (a_{i} | s_{i}, g_{i}; \lambda_{T}) } \left[ \left\| s_{i + 1} - g_{i} \right\|_{p} \right]
\end{gather}

By recognizing that in the finite horizon setting the expected sum of rewards is equal to the meta policy's Q function and the expected sum of distances to goals is the worker policy's Q function for deterministic policies, we can separate the dual problem into a bi-level optimization problem first over the policies. 
\begin{gather}
    \max_{\pi_{T}, \omega_{T}} Q_{m}(s_{T}, g_{T}, a_{T}) - \lambda_{T} Q_{w}(s_{T}, g_{T}, a_{T})\\
    \min_{\lambda_{T}\geq 0} \lambda_{T} \delta + \lambda_{T} Q_{w}(s_{T}, g_{T}, a_{T})
\end{gather}

By solving the iterated maximization backwards in time, solutions for $t<i\leq T$ are a constant $c_{t:T}$ with respect to meta policy $\pi_{t}$, worker policy $\omega_{t}$ and Lagrange multiplier $\lambda_{t}$. Dropping the time dependencies gives us an approximate solution to the dual problem for the stationary policies used in practice, which we parameterize using neural networks. 
\begin{gather}
    \max_{\pi, \omega} Q_{m}(s_{t}, g_{t}, a_{t}) - \lambda Q_{w}(s_{t}, g_{t}, a_{t}) + c_{t:T}\\
    \min_{\lambda\geq 0} \lambda \delta + \lambda Q_{w}(s_{t}, g_{t}, a_{t}) + c_{t:T}
\end{gather}

The final approximation we make assumes that, for a worker policy that is maximizing a mixture of the meta policy reward, and the worker goal-reaching reward, the goal-reaching term tends to be optimized the most strongly of the two, leading to the following approximation.
\begin{gather}
    \max_{\pi} Q_{m}(s_{t}, g_{t}) + \lambda \max_{\omega} Q_{w}(s_{t}, g_{t}, a_{t}) + c_{t:T}\\
    \min_{\lambda\geq 0} \lambda \delta + \lambda Q_{w}(s_{t}, g_{t}, a_{t}) + c_{t:T}
\end{gather}

\section{\methodName Algorithm}

For completeness, we provide the \methodName algorithm below.

\begin{algorithm}[H]
\captionsetup{}
\caption{\methodName}
\label{alg:training}
\begin{algorithmic}[1]
\State $\text{Initialize policy parameters}~\theta_w, \theta_m,~\text{memory}~\mathcal{D},~\text{and cooperative term}~\lambda$
\State $\text{For dynamic updates of}~\lambda,~\text{initialize}~\delta$
\While{True}
\ForEach{$t = 0,\dots,T$}
    \State $g_{t} \sim \pi_{\theta_{m}}(g_t | s_t)$ \algorithmiccomment{Sample manager action}

    \State $a_{t} \sim \pi_{\theta_{w}}(a_t| s_t, g_{t})$ \algorithmiccomment{Sample worker action}
    
    \State $s_{t+1}, r^{m}_{t} \gets \text{env}.\text{step} \; (a_t)$
    
    \State $r^{w}_{t} \gets r_{w}(s_{t}, g_{t}, s_{t + 1})$  \algorithmiccomment{Worker reward}
\EndForEach

\State $\theta_w \gets \theta_w + \alpha \nabla_{\theta_w} J_w$  \algorithmiccomment{Train worker}

\State $\theta_m \gets \theta_m + \alpha \nabla_{\theta_m} (J_m + \lambda J_w)$  \algorithmiccomment{Train manager}
    
\If{$\delta$ initialized}
    \State $\lambda \gets \lambda + \alpha \nabla_{\lambda}\left( \lambda \delta - \lambda Q_w \right)$ \algorithmiccomment{Train $\lambda$}
\EndIf
\EndWhile
\end{algorithmic}
\end{algorithm}

\section{Environment details} \label{sec:environment-details}

In this section, we highlight the environmental setup and simulation software used for each of the environments explored within this article.

\subsection{Environments} \label{sec:environments}

\paragraph{\antgather}
In this task shown in Figure~\ref{fig:antgather-env}, an agent is placed in a $20\times20$ space with $8$ apples and $8$ bombs. The agent receives a reward of $+1$ or collecting an apple (denoted as a green disk) and $-1$ for collecting a bomb (denoted as a red disk); all other actions yield a reward of $0$. Results are reported over the average return from the past $100$ episodes.

\paragraph{\antmaze}
For this task, immovable blocks are placed to confine the agent to a U-shaped corridor, shown in Figure~\ref{fig:antmaze-sim}. The agent is initialized at position $(0,0)$, and assigned an $(x,y)$ goal position between the range $(-4,-4)\times(20,20)$ at the start of every episode. The agent receives reward defined as the negative $L_2$ distance from this $(x,y)$ position. The performance of the agent is evaluated every $50,000$ steps at the positions $(16,0)$, $(16,16)$, and $(0,16)$ based on a ``success'' metric, defined as achieved if the agent is within an $L_2$ distance of $5$ from the target at the final step. This evaluation metric is averaged across $50$ episodes.

\paragraph{\antfourrooms}
An agent is placed on one corner of the classic four rooms environment~\citep{sutton1999between} and attempts to navigate to one of the other three corners. The agent is initialized at position $(0,0)$ and assigned an $(x,y)$ goal position corresponding to one of the other three corner ($(0, 20)$, $(20, 0)$, or $(20, 20)$). The agent receives reward defined as the negative $L_2$ distance from this $(x,y)$ position. ``Success'' in this environment is achieved if the agent is within an $L_2$ distance of $5$ from the target at the final step. The success rate is reported over the past $100$ episodes.


\paragraph{\ringroad}
This environment, see Figure~\ref{fig:ring-env}, is a replication of the environment presented by~\citep{wu2017flow}.
A total of 22 vehicles are place in a ring road of variable length (220-270m). In the absence of autonomy, instabilities in human-driven dynamics result in the formation of stop-and-go traffic~\citep{sugiyama2008traffic}. During training, a single vehicle is replaced by a learning agent whose accelerations are dictated by an RL policy. The vehicle perceives its speed, as well as the speed and gap between it immediate leader and follower from the five most recent time steps (resulting in a observation space of size 25). In order to maximize the throughput of the network, the agent is rewarded via a reward function defined as: $r_\text{env} = \frac{0.1}{n_v}\left[\sum_{i=1}{n_v}v_i\right]^2$, where $n_v$ is the total number of vehicles in the network, and $v_i$ is the current speed of vehicle $i$. Results are reported over the average return from the past $100$ episodes.

\begin{figure*}[h!]
\begin{subfigure}[b]{0.075\textwidth}
\begin{tikzpicture}
    \newcommand \laneheight {2.50cm}
    \newcommand \laneoffset {0.25cm}
    \newcommand \lanewidth  {0.25cm}

    \draw (0, -0.35cm) node {};


    \filldraw [black!40!white] 
    (\laneoffset, 0) rectangle 
    (\laneoffset + \lanewidth, \laneheight);

    \draw 
    (\laneoffset, 0) -- 
    (\laneoffset, \laneheight);

    \draw 
    (\laneoffset + \lanewidth, 0) -- 
    (\laneoffset + \lanewidth, \laneheight);

    
    \node[inner sep=0pt] at 
    (\laneoffset + 0.5 * \lanewidth, 0.1 * \laneheight)
    {\includegraphics[angle=90,origin=c,width=0.15cm]{figures/car.png}};
    \node[inner sep=0pt] at 
    (\laneoffset + 0.5 * \lanewidth, 0.25 * \laneheight)
    {\includegraphics[angle=90,origin=c,width=0.15cm]{figures/car.png}};
    \node[inner sep=0pt] at 
    (\laneoffset + 0.5 * \lanewidth, 0.5 * \laneheight)
    {\includegraphics[angle=90,origin=c,width=0.15cm]{figures/car.png}};
    \node[inner sep=0pt] at 
    (\laneoffset + 0.5 * \lanewidth, 0.9 * \laneheight)
    {\includegraphics[angle=90,origin=c,width=0.15cm]{figures/car.png}};


    \draw [->] 
    (0, 0.25 * \laneheight) -- 
    (0, 0.75 * \laneheight);

\end{tikzpicture}
\end{subfigure}
\hfill
\begin{subfigure}[b]{0.30\textwidth}
    \centering
    \includegraphics[height=2.5cm]{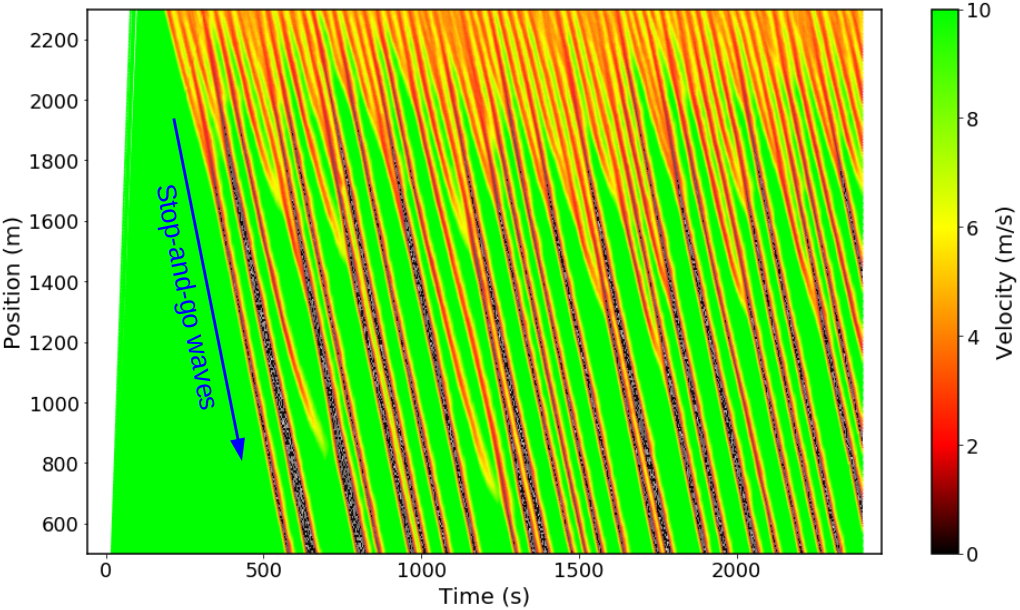}
    \caption{Human baseline}
    \label{fig:ts-no-control}
\end{subfigure}
\hfill
\begin{subfigure}[b]{0.30\textwidth}
    \centering
    \includegraphics[height=2.5cm]{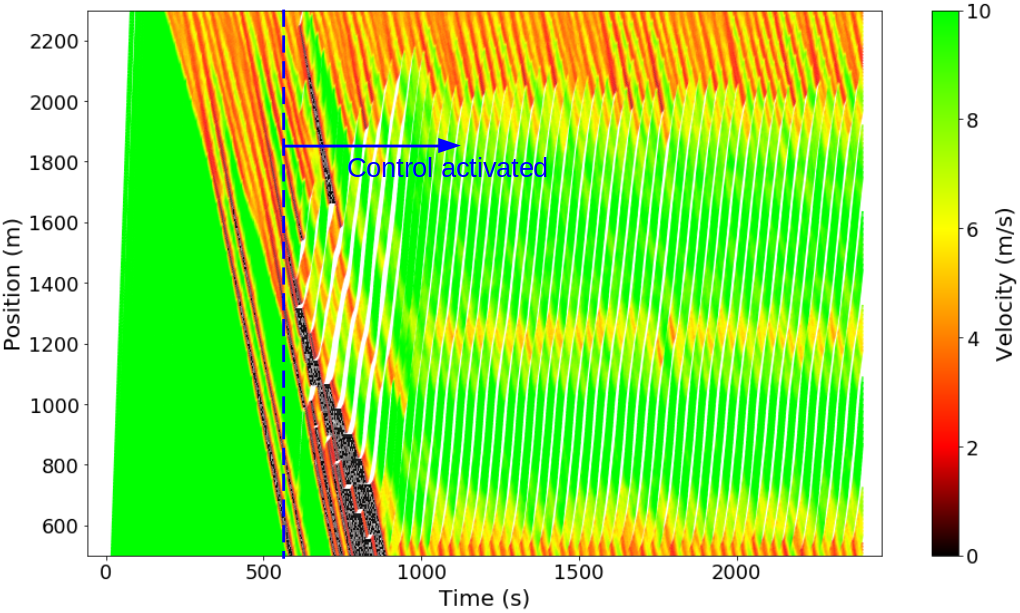}
    \caption{\methodName controller}
    \label{fig:ts-control}
\end{subfigure}
\hfill
\begin{subfigure}[b]{0.30\textwidth}
    \centering
    \includegraphics[height=2.5cm]{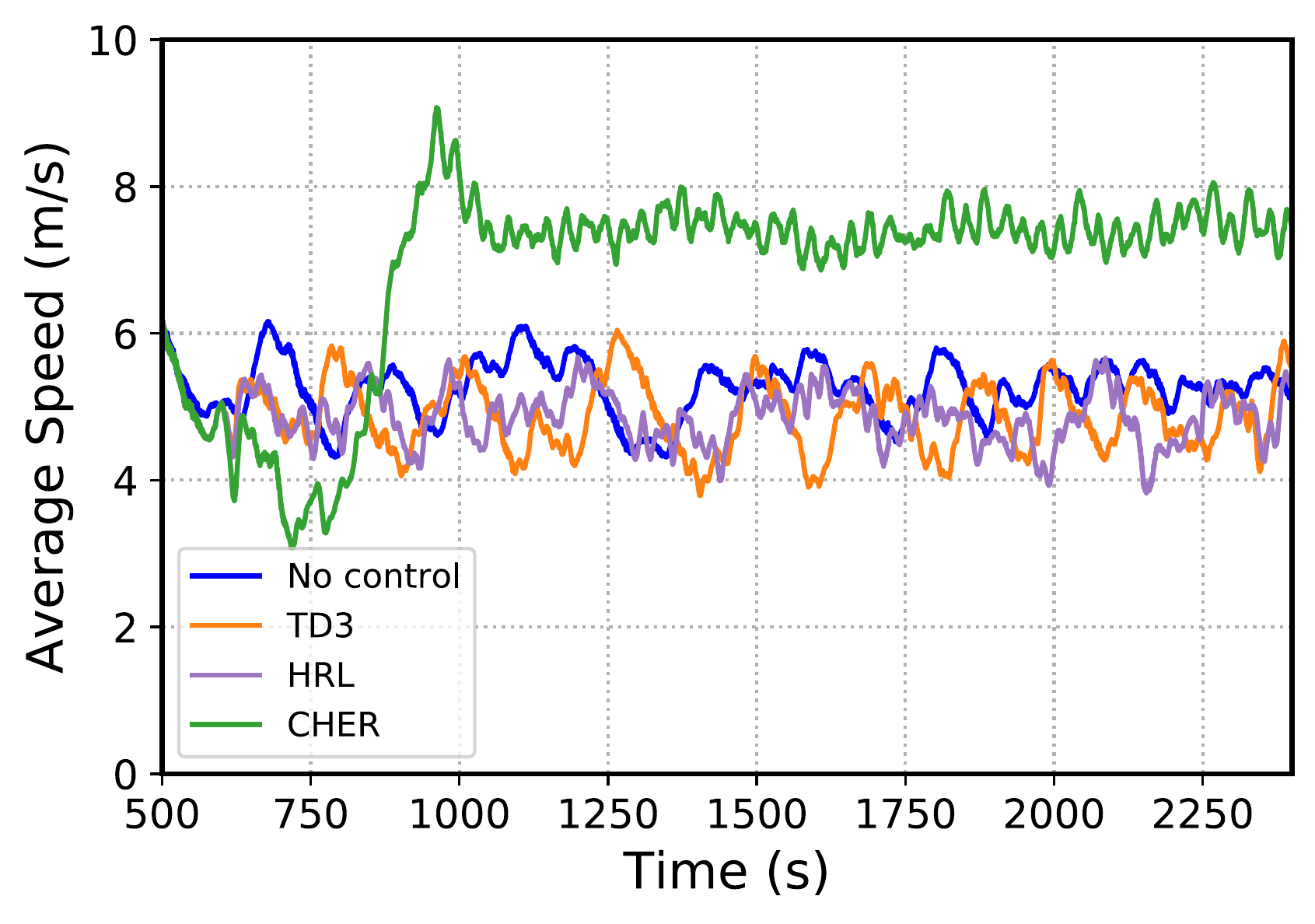}
    \caption{Average speed for each method}
    \label{fig:avg-speed}
\end{subfigure}
\caption{Traffic flow performance of vehicles within the \highwaysingle environment. \textbf{a)} In the absence of control, downstream traffic instabilities result in the propagation of congestion in the form of stop-and-go waves, seen as the red diagonal streaks. \textbf{b)} The control strategy generated via \methodName results in vehicles forming gaps that prematurely dissipate these waves. \textbf{c)} This behavior provides significant improvements to traveling speed of vehicles; in contrast, other methods are unable to improve upon the human-driven baseline.}
\label{fig:highwaysingle-results}
\end{figure*}

\paragraph{\highwaysingle}
This environment, see Figure~\ref{fig:highwaysingle-env}, is an open-network extension of the \ringroad environment. In this problem, downstream traffic instabilities resulting from a slower lane produce congestion in the form of stop-and-go waves, represented by the red diagonal lines in Figure~\ref{fig:ts-control}. The states and actions are designed to match those presented for the \ringroad environment, and is concatenated across all automated vehicles to produce a single centralized policy~\citep{kreidieh2018dissipating}. Results are reported over the average return from the past $100$ episodes.

\subsection{Simulation details} \label{sec:simulation-details}

The simulators and simulation horizons for each of the environments are as follows:
\begin{itemize}[noitemsep,leftmargin=1.5em]
    \item The \antmaze, \antfourrooms, and \antgather environments are simulated using the MuJoCo physics engine for model-based control~\citep{todorov2012mujoco}. The time horizon in each of these tasks is set to 500 steps, with dt = 0.02 and frame skip = 5.
    \item The \ringroad and \highwaysingle environments are simulated using the Flow~\citep{wu2017flow} computational framework for mixed autonomy traffic control. During resets, the simulation is warmed up for 1500 steps to allow for regular and persistent stop-and-go waves to form. The horizon of these environments are set to 1500 steps. Additional simulation parameters are for each environment are:
    \begin{itemize}
        \item \ringroad: simulation step size of 0.2 seconds/step
        \item \highwaysingle: simulation step size of 0.4 seconds/step
    \end{itemize}
\end{itemize}

Finally, the \antgather environment is terminated early if the agent falls/dies, defined as the z-coordinate of the agent being less than a certain threshold. If an agent dies, it receives a return of 0 for that time step.

\section{Algorithm details}

\subsection{Choice of hyperparamters} \label{sec:hyperparams}

\begin{itemize}[noitemsep,leftmargin=1.5em]
    \item Network shapes of (256, 256) for the actor and critic of both the manager and worker with \texttt{ReLU} nonlinearities at the hidden layers and \texttt{tanh} nonlinearity at the output layer of the actors. The output of the actors are scaled to match the desired action space.
    \item Adam optimizer; learning rate: 3e-4 for actor and critic of both the manager and worker
    \item Soft update targets $\tau =$ 5e-3 for both the manager and worker.
    \item Discount $\gamma = 0.99$ for both the manager and worker.
    \item Replay buffer size: 200,000 for both the manager and worker.
    \item Lower-level critic updates 1 environment step and delay actor and target critic updates every 2 steps.
    \item Higher-level critic updates 10 environment steps and delay actor and target critic updates every 20 steps.
    \item Huber loss function for the critic of both the manager and worker.
    \item No gradient clipping.
    \item Exploration (for both the manager and worker): 
    \begin{itemize}
        \item Initial purely random exploration for the first 10,000 time steps
        \item Gaussian noise of the form $\mathcal{N}(0, 0.1 \times \text{action range})$ for all other steps
    \end{itemize}
    \item Reward scale of 0.1 for \antmaze and \antfourrooms, 10 for \antgather, and 1 for all other tasks.
    \item Number of candidate goals = 10\footnote{This hyperparameter is only used by the HIRO algorithm.}
    \item Subgoal testing rate = 0.3\footnote{This hyperparameter is only used by the HAC algorithm.}
    \item Cooperative gradient weights ($\lambda$):
    \begin{itemize}
        \item fixed $\lambda$: We perform a hyperparameter search for $\lambda$ values between $[0.005, 0.01]$, and report the best performing policy. This corresponds to $\lambda=0.01$ for the \antgather, \ringroad, and \highwaysingle environments, and $\lambda=0.005$ for the \antmaze and \antfourrooms environments.
        \item dynamic $\lambda$: We explore varying degrees of cooperation by setting $\delta$ such then it corresponds to $25\%$, $50\%$, or $75\%$ of the worker expected return when using standard HRL, and report the best performing policy. This corresponds to $25\%$ for the \antgather environment, and $75\%$ for all other environments.
    \end{itemize}
\end{itemize}

\subsection{Manager goal assignment}

As mentioned in Section~\ref{sec:hyperparams}, the output layers of both the manager and worker policies are squashed by a \texttt{tanh} function and then scaled by the action space of the specific policy. The scaling terms for the worker policies in all environments are provided by the action space of the environment.

For the \antmaze, \antfourrooms, and \antgather environments, we follow the scaling terms utilized by~\citep{nachum2018data}. The scaling term are accordingly $\pm$10 for the desired relative x,y; $\pm$0.5 for the desired relative z; $\pm$1 for the desired relative torso orientations; and the remaining limb angle ranges are available from the ant.xml file.


Finally, for the \ringroad and \highwaysingle environments, assigned goals represent the desired speeds by controllable/automated vehicles. The range of desired speeds is set to 0-10 m/s for the \ringroad environment and 0-20 m/s for the \highwaysingle environment.

\subsection{Egocentric and absolute goals} \label{sec:ego-goals}

\citet{nachum2018data} presents a mechanism for utilizing egocentric goals as a means of scaling hierarchical learning in robotics tasks. In this settings, managers assign goals in the form of desired changes in state. In order to maintain the same absolute position of the goal regardless of state change, a goal-transition function $h(s_t,g_t,s_{t+1}) = s_t + g_t - s_{t+1}$ is used between goal-updates by the manager policy. The goal transition function is accordingly defined as:
\begin{equation} \label{eq:goal-transition}
    g_{t+1} = 
    \begin{cases}
        \pi_m(s_t) & \text{if } t \text{ mod } k = 0\\
        s_t + g_t - s_{t+1} & \text{otherwise}
    \end{cases}
\end{equation}

For the mixed-autonomy control tasks explored in this article, we find that this approach produces undesirable goal-assignment behaviors in the form of \emph{negative} desired speeds, which limit the applicability of this approach. Instead, for these tasks we utilize a more typical \emph{absolute} goal assignment strategy, defined as:
\begin{equation} \label{eq:goal-transition-fixed}
    g_{t+1} = 
    \begin{cases}
        \pi_m(s_t) & \text{if } t \text{ mod } k = 0\\
        g_t & \text{otherwise}
    \end{cases}
\end{equation}
In this case, the goal-transition function is simply $h(s_t,g_t,s_{t+1}) = g_t$.

\subsection{Worker intrinsic reward}
\label{sec:intrinsic-reward}

We utilize an intrinsic reward function used for the worker that serves to characterize the goals as desired relative changes in observations, similar to~\citet{nachum2018data}. When the manager assigns goals corresponding to \emph{absolute} desired states, the intrinsic reward is:
\begin{equation}
    r_w(s_t, g_t, s_{t+1}) = -||g_t - s_{t+1}||_2
\end{equation}

Moreover, when the manager assigns goals corresponding to \emph{egocentric} desired states, the intrinsic reward is:
\begin{equation}
    r_w(s_t, g_t, s_{t+1}) = -||s_t + g_t - s_{t+1}||_2
\end{equation}

\subsection{Reproducibility of the HIRO algorithm} \label{sec:hiro-reproducibility}

\begin{figure}[ht]
\centering
\begin{subfigure}[b]{0.45\textwidth}
    \centering
    \includegraphics[width=\linewidth]{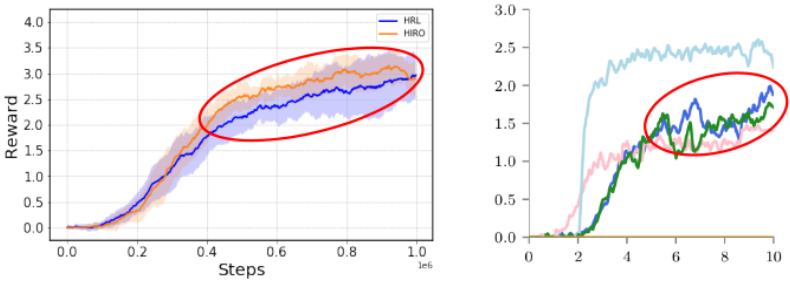}
    \caption{\antgather} 
\end{subfigure}
\hfill
\begin{subfigure}[b]{0.45\textwidth}
    \centering
    \includegraphics[width=\linewidth]{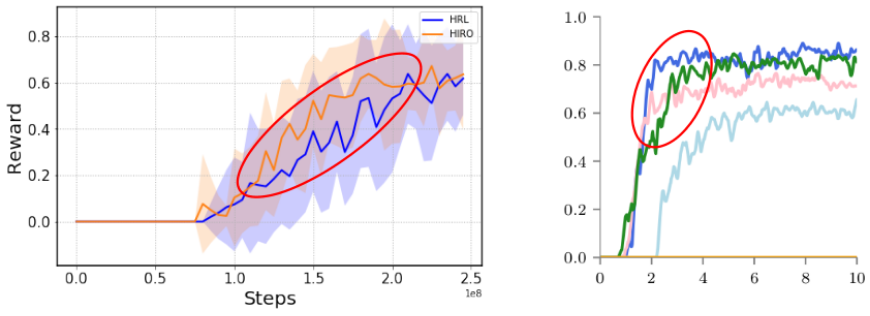}
    \caption{\antmaze} 
\end{subfigure}
\caption{Training performance of the original implementation of the HIRO algorithm with ours. The original performance of HIRO, denoted by the right figure within each subfigure, is adapted from the original article, see~\citep{nachum2018data}. While the final results do not match exactly, the relative evolution of the curves exhibit similar improvements, as seen within the regions highlighted by the red ovals.}
\label{fig:hiro-comparison}
\end{figure}

Our implementation of HIRO, in particular the use of egocentric goals and off-policy corrections as highlighted in the original paper, are adapted largely from the original open-sourced implementation of the algorithm, as available in \url{https://github.com/tensorflow/models/tree/master/research/efficient-hrl}. Both our implementation and the original results from the paper exhibit similar improvement when utilizing the off-policy correction feature, as seen in Figure~\ref{fig:hiro-comparison}, thereby suggesting that the algorithm was successfully reproduced. We note, moreover, that our implementation vastly outperforms the original HIRO implementation on the \antgather environment. Possible reasons this may be occurring could include software versioning, choice of hyperparameters, or specific differences in the implementation outside of the underlying algorithmic modification. More analysis would need to be performed to pinpoint the cause of these discrepancies.

\subsection{Hierarchical Actor-Critic with egocentric goals}
\label{sec:hac-egocentric}

To ensure that the hindsight updates proposed within the Hierarchical Actor-Critic (HAC) algorithm~\citep{levy2017hierarchical} are compared against other algorithms on a level playing field, we modify the non-primary features of this algorithm to result in otherwise similar training performance. For example, while the original HAC implementation utilizes DDPG as the underlying optimization procedure, we use TD3. Moreover, while HAC relies on binary intrinsic rewards that significantly sparsify the feedback provided to the worker policy, we on the other hand use distance from the goal.

We also extend the HAC algorithm to support the use of relative position, or egocentric, goal assignments as detailed in Appendix~\ref{sec:ego-goals}. This is done by introducing the goal-transition function $h(\cdot)$ to the hindsight goal computations when utilizing hindsight goal and action transitions. More concretely, while the original HAC implementation defines the hindsight goal $\tilde{g}_t$ at time $t$ as $\tilde{g}_t = \tilde{g}_{t+1} = \dots = \tilde{g}_{t+k} = s_{t+k}$, the hindsight goal under the relative position goal assignment formulation is $\tilde{g}_{t+i} = s_{t+k} - s_{t+i}, \ i=0,\dots,k$. This function results in the final goal $\tilde{g}_{t+k}$ before a new one is issued by the manager to equal zero, thereby denoting the worker's success in achieving its allocated goal. These same goals are used when computing the intrinsic (worker) rewards in hindsight.

\section{Additional results}

\subsection{\highwaysingle} \label{appendix:highway-results}

\subsection{Visual \antmaze} \label{sec:visual-antmaze}

\begin{figure}[ht]
    \centering
    \includegraphics[width=0.25\linewidth]{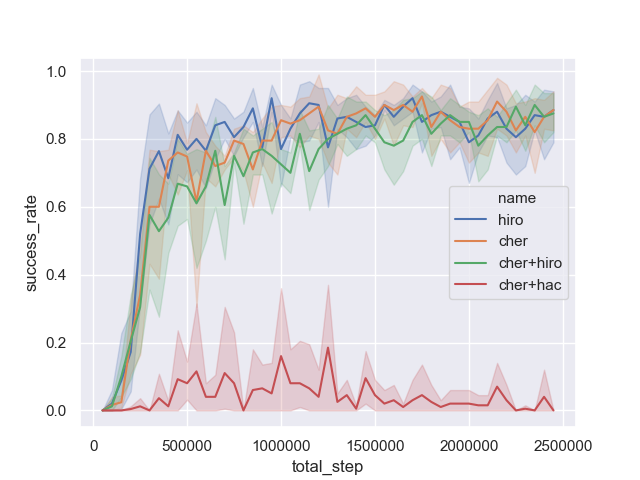}
    \quad
    \includegraphics[width=0.25\linewidth]{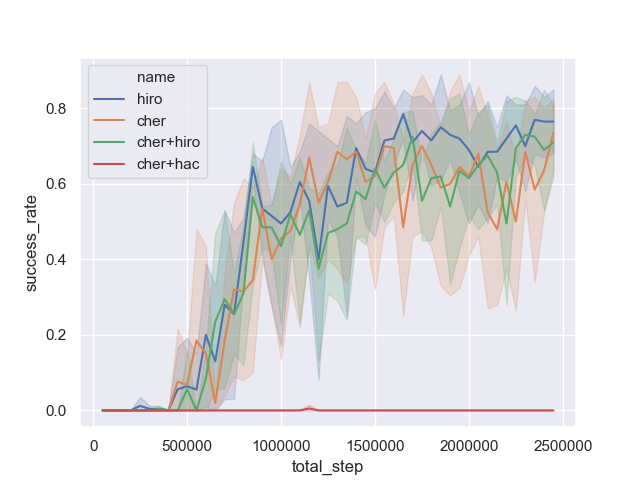}
    \quad
    \includegraphics[width=0.25\linewidth]{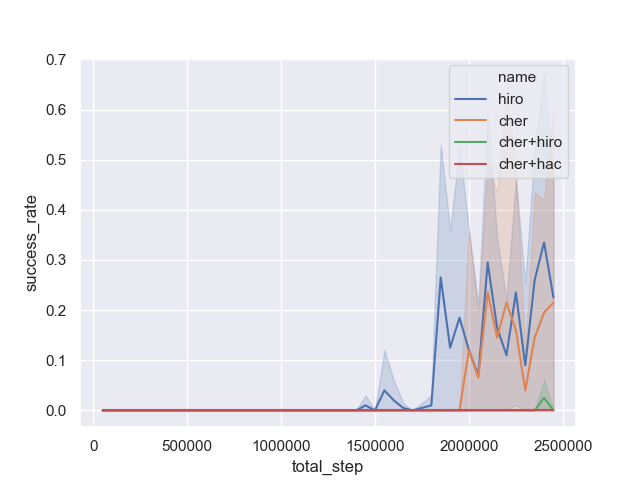}
    \quad
    \includegraphics[width=0.25\linewidth]{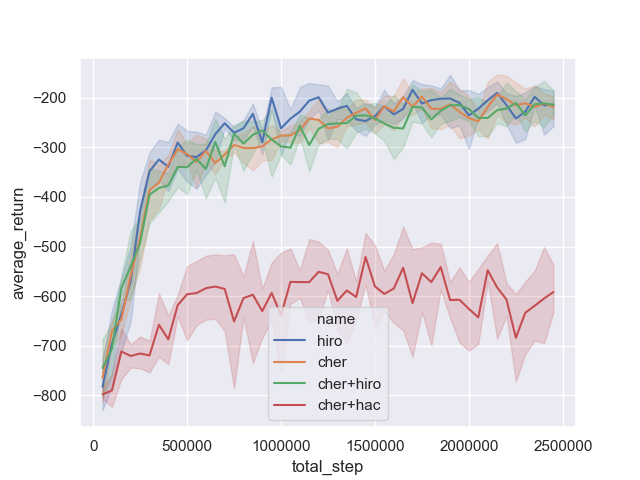}
    \quad
    \includegraphics[width=0.25\linewidth]{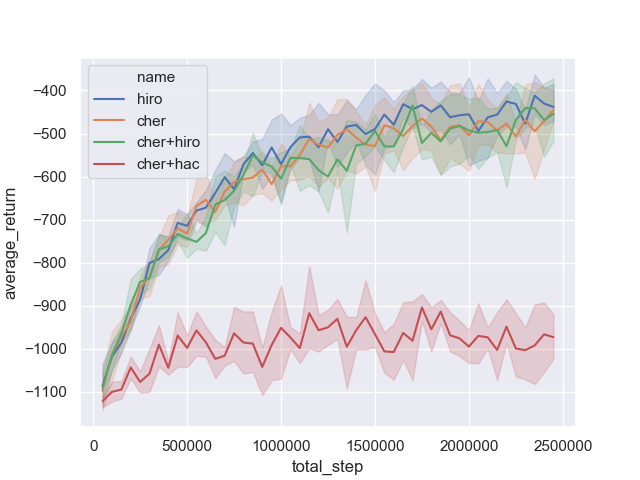}
    \quad
    \includegraphics[width=0.25\linewidth]{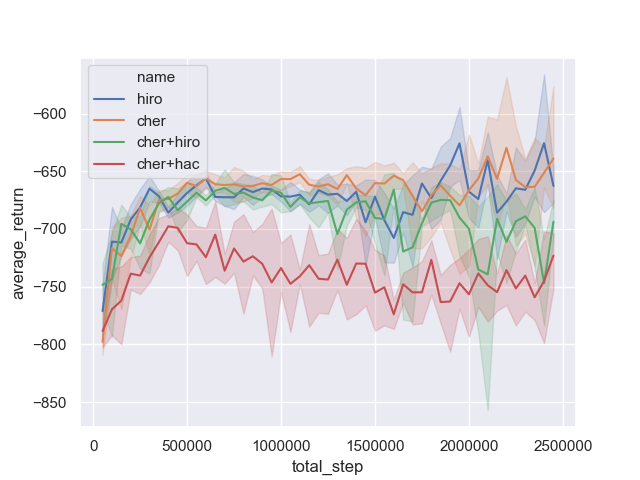}
    \caption{This figure shows the success rates, measured when the agent's center of mass enters within 5 units to the goal position, and an average return, calculated as the sum of negative distances from the agent's center of mass to the goal position. Total steps indicates the number of samples taken from the environment.}
    \label{fig:visual_ant_maze_total_steps}
\end{figure}

\begin{figure}[ht]
    \centering
    \includegraphics[width=0.25\linewidth]{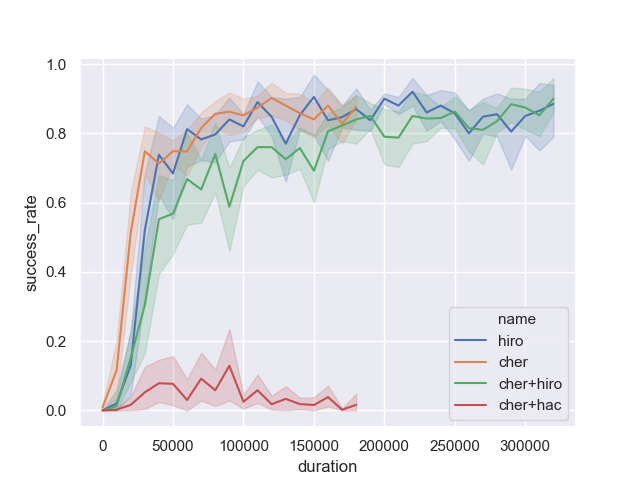}
    \quad
    \includegraphics[width=0.25\linewidth]{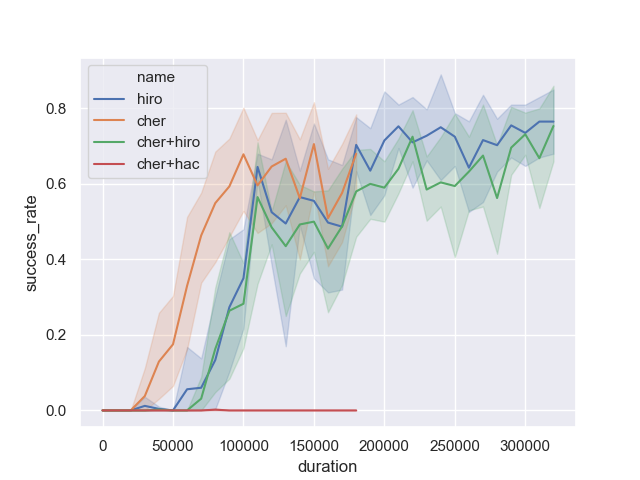}
    \quad
    \includegraphics[width=0.25\linewidth]{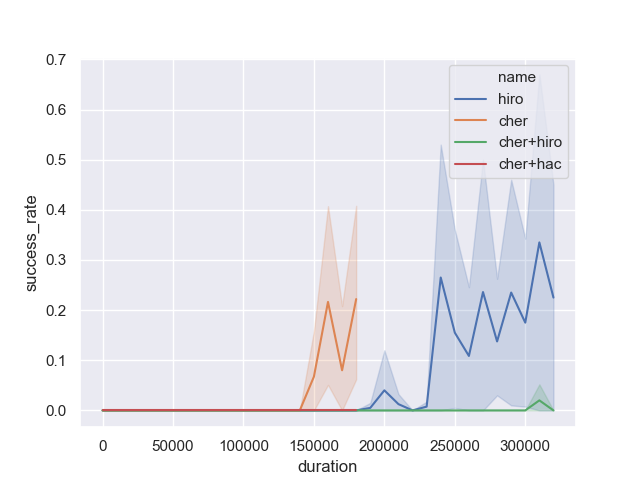}
    \quad
    \includegraphics[width=0.25\linewidth]{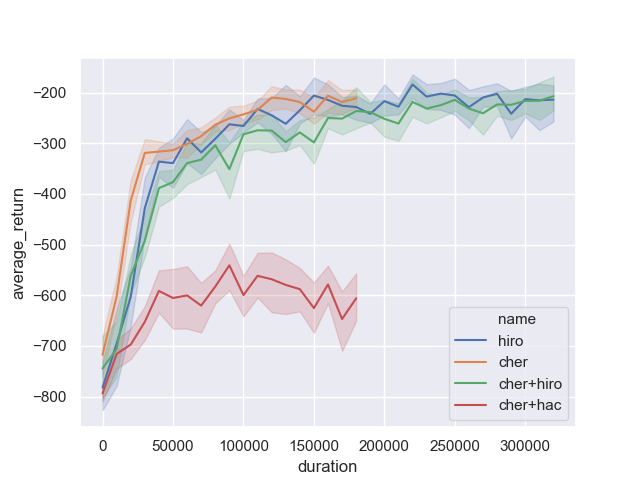}
    \quad
    \includegraphics[width=0.25\linewidth]{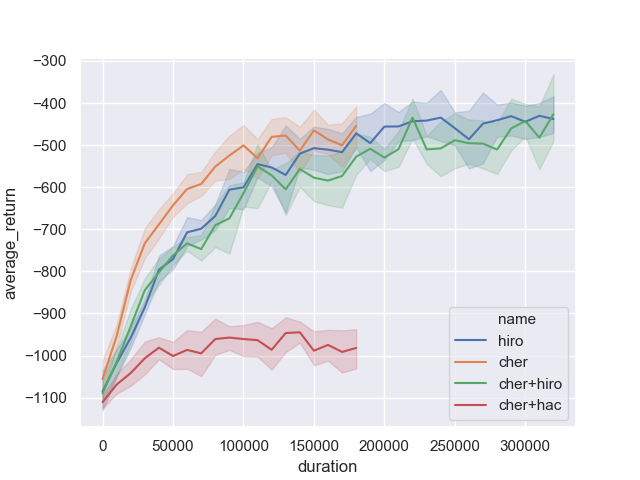}
    \quad
    \includegraphics[width=0.25\linewidth]{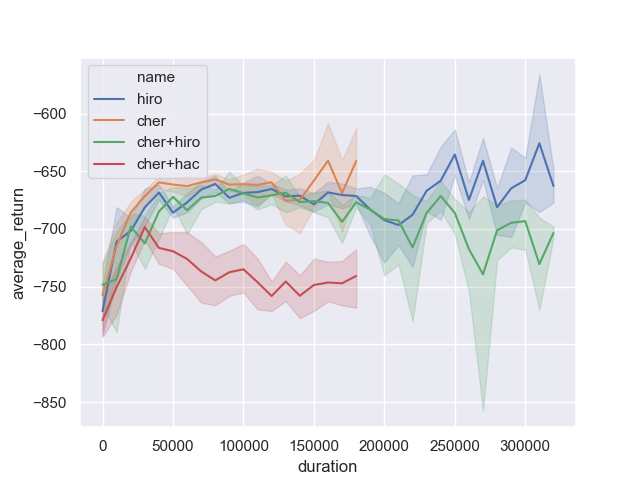}
    \caption{This figure shows the success rates, measured when the agent's center of mass enters within 5 units to the goal position, and an average return, calculated as the sum of negative distances from the agent's center of mass to the goal position. Unlike figure \ref{fig:visual_ant_maze_total_steps}, the x-axis in these plots is the duration of the experiment, measured by the wall clock time in seconds from start to 2.5 million environment steps.}
    \label{fig:visual_ant_maze_duration}
\end{figure}

Shown by figure~\ref{fig:visual_ant_maze_total_steps}, \methodName performs comparably to HIRO when trained on an image-based variant of our \antmaze environment. In this particular environment, the XY position of the agent is removed from its observation. Instead, a top down egocentric image is provided to the agent, colored such that the agent can recover the hidden XY positions as a nonlinear function of image pixels. Perhaps more interesting, while CHER achieves competitive performance with HIRO, CHER trains moderately faster than HIRO, which is shown in Figure~\ref{fig:visual_ant_maze_duration}. This is likely due to not requiring goal off policy relabelling at every step of gradient descent for the manager policy.

\end{document}